\documentclass{article}

\usepackage[nonatbib, final]{neuripsx}

\usepackage[utf8]{inputenc} %
\usepackage[T1]{fontenc}    %
\usepackage{hyperref}       %
\usepackage{url}            %
\usepackage{booktabs}       %
\usepackage{amsfonts}       %
\usepackage{nicefrac}       %
\usepackage{microtype}      %
\usepackage[dvipsnames]{xcolor}         %
\usepackage{enumerate}
\usepackage{enumitem}

\usepackage{wrapfig}

\usepackage{amsmath, amssymb, amsthm, bm}
\usepackage{math_commands}
\usepackage{cleveref}

\usepackage{macros}

\usepackage[square,numbers]{natbib}

\usepackage[ruled]{algorithm2e}

\usepackage{caption}
\usepackage{subcaption}
\usepackage{makecell}

\usepackage{tikz}
\usetikzlibrary{positioning, calc, fit, decorations.pathreplacing}

\theoremstyle{plain}
\newtheorem{theorem}{Theorem}[section]

\theoremstyle{definition}
\newtheorem{definition}[theorem]{Definition}

\crefname{claim}{claim}{claims}
\crefname{conjecture}{conjecture}{conjectures}
\crefname{assumption}{assumption}{assumptions}
\crefname{condition}{condition}{conditions}

\usepackage{xargs}
\usepackage[colorinlistoftodos,prependcaption,textsize=tiny]{todonotes}
\newcommandx{\priority}[2][1=]{\todo[linecolor=red,backgroundcolor=red!25,bordercolor=red,#1]{#2}}

\title{Tensor Programs VI: \\ Feature Learning in Infinite-Depth Neural Networks}

\author{%
  Greg Yang\thanks{Equal contribution.} \\
  xAI\\
  \And
  Dingli Yu$^*$\\
  Princeton Language\\ and Intelligence\\
  \And
  Chen Zhu \\
  Nvidia\\
  \And 
  Soufiane Hayou\thanks{Work partially done at the National University of Singapore.} \\
  Simons Institute\\ UC Berkeley\\
}

\begin{document}

\maketitle

\begin{abstract}

By classifying infinite-width neural networks and identifying the \emph{optimal} limit, \cite{yang2021tensor,yang2022tensor} demonstrated a universal way, called $\mu$P, for \emph{widthwise hyperparameter transfer}, i.e., predicting optimal hyperparameters of wide neural networks from narrow ones.
Here we investigate the analogous classification for \emph{depthwise parametrizations} of deep residual networks (resnets). 
We classify depthwise parametrizations of block multiplier and learning rate by their infinite-width-then-depth limits.
In resnets where each block has only one layer, we identify a unique optimal parametrization, called Depth-$\mu$P that extends $\mu$P and show empirically it admits depthwise hyperparameter transfer.
We identify \emph{feature diversity} as a crucial factor in deep networks, and  Depth-$\mu$P can be characterized as maximizing both feature learning and feature diversity.
Exploiting this, we find that absolute value, among all homogeneous nonlinearities,  maximizes feature diversity and indeed empirically leads to significantly better performance.
However, if each block is deeper (such as modern transformers), then we find fundamental limitations in all possible infinite-depth limits of such parametrizations, which we illustrate both theoretically and empirically on simple networks as well as Megatron transformer trained on Common Crawl.

\end{abstract} 
\section{Introduction}

Deep neural networks have showcased remarkable performance across a broad range of tasks, including image classification, game playing exemplified by AlphaGo \citep{alphago}, and natural language processing demonstrated by GPT-4 \citep{openai2023gpt4}. A prevailing trend in developing these networks is to increase their size and complexity, with empirical evidence indicating that using the same computation resources, models with more parameters tend to exhibit better performance. There are two ways to increase any network size: \emph{width} and \emph{depth}. The properties of the width (given a fixed depth) have been extensively studied in the literature: recent work by \citet{yang2022tensor} identified the \emph{Maximal Update Parametrization} ($\mu$P) that guarantees maximal feature learning in the infinite width limit.\footnote{Here maximal feature learning refers to $\Theta(1)$ change in features in the infinite width limit. This should be contrasted with the lazy training regime where the change in features is of order $\Theta(n^{-1/2})$.} Another benefit of $\mu$P is hyperparameter transfer which enables hyperparameter tuning on smaller models; the optimal hyperparameter choice for the smaller model remains optimal for larger models (i.e., models with larger width). However, despite the achievements of large-scale deep models and the theoretical understanding of scaling width, increasing the depth of neural networks still has both practical limitations and theoretical difficulties. In practice, increasing depth beyond some level often results in performance degradation and/or significant shifts in the optimal hyperparameters. In theory, unlike increasing width, increasing depth introduces new parameters that significantly change the training dynamics. In this paper, we aim to solve this problem by extending $\mu$P to include depth scaling. We call the depth scaling Depth-$\mu$P.

The issue of depth scaling has persisted over time. A decade ago, deep neural networks experienced significant degradation problems --- having more than a few dozen layers would increase the training error instead of improving the model's performance. This was partly due to the vanishing or exploding gradient problem that affects the efficient propagation of information through the network. The introduction of residual networks (ResNet)~\citep{he2016deep,he2016identity, srivastava2015highway} has partially resolved this issue, allowing for the training of deeper networks with improved performance. ResNet is constructed by layering \emph{residual blocks}, which are composed of a series of convolutional layers and then an element-wise addition with the input. This element-wise addition (commonly referred to as \emph{skip connection}) is a significant innovation of ResNet and remains an important ingredient in modern architectures including Transformers \citep{vaswani2017attention}. 

Specifically, in a residual architecture, the $l$-th residual block is formulated as
\[x^l=x^{l-1} + g^l(x^{l-1};W^l),\] 
where $x^{l-1}$ is the input, $x^l$ is the output, $W^l$ are the parameters of the block, and $g^l$ (often called the \emph{residual branch}) is a mapping that defines the layer (e.g. a stack of convolutions in ResNet, or SelfAttention and MLP in a Transformer). In this work, we focus on the case where $g^l$ is a biasless perceptron with (or without) activation.

\begin{table}[t]
\begin{center}
\caption{Difference between standard depth scaling and Depth-$\mu$P. The constants $a$ and $\eta$ in Depth-$\mu$P are transferable across depth, i.e., one can tune a smaller network and use the same constants for deeper networks. On the other hand, the learning rate of standard depth scaling requires separate tuning for models of different depth. }
\begin{tabular}{c c c c} 
 \hline
  & Branch Multiplier & Learning Rate  \\
 \hline
 \vspace{-0.7em}
 \\
 Standard & 1 & ? (tuned) \\ 

 Depth-$\mu$P (SGD) & {\color{Maroon} $a/\sqrt{\textrm{depth}}$} & {\color{Maroon} $\eta$} \\ 

 Depth-$\mu$P (Adam) & {\color{Maroon} $a/\sqrt{\textrm{depth}}$} & {\color{Maroon} $\eta/\sqrt{\textrm{depth}}$}  \\
\hline

\end{tabular}
\end{center}
\end{table}

The stacking of many residual blocks causes an obvious issue even at the initialization --- the norm of $x^l$ grows with $l$, so the last layer features do not have a stable norm when increasing the depth. Intuitively, one can stabilize these features by scaling the residual branches with a depth-dependent constant. However, scaling the residual branches with arbitrarily small constants might result in no feature learning in the large depth limit since the gradients will also be multiplied with the scaling factor. 

When each block $g^l$ has only one layer (one matrix multiply), we identify the parametrization we call Depth-$\mu$P as the optimal parametrization for deep networks.
It maximizes both \emph{feature learning} and \emph{feature diversity}\footnote{We give a formal definition of feature learning and feature diversity later in the paper.}  among all possible parametrizations of block multiplier and learning rate with depth.
Our framework extends the previous results on $\mu$P which deals with optimal width scaling \citep{yang2022tensor}. It completes the width scaling and hence provides a full width and depth scaling recipe that guarantees maximal feature learning and hyperparameter transfer across width and depth.  %
Depth-$\mu$P contains the following modifications to the standard practice:

\begin{enumerate}
    \item There is a multiplier for each residual branch before adding to its input, which is inversely proportional to the square root of $L$ (where $L$ is the depth). Formally, with a constant $a$ independent from $L$,
    \begin{align}
        x^{l}&=x^{l-1} +\frac a {\sqrt L} \cdot g^l(x^{l-1}; W^l). \label{eqn:resmlp_defn}
    \end{align}
    \item We set the learning rate of $W^l$ so that the update of $W^l$ during training is proportional to $1/\sqrt L$. We derive different learning rate schemes for different optimization algorithms based on this principle. For Adam, because it is scale-invariant to the gradient, the learning rate of $W^l$ is set to be $\eta/{\sqrt{L}}$. On the other hand, the learning rate of $W^l$ for SGD is set as a constant $\eta$ because the gradient of $W^l$ is already of size $1/\sqrt{L}$ due to the multiplier.
\end{enumerate}
In block depth $1$ (i.e., $g^{l}$ is a biasless perceptron, $W^{l}$ is a single matrix), this scaling leads to the following properties:
\begin{itemize}
    \item At the initialization, each one of the $L$ residual blocks contributes $\Theta(1/\sqrt L)$ to the main branch. These $L$ contributions are independent of each other, so the sum of them is of size $\Theta(1)$.
    \item During training, the contribution of the update of each residual block is $\Theta(1/L)$ due to the combining effect of the learning rate and multiplier. The contributions of the updates are highly correlated, so they sum up to $\Theta(1)$.
\end{itemize}
More detailed intuition of this scaling approach can be found in \Cref{sec:intuition} where we provide a simple analysis with linear networks after one gradient step. We give a complete classification of depthwise parametrizations in \cref{sec:convergence}.

\begin{wrapfigure}{r}{0.5\textwidth}
  \vspace{-1cm}
    \centering\includegraphics[width=0.5\textwidth]{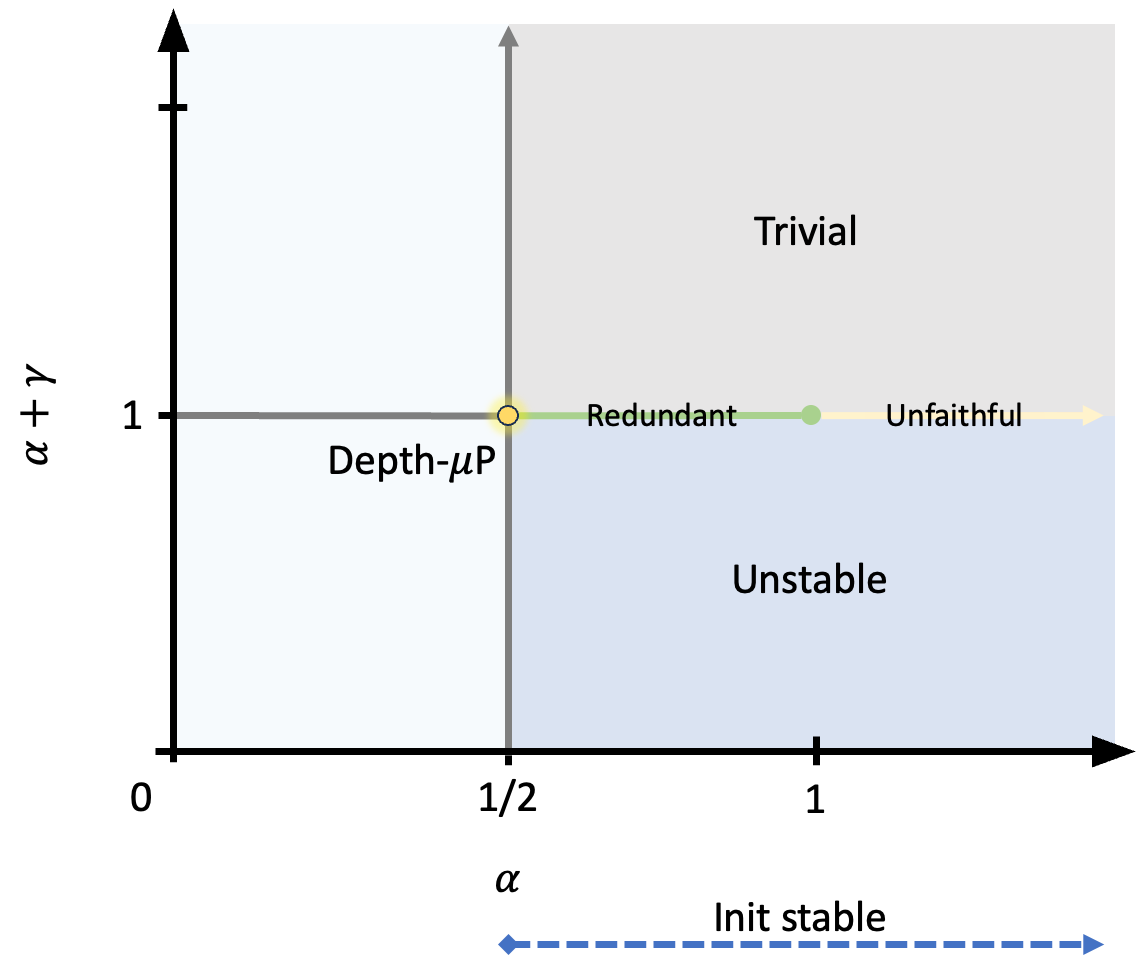}
  \caption{Behaviors of scaling strategies with a branch multiplier $L^{-\alpha}$ and parameter update proportional to $L^{-\gamma}$. }
  \label{fig:alpha-gamma}
  \vspace{-1cm}
\end{wrapfigure}
\subsection{Optimality of Depth-$\mu$P. } 

We thoroughly compare Depth-$\mu$P with other scaling strategies with a branch multiplier $\propto L^{-\alpha}$ and parameter update  $\propto L^{-\gamma}$.\footnote{It implies that the effective learning rate is proportional to $L^{-\gamma}$ for Adam and $L^{\alpha-\gamma}$ for SGD if the network is stable at initialization.}  As shown in \Cref{fig:alpha-gamma}, the space of $(\alpha, \gamma)$ is divided into several areas, each resulting in a different behavior when $L\to \infty$: 
\begin{itemize}[leftmargin=*]
    \item Having $\alpha\geq 1/2$ is required to stabilize the network at initialization. This ensures that he hidden activations and the network output do not explode at initialization;
    \item For any $\alpha+\gamma < 1$, the network is unstable during training. The change in hidden activations or the network output explodes with depth during training;
    \item For any $\alpha+\gamma > 1$, training outcome is trivial. The change of the network vanishes as depth increases;
\end{itemize}

\begin{itemize}[leftmargin=*]
    \item For any $\alpha+\gamma=1$ with $\alpha > 1$, the network is \emph{unfaithful} (a formal definition will provided later in the paper). The hidden activations explode during training as depth increases;
    \item For any $\alpha + \gamma = 1$ and $\alpha \in (1/2, 1]$, we show that the network converges to a \emph{redundant} limit that lacks \emph{feature diversity}, in that close layers have similar outputs (in a neural ODE fashion). 
    \item The only choice of $\alpha$ and $\gamma$ left is $\alpha = \gamma = 1/2$, which corresponds to Depth-$\mu$P.
\end{itemize}
The rigorous definitions and proofs are presented in \Cref{sec:convergence}. 

\subsection{Hyperparameter Transfer for Depth.} 
The optimality of Depth-$\mu$P implies (under some assumptions) that the optimal hyperparameters of the networks also converge as the depth ($L$) increases.
This convergence suggests that the optimal hyperparameters of shallower networks are approximately equal to those of deeper networks. As a direct implication, we can leverage this property to infer the hyperparameters for deeper networks from the shallower ones, effectively reducing the cost associated with hyperparameter tuning. With Depth-$\mu$P, we successfully train networks comprising thousands of residual blocks, while also showcasing the transferability of hyperparameters across depth.

\subsection{Impossibility Results for Block Depth $\ge 2$}

While the block depth 1 case admits a positive result, we show that the block depth $\ge 2$ case does not and cannot (\cref{sec:bd2}). The basic issue is the weights in different layers within a block is forced to interact additively instead of multiplicatively when depth is large, if one wants to retain diversity.
This causes block depth $\ge 2$ to have worse performance than block depth $1$ and for the optimal hyperparameters to shift with depth.
We demonstrate this pedagogically on resnet with MLP blocks but also on Megatron transformer~\citep{shoeybi2019megatron} trained on Common Crawl.
These observations entail the need to rethink the current approach to hyperparameter transfer.

\section{Related Works}
\subsection{Width Scaling and $\mu$P} The infinite-width limit of neural networks has been a topic of extensive research in the literature. Numerous studies have predominantly focused on examining the behavior of various statistical quantities at initialization. Some works have gone beyond the initialization stage to explore the dynamics of feature learning in neural networks.
\paragraph{Lazy training.} With standard parametrization, a learning rate of order $\mathcal{O}(n^{-1})$,\footnote{We also obtain the lazy infinite-width limit with the NTK parametrization and a $\mathcal{O}(n^{-1/2})$ learning rate.} $n$ being the width, yields the so-called lazy training regime in the infinite-width limit, where the features remain roughly constant throughout training \citep{chizat2020lazy, yang2022tensor}. This regime is also known as the Neural Tangent Kernel (NTK) regime and its convergence properties have been extensively studied in the literature \citep{jacot2020neural, allenzhu2019convergence, chizat2018global, zou2018stochastic}. 

\paragraph{Feature learning and $\mu$P.} Recent empirical studies (e.g. \cite{yang2022tensor}) have provided compelling evidence that feature learning plays a crucial role in the success of deep learning. It is widely acknowledged that the remarkable performance achieved by deep neural networks can be attributed to their ability to acquire meaningful representations through the process of training. Consequently, scaling the network architecture emerges as a natural choice to enhance the performance of such models.

In this context, $\mu$P (Maximal Update Parameterization), introduced in \cite{yang2022tensor}, has emerged as a promising approach for maximizing feature learning while simultaneously preventing feature explosion as the network width increases, given a fixed depth. Notably, $\mu$P facilitates hyperparameter transfer across varying network widths. This means that instead of tuning hyperparameters directly on large models, one can optimize them on smaller models and utilize the same set of hyperparameters for larger models.

The derivation of $\mu$P leverages the Tensor Programs framework \citep{yang2021tensor_i, yang2020scaling, yang2020tensor, yang2021tensor, yang2022tensor}, which provides valuable tools for capturing the behavior of neural networks in the infinite-width regime during the training process. 

\subsection{Depth Scaling} 
While increasing the width of neural networks can lead to improved performance, increasing the depth of the network also yields significant performance gains, and most state-of-the-art models use deep architectures. The introduction of skip connections \citep{he2016deep, he2016identity} played a pivotal role in enabling the training of deep networks. However, it became apparent that even with skip connections and normalization layers, training deep networks remains a challenging task \citep{liu2020understanding}. Moreover, tuning hyperparameters for large depth networks is a time-and-resource-consuming task.

To address the challenges associated with training deep networks, several studies have proposed scaling the network blocks using a depth-dependent scaler to ensure stability of features and gradients at initialization or in the kernel regime \citep{hayou21stable, hanin2018start, zhang2019fixup, noci2022signal, hayou2023infinitedepth, hayou2023width, noci2023shaped,zhang2023stabilize}. However, these works lack insights into the dynamics with feature learning. For instance, one might argue that features can still experience explosive growth if the learning rate is not properly chosen. Therefore, an effective depth scaling approach should not only ensure stability at initialization but also provide guidelines for scaling the learning rate.

This motivation underlies the development of Depth-$\mu$P, which offers a comprehensive framework for depth scaling. Depth-$\mu$P encompasses block multipliers and learning rate scaling, providing a complete recipe for training deep networks. In the case of Multi-Layer Perceptrons (MLPs) (no skip connections), \citet{jelassi2023depth} showed that a learning rate scaling of $depth^{-3/2}$ guarantees stability after the initial gradient step. However, it remains unclear how the learning rate should be adjusted beyond the first step, and this scaling is not suitable for architectures with residual connections.

\section{Warm-Up: An Intuitive Explanation with Linear Networks}\label{sec:intuition}

Let us begin with a simple example that provides the necessary intuition underpinning our depth scaling strategy. Given a depth $L$, width $n$, consider a linear residual network of the form
\begin{align*}
    x^0 & = U \xi, \\
    \forall l\in [L], \quad x^l & = x^{l - 1} + \frac1{\sqrt L} W^{l} x^{l-1},  \\
    f &= V^\top x^L,
\end{align*}
where the weight matrices $W^l \in \mathbb{R}^{n\times n}$ and $U, V$ are input and output weight matrices that we assume to be fixed during training. 

\subsection{Optimal Scaling of the Learning Rate}
To simplify the analysis, we consider gradient updates based on a single datapoint. The first gradient step is given by 
$$
W^l_1 = W^l_0 - \eta G^l_0, 
$$
where $\eta$ is the learning rate, and $G^l_0$ is a matrix with update directions. For instance, we have the following expressions for $G^l_0$ with SGD and Adam:
\begin{itemize}
    \item SGD: $G^l_0 = \frac{1}{\sqrt{L}} \del x^{l} \otimes x^{l-1}$, where $\del x^l \defeq \frac{\partial \ell}{\partial x^l}$ for some loss function $\ell$.\footnote{We use $\del$ for gradient because we want to distinguish from $d$ in depth differential equations that appear later in the paper. }
    \item Adam\footnote{For the sake of simplification, we consider SignSGD in this section, which can be seen as a memory-less version of Adam. The analysis is valid for any training algorithm that gives $\Theta(1)$ gradients.}:  $G^l_0 = \textrm{sign}\left(\frac{1}{\sqrt{L}} \del x^{l} \otimes x^{l-1}\right)$.
\end{itemize}
In both cases, $\del x^{l}$ and $x^{l-1}$ are computed for a single training datapoint $\xi_0$.
The last layer features $x^L$ (for some input $\xi$) are given by $x^L = \prod_{l=1}^L \left( I + \frac{1}{\sqrt{L}} W^l\right) x^0$.\footnote{ To avoid any confusion, here we define the matrix product by $\prod_{l=1}^L A_l = A_L \times A_{L-1} \dots \times A_1$.} We use the subscript $t$ to refer to training step. After the first gradient step, we have the following 
\begin{equation}\label{eq:first_order_linear}
x^L_1 = \prod_{l=1}^L \left( I + \frac{1}{\sqrt{L}} W^l_1\right) x^0 = x^L_0 - \frac{\eta}{\sqrt{L}} A_L + \mathcal{O}(\eta^2),
\end{equation}
where $A_L = \sum_{l=1}^L \left[\prod_{k>l} \left( I + \frac{1}{\sqrt{L}} W^k_0\right)\right] G^l_0 \left[\prod_{k<l} \left( I + \frac{1}{\sqrt{L}} W^k_0\right)\right] x^0$. 
We argue that $A_L$ behaves as $\Theta(L)$ (in $L_2$ norm). This is the due to the $1/\sqrt{L}$ scaling factor. To see this, we further simplify the analysis by considering the case $d_{in} = n = d_{out} = 1$ (single neuron per layer) and the squared loss. In this case, the term $A_L$ simplifies to 
$$
A_L = \sum_{l=1}^L \prod_{k \neq l}\left(1 + \frac{1}{\sqrt{L}} W^k_0\right) G^l_0 x_0.
$$

\paragraph{Scaling for SGD.} With SGD,  we have that $G^l_0 = \frac{1}{\sqrt{L}} \prod_{k \neq l} \left(1 + \frac{1}{\sqrt{L}} W^k_0\right) x_0 \del x^L$, where $\del x^L = (V x^L - y(\xi_0))$ and $y(\xi_0)$ is the target output. 
Therefore, it is easy to see that 
\begin{align*}
\E A_l^2 = \frac{1}{L} \E \left(\sum_{l=1}^L \prod_{k \neq l}\left(1 + \frac{1}{\sqrt{L}} W^k_0\right)^2 \del x^L x_0^2 \right)^2 = \Theta\left(\frac{1}{L} L^2\right) = \Theta(L),
\end{align*}
where we have used the fact that $\E\left(1 + \frac{1}{\sqrt{L}} W^k_0\right)^{2p} = 1 + \Theta(L^{-1})$, for any positive integer $p$.\\

Hence, the magnitude of the first order term in \cref{eq:first_order_linear} is given by 
$$
\E\left[ \left(\frac{\eta}{\sqrt{L}} A_l\right)^2\right] = \Theta(\eta^2),
$$
which shows that the update is stable in depth as long as $\eta = \Theta(1)$ in depth. More precisely, this is the maximal choice of learning rate that does not lead to exploding features as depth increases.

\paragraph{Scaling for Adam.} With Adam, we have $G^l_0 = \pm 1$, and therefore we obtain
\begin{align*}
\E A_l^2 =  \E \left(\sum_{l=1}^L \prod_{k \neq l}\left(1 + \frac{1}{\sqrt{L}} W^k_0\right) x_0 \right)^2 = \Theta\left(L^2\right),
\end{align*}
where we have used the same arguments as before. In this case, the first order term in \cref{eq:first_order_linear} is given by
$$
\E\left[ \left(\frac{\eta}{\sqrt{L}} A_l\right)^2\right] = \Theta(\eta^2 L^{-1}).
$$
Therefore, the maximal learning rate that one can choose without exploding the features is given by $\eta = \Theta(L^{-1/2})$.

\underline{\textit{Summary}:} By ensuring that parameter update is $\Theta(1/\sqrt{L})$, the features remain stable while feature update is $\Theta(1)$. This $\Theta(1)$ update is due to the accumulation of $\Theta(1/L)$ correlated terms across depth.

\subsection{Convergence when Depth goes to $\infty$}
Let us look at $x^L_1$ again in the simple case $d_{in} = d_{out} = n = 1$ and analyze its behaviour when $L \to \infty$. This paragraph is only intended to give an intuition for the convergence. A rigorous proof of such convergence will be later presented in the paper. Let us consider the case with SGD training with learning rate $\eta = 1$ and let $M_{L,l}=\prod_{k \neq l} \left(1 + \frac{1}{\sqrt{L}} W_0^k\right)$ and $\tau = (V x^L_0 - y(\xi_0))x^0$. With this, we have the following 
\begin{equation}
x^L_1 = \prod_{l=1}^L \left( 1 + \frac{1}{\sqrt{L}} W^l_0 - \frac{1}{L} \tau M_{L,l}   \right) x^0.
\end{equation}
WLOG, let us assume that $x^0_0 > 0$. Then, with high probability (the event that $W^l_0 \ll \sqrt{L}$, for some notion of ``$\ll$'', occurs with a probability of at least $1 - e^{-L^{\alpha}}$ for some $\alpha>0$)\footnote{This follows from simple concentration inequalities for sub-exponential random variables.}, we have that $x^L_1 > 0$. We can therefore look at $\log(x^L_1)$ which simplifies the task. Taking the log and using Taylor expansion under a high probability event, we obtain
\begin{equation*}
\begin{aligned}
\log(x^L_1 / x^0) &= \frac{1}{\sqrt{L}} \sum_{l=1}^L W^l_0 - \frac{1}{L} \sum_{l=1}^L \tau M_{L,l}  + \frac{\sum_{l=1}^L (W^l_0)^2}{L} + \mathcal{O}(L^{-1+\epsilon})\\
&= \frac{1}{\sqrt{L}} \sum_{l=1}^L W^l_0 -  \tau x^L_0  \frac{1}{L} \sum_{l=1}^L \frac{1}{1 + \frac{1}{\sqrt{L}}W^l_0}  + \frac{\sum_{l=1}^L (W^l_0)^2}{L} + \mathcal{O}(L^{-1+\epsilon}),
\end{aligned}
\end{equation*}
for some $\epsilon > 0$. The first and third terms $\frac{1}{\sqrt{L}} \sum_{l=1}^L W^l_0 $ and $\frac{\sum_{l=1}^L (W^l_0)^2}{L}$ converge (almost surely) to a standard Gaussian and $1$, respectively. The second term also converges naturally, since $x^L_0$ converges in $L_2$ to a Log-Normal random variable (\cite{hayou2023infinitedepth}) and with a delicate treatment (involving high probability bounds), one can show that the term $\frac{1}{L} \sum_{l=1}^L \frac{1}{1 + \frac{1}{\sqrt{L}}W^l_0}$ converges (in $L_2$ norm) at large depth. This implies that one should expect $x^L_1$ to have some notion of weak convergence as depth grows. Note that the same analysis becomes much more complicated for general width $n > 0$. To avoid dealing with high probability bounds, a convenient method consists of taking the width to infinity first $n \to \infty$, then analyzing what happens as depth increases. We discuss this in the next section.
\subsection{A Discussion on the General Case}
\paragraph{Difficulty of generalizing to the nonlinear case. } The extension to the general width scenario ($n > 1$) necessitates a more intricate treatment of the term $A_l$ to find optimal scaling rules, yet the proposed scaling remains optimal for general width. This preliminary analysis lays the groundwork for proposing a specific learning rate scaling scheme that maximizes feature learning. Moreover, demonstrating the optimality of this scaling strategy in the presence of non-linearities is a non-trivial task. The primary challenge stems from the correlation among the post-activations induced during the training process.  Overcoming these challenges requires a rigorous framework capable of addressing the large depth limit of crucial quantities in the network.

For this purpose, we employ the Tensor Program framework to investigate the behavior of essential network quantities in the infinite-width-then-depth limit. By leveraging this framework, our theoretical findings establish that the aforementioned scaling strategy remains optimal for general networks with skip connections. Our framework considers the setup where the width is taken to infinity first, followed by depth. This represents the case where $1 \ll depth \ll width$, which encompasses most practical settings (e.g. Large Language Models).

\paragraph{The critical role of  Initialization.} A naive approach to depth scaling can be as follows: since the weights $W^k_t$ might become highly correlated during training, one has to scale the blocks with $1/L$. To understand this, let us assume a block multiplier of $L^{-\alpha}$ and consider the scenario of perfect correlation where all weights are equal, i.e., $W^k_t = W$ for every $k \in {1, \dots, L}$. In this case, the last layer features can be expressed as $x^L = \left( I + L^{-\alpha} W\right)^L x_0$. When $\alpha = 1/2$, the features are likely to exhibit an explosive growth with increasing depth, while opting for $\alpha=1$ is guaranteed to stabilize the features.

However, in this paper, we demonstrate that this intuition does not align with practical observations. Contrary to expectations, the features do not undergo an explosive growth as the depth increases when $\alpha = 1/2$. This phenomenon is attributed to two key factors: random initialization and learning rate scaling with depth. These factors ensure that the weight matrices never become highly correlated in this particular fashion during the training process.

In summary, while a naive depth scaling strategy based on scaling blocks might suggest the need for $\alpha=1$ to stabilize the features, our findings reveal that in practice, this is not the case. The interplay of random initialization and learning rate scaling effectively prevents the features from experiencing explosive growth, even with the choice of $\alpha = 1/2$.

\section{SGD Training Dynamics of Infinitely Deep Linear Networks}\label{sec:linearsgd}

In this section, we continue to study the linear neural network with residual connections under Depth-$\mu$P. Using the Tensor Program framework~\citep{yang2023tensor}, we rigorously derive the training dynamics of SGD for the linear residual network when the width and the depth sequentially go to infinity.  The road map of our analysis consists the following three steps.
\begin{enumerate}
    \item We first take the width of the network to infinity by the Tensor Program framework~\cite{yang2023tensor}. As a result, instead of tracking vectors and matrices along the training trajectory, we track random variables that correspond to the vectors, that is, for a vector $x\in \R^n$ that appears in the computation of the training, the coordinates of $x$ can be viewed as iid copies of random variable $\ket x$ (called a \emph{ket}) when $n\to\infty$. \footnote{The definition of $\ket x$ requires the coordinates of $x$ is $\cO(1)$ w.r.t. $n$, and $\ket x$ is trivial if the coordinates of $x$ is $o(1)$ w.r.t. $n$. Therefore, for $x$ whose coordinates are not $\Theta(1)$, we normalize $x$ by multiplying polynomial of $n$ so the resulting vector has coordinates $\Theta(1)$. } 
    \item Since the network is linear, every random variable can be written as a linear combination of a set of zero-mean ``base'' random variables by the Master Theorem of Tensor Programs~\cite{yang2023tensor}. Therefore, we can track the random variables by analyzing the coefficients of their corresponding linear combinations, along with the covariance between the ``base'' random variables.
    \item Since the number of random variables and the number of ``base'' random variables scale linearly with $L$, the coefficients of all random variables can be represented by a six-dimensional tensor, where two of the dimensions have shape $L$. We  then map the tensor to a set of functions whose input domain is $[0,1]\times [0,1]$. Finally, we claim that the functions converge when $L\to\infty$, and identify their limits as the solution of a set of functional integrals.
\end{enumerate}
In \Cref{sec:verify_linear}, we conduct a thorough empirical verification of our theory in the linear case. The experiments clearly show the convergence of deep linear residual networks under Depth-$\mu$P. 

\paragraph{Assumptions and Notations}
Recall the linear network is given by
\begin{align*}
    x^0 & = U \xi, \\
    \forall l\in [L], \quad x^l & = \frac a{\sqrt L} W^{l} x^{l-1} + x^{l - 1},  \\
    f &= V^\top x^L.
\end{align*}
For convenience, we assume $a=1$, the SGD learning rate of $W^l$ is $1$. We add $t$ as a subscript to any notation to denote the same object but at $t$-th training step, e.g., the input at step $t$ is a single datapoint $\xi_t$, the hidden output of $l$-th layer at step $t$ is $x_t^l$, and the model output at step $t$ is $f_t$. Let $T$ be the number of training steps. Let $\ell_t$ be the loss function absorbing the label at time $t$, and $\chi_t$ be the derivative of the loss at time $t$, i.e., $\chi_t=\ell'_t(f_t)$. Let $\del x_t^l=\partial \ell_t/\partial x_t^l$, and $\tilde\del x_t^l=n \del x_t^l$ is the normalized version of $\del x_t^l$.

The Tensor Program analysis heavily depends on the scaling of initialization and learning rate of $U, V, W$ w.r.t $n$. In this paper, we use $\mu$P as the scaling w.r.t. $n$ since it maximizes feature learning in the large width limit \citep{yang2021tensor}. 
Without loss of generality, we follow \cite{yang2021tensor} and assume the input and output dimension is $1$, i.e., $\xi\in \RR$, $f\in \RR$. For a clean presentation, we additionally assume $U, V$ are frozen during training in this section and each coordinate of $W$ is initialized with i.i.d. Gaussian of variance $1/n$.   %

\subsection{Width Limit under $\mu$P}
As the first step, we take the width of the network $n$ to infinity using Tensor Programs (TP). As briefly mentioned in the road map of the section, the TP framework characterizes each vector involved in the training procedure by a random variable when $n\to\infty$. For a vector $x\in\R^{n}$ that has roughly iid coordinates, we write $\ket x \in \R$ (called a \emph{ket}) to denote a random variable such that $x$'s entries look like iid
copies of $\ket{x}$. Then for any two vector $x, y\in \R^n$ that have roughly iid coordinates, their limiting inner product by $n$ can be written as $\lim_{n\to\infty} \frac{x^\top y} n=\E \ket{x}\cdot \ket{y}$, which we write succinctly as $\braket{x}{y}$. %
Deep linear network with SGD is a simple example for this conversion from vectors to random variables. As shown in Program~\ref{alg:tp-linear}, we define a series of scalars ($\mathring f_t$ and $\mathring \chi_t$) and random variables ($\ket{U}, \ket{nV}, \ket{x_t^l}, \ket{\delta x_t^l},\ket{W_t^l x_t^{l-1}}, \ket{W_t^{l\top} \delta x_t^l}$) using the ket notations. For better understanding, we provide a brief introduction to TP below.

\SetAlgorithmName{Program}{program}{List of Programs}
\SetKwInput{KwData}{Initial random variables}
\begin{algorithm}[ht]
\caption{Random Variables induced from Tensor Program for the Linear Network with LR $\eta =1$ and frozen $U, V$. }\label{alg:tp-linear}
\KwData{$\ket{U}, \ket{nV}$ are independent standard Gaussian.}
\For{$t=0,\ldots, T-1$}{
  $\ket{x_t^0} \odefeq \xi_t \ket{U}$\;
  \For{$l=1,\ldots, L$}{
  $\ket{W_t^l x_t^{l-1}} \odefeq \ket{W_0^l x_t^{l-1}} - \frac1{\sqrt L} \sum_{s=0}^{t-1} \ket{\tilde\del x_s^{l}} \braket{x_s^{l-1}}{x_t^{l-1}}$\;
  $\ket{x_t^l} \odefeq \ket{x_t^{l-1}} + \frac1{\sqrt L} \ket{W_t^l x_t^{l-1}}$\;
  }
  $\mathring f_t \odefeq \braket{x_t^L}{nV}$\;
  $\mathring \chi_t \odefeq \ell_t'(\mathring f_t)$\;
  $\ket{\delta x_t^L} \odefeq \mathring \chi_t \ket{nV}$\;
  \For{$l=L,\ldots, 1$}{
  $\ket{W_t^{l\top} \tilde\del x_t^l} \odefeq  \ket{W_0^{l\top} \tilde\del x_t^l} - \frac1{\sqrt L} \sum_{s=0}^{t-1} \ket{x_s^{l-1}} \braket{\tilde\del x_s^l}{\tilde\del x_t^l}$\;
  $\ket{\tilde\del x_t^{l-1}} \odefeq \ket{\tilde\del x_t^l} + \frac1{\sqrt L} \ket{W_t^{l\top} \tilde\del x_t^l}$\;
  }
}
where $\ket{W_0^l x_t^{l-1}}$ and $\ket{W_0^{l\top} \tilde\del x_t^l}$ are defined in \Cref{def:zhatzdot}.
\end{algorithm}

\paragraph{Tensor Programs (TP) in a nutshell.} When training a neural network, one can think of this procedure as a process of successively creating new vectors and scalars from an initial set of random vectors and matrices (initialization weights), and some deterministic quantities (dataset in this case). In the first step, the forward propagation creates the features $x^l_0$ where the subscript $0$ refers to initialization, and the scalar $f_0$, which is the network output. In the first backward pass, the output derivative $\chi_0$ is computed, then the gradients $\del x^l_0$ are backpropagated. (Since the coordinates of  $\del x_0^l$ vanish to 0 when $n\to \infty$, TP instead tracks its normalized version $\tilde\del x_0^l\defeq n\cdot \del x_0^l$.) New vectors are created and appended to the TP as training progresses. When the width $n$ goes to infinity, vectors of size $n$ in the TP (e.g., the features $x^l_t$, and normalized gradients $\tilde\del x^l_t$) see their coordinates converge to roughly iid random variables (e.g., $\ket{x^l_t}$ and $\ket{\tilde\del x^l_t}$ in Program~\ref{alg:tp-linear}), and other scalar quantities (e.g., $f_t$ and $\chi_t$) converge to deterministic values (e.g., $\mathring f_t$ and $\mathring \chi_t$ in Program~\ref{alg:tp-linear}) under proper parametrization ($\mu$P). The Master Theorem \citep{yang2022tensor} captures the behaviour of these quantities by characterizing the \emph{infinite-width} limit of the training process. For more in-depth definitions and details about TP, we refer the reader to \cite{yang2022tensor}.

Now when we look back to Program~\ref{alg:tp-linear}, the definitions of scalars and random variables should be clear (except for $\ket{W_0^l x_t^{l-1}}$ and $\ket{W_0^{l\top} \tilde\del x_t^l}$). One can find straightforward correspondence between those and their finite counterpart, for example: \begin{itemize}
    \item $\mathring f_t$ corresponds to $f_t$, and $\mathring \chi_t$ corresponds to $\chi_t$;
    \item $\ket{x_t^l}$ corresponds to $x_t^l$ and $\ket{\tilde\del x_t^l}$ corresponds to $\tilde\del x_t^l$. (Recall $\tilde\del x_t^l=n\cdot \del x_t^l$ is the normalized version of $\del x_t^l$.)
    \item By SGD, $W_t^l = W_0^l - \frac{1}{\sqrt L}\sum_{s=0}^{t-1} \del x_s^l \otimes x_s^{l-1} $, which corresponds to $\ket{W_t^l x_t^{l-1}} = \ket{W_0^l x_t^{l-1}} - \frac1{\sqrt L} \sum_{s=0}^{t-1} \ket{\tilde\del x_s^{l}} \braket{x_s^{l-1}}{x_t^{l-1}}$.
\end{itemize}

Now we can dive into the definition of $\ket{W_0^l x_t^{l-1}}$ and $\ket{W_0^{l\top} \tilde\del x_t^l}$. Let $\cW$ be the set of initial random matrices of size $n\times n$, i.e., $\{W_0^1, \ldots, W_0^L\}$, and $\cW^\top\odefeq \{W^\top: W\in \cW\}$. Let $\cV_W$ denote the set of all vectors in training of the form $W y$ for some $y$. Then for every $W\in \cW\cup \cW^\top$, and $Wy\in \cV_W$, we can decompose $\ket{Wy}$ into the sum of $\hatket{Wy}$ and $\dotket{Wy}$, where $\hatket{Wy}$ is a random variable that act as if $W$ were independent of $y$, and $\dotket{Wy}$ is the random variable capturing the correlation part between $W$ and $y$. Specifically, let us briefly track what happens to $W^l_0 x^{l-1}_t$ during training. In the first step, we have $W^l_0 x^{l-1}_0$ which has roughly Gaussian coordinates (in the large width limit). In this case, we have $\dotket{W^l_0 x^{l-1}_0} = 0$. 
After the first backprop, we have $ \del x^{l-1}_0 = \del x^{l}_0 + \frac{1}{\sqrt{L}} W^{l\top}_0 \del x^{l}_0$, which means that the update in $W^{l-1}$ will contain a term of the form $W^{l\top}_0 z$ for some vector $z$. This implies that $W^{l}_0 x^{l-1}_1$ will contain a term of the form $W^{l}_0 W^{l\top}_0 z'$ for some vector $z'$. This term induces an additional correlation term that appears when we take the width to infinity. The $\dotket{W^{l}_0 x^{l-1}_1}$ is defined by isolating this additional correlation term from $W^{l}_0 W^{l\top}_0 z'$. The remaining term is Gaussian in the infinite-width limit, which defines the term $\hatket{W^{l}_0 x^{l-1}_1}$. Formally, we present the following definition.
\begin{defn}\label{def:zhatzdot}

    We define $\ket{Wy} \odefeq \hatket{Wy} + \dotket{Wy}$ for every $W \in \calW\cup \calW^\top$ and $Wy\in \calV_W$, where
    \begin{itemize}
        \item $\hatket{Wy}$ is a Gaussian variable with zero mean. $\forall W \in \calW\cup \calW^\top, Wy, Wz\in \calV_W$,
        \[ \Cov\left(\hatket{Wy}, \hatket{Wz}\right) \odefeq \braket{y}{z}.\]
        $\forall W, W' \in \calW\cup \calW^\top, Wy\in \calV_W, W'z\in \calV_{W'}$, $\hatket{Wy}$ and $\hatket{W'z}$ are independent if $W\neq W'$.
        $\hatket{Wy}$ is also independent from $\ket{U}$ and $\ket{nV}$.
        \item $\dotket{Wy}$ is defined to be a linear combination of $\{\ket{z}: W^\top z \in \calV_{W^\top}\}$. Then we can unwind any $\ket{y}$ inductively as a linear combination of $\hatket \bullet$, $\ket{U}$ and $\ket{nV}$, which allows us to fully define 
        \[\dotket{Wy}\odefeq\sum_{W^\top z\in \calV_{W^\top}} \ket{z} \cdot \frac{\partial \ket{y}}{\partial \hatket{W^\top z}}. \]
    \end{itemize}
\end{defn}

\subsection{Depthwise Scaling of Random Variables}

As mentioned in \Cref{def:zhatzdot}, both $\ket{x_t^l}$ and $\ket{\tilde\del x_t^{l-1}}$ can be written as linear combination of ``base'' random variables: $\{\hatket{W_0^m x_s^{m-1}}\}_{s\in \{0,\ldots, t\}, m\in [L]},\{\hatket{W_0^{m\top} \tilde\del x_s^{m}}\}_{s\in \{0,\ldots, t\}, m\in [L]}, \ket{U}$ and $\ket{nV}$. Moreover, the coefficients of the linear combinations can be calculated in a recursive way: 
by expanding $\ket{W_0^l x_t^{l-1}}$ using \Cref{def:zhatzdot}, we have 
\[\ket{x_t^l} = \ket{x_t^{l-1}} + \frac1{\sqrt L} \hatket{W_0^l x_t^{l-1}} + \frac1{\sqrt L} \sum_{s=1}^{t-1} \ket{\tilde\del x_s^{l}} \left(\frac{\partial \ket{x_t^{l-1}}}{\partial \hatket{W_0^{l\top} \tilde\del x_s^{l}}} - \frac1{\sqrt L} \braket{x_s^{l-1}}{x_t^{l-1}} \right).\]
The recursive formula for $\ket{\tilde\del x_t^l}$ is similar. 

Using this induction, we claim in the linear combinations, the coefficient of every $\hatket{\bullet}$ is $\cO(1/\sqrt{L})$, and the coefficient of $\ket{U}$ and $\ket{nV}$ is $\cO(1)$. We also claim the covariance between any pairs of random variables in the form of $\ket{x_t^l}$ and $\ket{\tilde\del x_t^{l-1}}$ is $\cO(1)$. 
\begin{prop}\label{prop:derivatives}
     $\forall t, \forall s \leq t, \forall l,m$, $\forall \ket{y}\in \{\ket{x_t^l}, \ket{\tilde\del x_t^l}\}$, 
     \[\frac{\partial \ket{y}}{\partial \hatket{W_0^mx_s^{m-1}}} = \cO\left(\frac1{\sqrt L}\right),
     \frac{\partial \ket{y}}{\partial \hatket{W_0^{m\top}\tilde\del x_s^m}} = \cO\left(\frac1{\sqrt L}\right),
     \frac{\partial \ket{y}}{\partial \ket{U}} = \cO\left(1\right),
     \frac{\partial \ket{y}}{\partial \ket{nV}} = \cO\left(1\right).\]
     $\forall t, s,l,m$, $\forall \ket{y}\in \{\ket{x_t^l}, \ket{\tilde\del x_t^l}\}$, $\forall \ket{z}\in \{\ket{x_s^m}, \ket{\tilde\del x_s^m}\}$,
     \[\braket y z=\cO(1).\]
\end{prop}

The reasoning of \Cref{prop:derivatives} is provided in \Cref{app:proof_linear}. Note the computation of covariance can also be written as a recursive formula. The reasoning relies essentially on an inductive argument. 

\subsection{Infinite Depth Limit}

Now we formalize our argument above and obtain the formula describing the dynamics of the network when $L\to \infty$. We first write the coefficients of the linear combinations as a six dimensional tensor $\mathbf \Gamma_{t,s,a,b, l, m}$, where $t,s\in \{0,\ldots, T-1\}, a,b\in \{0, 1\}, l, m\in [L]$. Specifically, $\mathbf \Gamma_{t,s,a,b, l, m}$ represents the derivative of $\ket{x_t^l}$ and  $\ket{\tilde\del x_t^l}$ w.r.t. $\hatket{W_0^mx_s^{m-1}}$ and $\hatket{W_0^{m\top}\tilde\del x_s^m}$. Here, we use $0$ to denote kets appears in the forward pass ($\ket{x_t^l}$ and $\hatket{W_0^mx_s^{m-1}}$), and $1$ to denote kets in the backward pass ($\ket{\tilde\del x_t^l}$ and $\hatket{W_0^{m\top}\tilde\del x_s^m}$). Formally, $\mathbf \Gamma_{t,s,0,0,l,m}=\frac{\partial \ket{x_t^l}}{\partial \hatket{W_0^mx_s^{m-1}}}$, $\mathbf \Gamma_{t,s,0,1,l,m}=\frac{\partial \ket{x_t^l}}{\partial \hatket{W_0^{m\top}\tilde\del x_s^m}}$, $\mathbf \Gamma_{t,s,1,0,l,m}=\frac{\partial \ket{\tilde\del x_t^l}}{\partial \hatket{W_0^{m} x_s^{m-1}}}$, $\mathbf \Gamma_{t,s,1,1,l,m}=\frac{\partial \ket{\tilde\del x_t^l}}{\partial \hatket{W_0^{m\top}\tilde\del x_s^m}}$.

However, it is hard to describe the limit of $\mathbf \Gamma$ because its size increases along with $L$. Therefore, we define the following set of functions $\{\Gamma_{t,s,a,b}:[0,1]\times (0, 1] \to \R\}_{t \in \{0,\ldots, T-1\},s\in \{-1, \ldots, t\}, a,b\in \{0,1\}}$:
For $s \geq 0$, %
\[\Gamma_{t,s,a,b}\left(p,q\right) = \sqrt L \cdot \mathbf{\Gamma}_{t,s,a,b,\lceil Lp\rceil,\lceil Lq\rceil}\]
For $s=-1$, $\Gamma_{t,-1,0,0}\left(p,q\right)=\frac{\partial \ket{x_t^{\lceil Lp\rceil}}}{\partial \ket{U}}, 
\Gamma_{t,-1,0,1}\left(p,q\right)=\frac{\partial \ket{x_t^{\lceil Lp\rceil}}}{\partial \ket{nV}}, \Gamma_{t,-1,1,0}\left(p,q\right)=\frac{\partial \ket{\tilde\del x_t^{\lceil Lp\rceil}}}{\partial \ket{U}}, 
\Gamma_{t,-1,1,1}\left(p,q\right)=\frac{\partial \ket{\tilde\del x_t^{\lceil Lp\rceil}}}{\partial \ket{nV}}.$

Here $l, m$ are normalized to $[0,1]$ so the input domain of $\Gamma$s are identical for different $L$; $\mathbf{\Gamma}_{t,s,a,b,l,m}$ is multiplied by $\sqrt L$ because $\mathbf{\Gamma}_{t,s,a,b,l,m}=\cO(1/\sqrt L)$ by \Cref{prop:derivatives}; and the extra $s=-1$ case helps us also capture the derivative w.r.t. $\ket U$ and $\ket{nV}$.

Similarly, we can also define another set of function $\{C_{t,s,a}:(0, 1]\to \R\}_{t,s\in \{-1, \ldots, T-1\}, a\in \{0,1\}}$ to describe the covariance between the ``base'' random variables:  $\forall p\in (0,1]$, let $l=\lceil Lp\rceil$,
\begin{itemize}
    \item $C_{t,s,0}\left(p\right) \odefeq \Cov(\hatket{W_0^lx_t^{l-1}}, \hatket{W_0^lx_s^{l-1}})=\braket{x_t^{l-1}}{x_s^{l-1}}$,
    \item $C_{t,s,1}\left(p\right) \odefeq \Cov(\hatket{W_0^{l\top}\tilde\del x_t^l}, \hatket{W_0^{l\top}\tilde\del x_s^l})=\braket{\tilde\del x_t^l}{\tilde\del x_s^l}$,
\end{itemize}
For $t=-1$, $C_{-1,-1,0}\left(p\right) \odefeq \Cov(\ket{U}, \ket{U})=1$, and $C_{-1,-1,1}\left(p\right) \odefeq \Cov(\ket{nV}, \ket{nV})=1$,
By \Cref{def:zhatzdot}, the ``base'' random variables of different ``groups'' are independent, so we only tracks the covariance listed above.

Using this definition of $\Gamma$ and $C$, it is convenient to write their  recursive formula in the following lemma. 

\begin{lemma}[Finite depth recursive formula for $\Gamma$ and $C$ (Informal version of \Cref{lemma:finite_depth_gamma_formal})]\label{lemma:finite_depth_gamma}
$\Gamma$ and $C$ can be computed recursively as follows:
    \begin{align*} 
     \Gamma_{t,r,0,b}\left(\frac l L, q\right) = &~\Gamma_{t,r,0,b}\left(\frac{l-1}L, q\right) + \ind_{[(t=r) \wedge (b=0) \wedge (l=\lceil L q\rceil)]} \\
     & + \frac1L \sum_{s=0}^{t-1} \Gamma_{s,r,1,b}\left(\frac l L, q\right)\left(\Gamma_{t,s,0,1}\left(\frac{l-1}L, \frac l L\right)-C_{t,s,0}\left(\frac l L\right)\right).
    \end{align*}
    \begin{align*} 
     \Gamma_{t,r,1,b}\left(\frac {l-1} L, q\right) = &~\Gamma_{t,r,1,b}\left(\frac{l}L, q\right) + \ind_{[(t=r) \wedge (b=1) \wedge (l=\lceil L q\rceil)]} \\
     & + \frac1L \sum_{s=0}^{t-1} \Gamma_{s,r,0,b}\left(\frac {l-1} L, q\right)\left(\Gamma_{t,s,1,0}\left(\frac{l}L, \frac l L\right)-C_{t,s,1}\left(\frac l L\right)\right).
     \end{align*}
     \[C_{t,s,a}(p)=\sum_{t'=-1}^{t}\sum_{s'=-1}^{s}\sum_{b\in\{0,1\}}\int_0^1 \Gamma_{t,t',a,b}(l/L,q) C_{t',s',b}(q)\Gamma_{s,s',a,b}(l/L,q) \dd q,\] where $l=\lceil Lp\rceil -1$ if $a=0$, and  $l=\lceil Lp\rceil$ if $a=1$.
\end{lemma}

The proof of \Cref{lemma:finite_depth_gamma} is straightforward from Program \ref{alg:tp-linear}. 
In \Cref{app:proof_linear}, we also give a formal proof that $\Gamma$ and $C$ converge when $L$ grows to infinity, in the case where $L$ is powers of $2$. The restriction on $L$ being powers of $2$ is imposed for the convenience of the proof, and the convergence of $\Gamma$ and $C$ is true in the general case.
Moreover, we derive the infinite depth behavior based on the recursion of $\Gamma$ and $C$ in \Cref{lemma:finite_depth_gamma}.

\begin{prop}[Infinite depth limit of $\Gamma$ and $C$ (Informal version of \Cref{prop:infinite_depth_linear_formal})]\label{prop:infinite_depth_linear}
In the limit $L\to \infty$, we  have 
\begin{align*}
  & \Gamma_{t,r,0,b}(p, q)=\ind_{[(t=r) \wedge (b=0) \wedge (p\geq q)]} + \int_0^p \sum_{s=0}^{t-1}\Gamma_{s,r,1,b}(p',q)\cdot (\Gamma_{t,s,0,1}(p', p')- C_{t,s,0}(p'))\dd p'; \\
     & \Gamma_{t,r,1,b}(p, q) = \ind_{[(t=r) \wedge (b=1) \wedge (p\leq q)]} + \int_p^1 \sum_{s=0}^{t-1} \Gamma_{s,r,0,b}(p',q) \cdot (\Gamma_{t,s,1,0}(p',p') - C_{t,s,1}(p'))\dd p';\\
     & C_{t,s,a}(p)=\sum_{t'=-1}^{t}\sum_{s'=-1}^{s}\sum_{b\in\{0,1\}}\int_0^1 \Gamma_{t,t',a,b}(p,q) C_{t',s',b}(q)\Gamma_{s,s',a,b}(p,q) \dd q.
\end{align*}
\end{prop}

The proof of \Cref{prop:infinite_depth_linear} follows from Lemma \ref{lemma:finite_depth_gamma}. A rigorous proof requires first showing the existence of a solution of the integral functional satisfied by the couple $(\Gamma, C)$. The solution is typically a fixed point of the integral functional in \Cref{prop:infinite_depth_linear}. After showing the existence, one needs to show that $(\Gamma, C)$ converges to this limit. This typically requires controlling the difference between finite-depth and infinite-depth solutions and involves obtaining upper-bounds on error propagation. The existence is guaranteed under mild conditions on the integral functional. We omit here the full proof for existence and assume that the functional is sufficiently well-behaved for this convergence result to hold. The formal proof of the convergence of $\Gamma$ and $C$ for $L=2^k$ ($k\in \N$) in \Cref{app:proof_linear} is a showcase of the correctness of the proposition.

This gives a convergence in distribution:

\begin{theorem}
    In the $L \to \infty$ limit, the kets $\ket{x^L_s}, s = 0, 1, \ldots,$ converge in distribution as a zero-mean Gaussian process with kernel
    \[\braket{x^L_s}{x^L_t} = C_{t,s,1}(1).\]
    Thus, for each fixed neuron index $\alpha$, the collection $\{x^L_{\alpha s}\}_{s\ge 0}$ converges in distribution to a zero-mean Gaussian process with kernel $C_{t,s,1}(1)$ in the $n\to\infty$ then $L\to\infty$ limit.
\end{theorem}

For audience familiar with stochastic processes, we in fact have a weak convergence of the entire continuous-depth-indexed process $\{\ket{x^{p}_s}, \ket{\delta x^{p}_s}\}_{p \in [0, 1],s\ge0}$ in the Skorohod topology.

\section{What Causes Hyperparameter Transfer?}\label{sec:cause_hp}

In a popular misconception, hyperparameter transfer is implied by the existence of a limit.
For example, the fact that $\mu$P transfers hyperparameters, in this misconception, is because of the existence of the feature learning limit (aka the $\mu$ limit), the limit of $\mu$P as width goes to infinity.
However, this is not the case.
Indeed, there are a plethora of infinite-width limits, such as the NTK limit, but there can only be one way how the optimal hyperparameters scale, so existence cannot imply transfer.
In a stronger version of this misconception, transfer is implied by the existence of a ``feature learning'' limit. But again, this is False, because there are infinite number of feature learning limits (where the $\mu$ limit is the unique maximal one).

Instead, what is true is that the \emph{optimal} limit implies the transfer of \emph{optimal} hyperparameters.
For example, in the width limit case, $\mu$P is the unique parametrization that yields a maximal feature learning limit. Compared to all other limits, this is obviously the optimal one. Hence $\mu$P can transfer hyperparameters across width.

So far, there is no \emph{a priori} definition for the ``optimality'' of a limit: One can only tell by \emph{classifying} all possible limits; it turns out only a small number of different behavior can occur in the limit, and thus one can manually inspect for which limit is the optimal one.

Similarly, in this work, to \emph{derive} a depthwise scaling that allows transfer, we need to \emph{classify} all possible infinite depth limits --- and Depth-$\mu$P will turn out to be optimal in a sense that we define later in the paper.%
\footnote{There are important nuances here that will be spelled out in an upcoming paper. For example, if the space of hyperparameters is not chosen correctly, then it could appear that no limit is \emph{optimal} in any manner. For example, if one in (widthwise) SP, one only thinks about the 1D space of the global learning rate, then all infinite-width limits are defective --- and indeed there is no hyperparameter transfer where the bigger always does better.}
More interestingly than the width case, here we have multiple modes of feature learning when taking the depth limit and it is important to discern which mode of feature learning is optimal.
Thus, again, it is \emph{insufficient} to derive any one limit, even with feature learning, and be able to infer it yields HP transfer.

In \cref{sec:exp}, we provide experiments with $1/L$ block scaling $(\alpha, \gamma) = (1, 0)$, aka ODE scaling, which provably induces feature learning in the infinite-depth limit, but is sub-optimal. Our results show a significant shift in the optimal learning rate with this parametrization. 
\section{Preliminaries for the General Case}\label{sec:preliminaries}
For the general case, we recall and extend the notation from the previous sections and also define new ones.
\paragraph{Notation} Let $L$ be the depth of the network, i.e., the number of residual blocks, and $n$ be the width of the network, i.e. the dimension of all hidden representations $x^0, \ldots, x^L$. Let $\xi\in \RR^\din$ be the input of the network, $U \in \RR^{n\times \din}$ be the input layer, and $V\in \RR^{n\times\dout}$ be the output layer, so that $x^0 = U\xi$ and the model output w.r.t. $\xi$ is $f(\xi)\triangleq V^\top x^L$.  Let $\ell$ be the loss function absorbing the label, and $\del x^l$ be the gradient of $x^l$ w.r.t. the loss. 
We denote variables at $t$-th training step by adding $t$ as a subscript, e.g., the input at step $t$ is $\xi_t$\footnote{Here, the input is used to perform one gradient step at training step $t$. We will see later that our claims should in principle hold for batched versions of the training algorithm.}, the hidden representation of $l$-th layer at step $t$ is $x_t^l$, and the model output at step $t$ is $f_t$. Let $T$ be the number of training steps. 

\subsection{Unified Scaling for SGD, Adam, and All Entrywise Optimizers}\label{sec:unified_scaling}

We extend the definition of entrywise update (\cite{yang2023tensor}) for depth scaling, allowing us to study the unified depth scaling for SGD, Adam, and other optimization algorithms that perform only entrywise operations. 

\begin{definition}\label{def:unified-scaling}
A gradient-based update of parameter $w$ with both width and depth scaling is defined by a set of functions $\QQ = \{Q_t:\R^{t+1}\to\R\}_{t\ge0}$, and $c, d, \delta, \gamma, \eta$.
The update at time $t$ of the optimization is 
\[
w \gets w -\eta n^{-c}L^{-\gamma} Q_t(n^dL^\delta g_0, \ldots, n^dL^\delta g_t),
\]
where $g_{s},s=0,\ldots,t$, are the gradients of $w$ at time $s$.
\end{definition}

For SGD, $Q_t(n^dL^\delta g_0, \ldots, n^dL^\delta g_t)=n^dL^\delta g_t$, and the ``true'' learning rate is $\eta n^{-c+d} L^{-\gamma+\delta}$.
For Adam, \[Q_t(n^dL^\delta g_0, \ldots, n^dL^\delta g_t) = \frac{\frac{1-\beta_1 }{1-\beta_1^{t+1}}\sum_{s=0}^t \beta_1^{t-s} n^dL^\delta g_s}{ \sqrt{\frac{1-\beta_2 }{1-\beta_2^{t+1}}\sum_{s=0}^t \beta_2^{t-s} (n^dL^\delta g_s)^2 + \epsilon}},\]
and the ``true'' learning rate is $\eta n^{-c} L^{-\gamma}$. 

The purpose of multiplying the gradients $n^dL^\delta$ before $Q_t$ is to make sure the inputs to $Q_t$ are $\Theta(1)$ w.r.t. $n$ and $L$\footnote{It is called faithfulness in \citet{yang2023tensor}.}; otherwise, the update might be trivial when $n$ and $L$ become large. For example, if gradients are $o(1)$ entrywise, then, in Adam, directly feeding gradients to $Q_t$ will always give an output of $0$ because of the constant $\epsilon > 0$. 

In this paper, we will only consider $d, \delta$ such that $n^d L^\delta g$ is $\Theta(1)$.\footnote{Note $c, d, \delta, \gamma, \eta$ in \Cref{def:unified-scaling} can be different for parameters, so it is possible to make every parameter to satisfy the condition.} As a result, the output of $Q_t$ is also $\Theta(1)$ in general. Therefore, $n^{-c}L^{-\gamma}$ decides the scale of the update and should be our focus. We call $\eta n^{-c}L^{-\gamma}$ the \emph{effective learning rate}. 

\subsection{$\mu$P and Widthwise Scaling}

Maximal update parametrization ($\mu$P) \citep{yang2020tensor} considers the change of initialization and learning rate of each weight matrix in the network when width scales up.\footnote{Reparametrization is also included in the original $\mu$P, but it is not necessary for the purpose of this paper. } It provides a unique initialization and learning rate of each weight matrix as a function of width $n$ that makes the update of each weight matrix maximal (up to a constant factor). The benefit of $\mu$P is not only the theoretical guarantee but also the hyperparameter stability when scaling up the width
~\citep{yang2021tensor}.

In this paper, we assume the widthwise scaling follows $\mu$P. That is, the $c$ in the effective learning rate $\eta n^{-c}L^{-\gamma}$ and the initialization variance of each weight matrix follows \Cref{tab:mup}. 
\begin{table}[ht]
    \centering
    \caption{Widthwise scaling of $\mu$P, where $c$ (defined in \Cref{def:unified-scaling}) describes the widthwise scaling of the effective learning rate.}
    \begin{tabular}{c|c|c|c}
    \Xhline{2\arrayrulewidth}
        &  Input weights  &  Output weights & Hidden weights \\\hline
       Init. Var. & $1$ & $n^{-2}$ & $n^{-1}$ \\\hline
       $c$  & $0$ & $1$ & $1$\\\Xhline{2\arrayrulewidth}
    \end{tabular}
    \label{tab:mup}
\end{table}

\subsection{Our Setup}

\global\long\def\MS{\mathrm{MS}}%

We consider an $L$-hidden-layer residual network with biasless perceptron blocks:
\begin{align*}
    x^0 & = U \xi, \\
    \forall l\in [L], \quad x^l & = L^{-\alpha} \, \MS(\phi(h^l)) + x^{l - 1}, \quad h^l=W^l x^{l-1},\\
    f &= V^\trsp x^L.
\end{align*}
where $\MS$ refers to Mean Subtraction and is given by $\MS(x)=x-\langle x,1\rangle/n=Gx$ with $G=I-11^{\top}/n$,
for any $x\in\R^{n}$.
The initialization and learning rate of $U, V$ follows $\mu$P. The initialization of $W^l$ follows $\mu$P, and the learning rate of $W^l$ is $\eta n^{-1} L^{-\gamma}$.
\paragraph{Mean Subtraction ($\MS$).} 
In general, without mean subtraction, the mean of $\phi$ will dominate the depthwise dynamics. For example, when $\phi$ is relu, each layer will only add nonnegative quantities to $x^l$ that on average is positive. Its accumulation over depth either causes the network output to blow up if the multiplier $L^{-\alpha}$ is too large, or lack feature diversity otherwise.
As we shall see, mean subtraction removes this failure mode and enable more powerful infinite-depth limits.%
\footnote{Note that using an \emph{odd} nonlinearity will also achieve similar results because they have no mean under a symmetrically distributed input, which is approximately the case for $h^l$ throughout training. This is the case for $\phi$ = identity that we discussed earlier. But it turns out odd nonlinearities minimize feature diversity, so mean subtraction is a much better solution.}

\begin{definition}\label{defn:abcd}
  Fix a set of update functions $\QQ = \{Q_t:\R^{t+1}\to\R\}_{t\ge0}$.
  A \emph{depthwise parametrization} of the MLP residual network above is specified by a set of numbers $\{\alpha, \gamma, \delta\}$
  such that
  \begin{enumerate}[label=(\alph*)]
  \item We independently initialize each entry of $W^{l}$ from $\Gaus(0,n^{-1} )$
  \item The gradients of $W^{l}$ are multiplied by $n L^{\delta}$ before being
  processed by $Q_t$: i.e., the update at time $t$ is
  \begin{equation}
  W^{l}\gets W^{l}-\eta n^{-1} L^{-\gamma}Q^l_t(n L^{\delta} g_{0},\ldots,n L^{\delta} g_{t})
  \label{eqn:abcdupdate}
  \end{equation}
  where $g_{s},s=0,\ldots,t$, are the gradients of $W^{l}$ at time
  $s$ and $Q_t$ is applied entrywise.
  \end{enumerate}
\end{definition}

\paragraph{Miscellaneous notations.}  For a vector $x$, let $[x]_i$ be its $i$-th coordinate. For a matrix $M$, let $[M]_i$ be its $i$-th row. Let $I$ be the identity matrix, and $\one$ be the full one vector. For $m\in \N^+$, let $[m] = \{1, \ldots, m\}$. Let $\otimes$ be the Kronecker product.

\global\long\def\MS{\mathrm{MS}}%

\global\long\def\rav{\check{\ra}}%

\global\long\def\lav{\check{\la}}%

\global\long\def\dket#1{\ob#1 \rav}%

\global\long\def\dbra#1{\lav#1 \ob}%

\global\long\def\dotob{\dot{\ob}}%

\global\long\def\hatob{\hat{\ob}}%

\global\long\def\hatoplim#1{\hatob#1 \hatob}%

\global\long\def\dotoplim#1{\dotob#1 \dotob}%
\section{Classification of Depthwise Parametrizations}\label{sec:convergence}

In this section, we provide a comprehensive description of the impact of depth parametrization on stability and update size. For this purpose, we only have two scalings to keep track of: the branch multiplier and
the learning rate scaling because the initialization scale is fixed
by the faithfulness property (defined below). Requiring that the features don't blow up at initialization means that the branch multipliers must be at
most $\Theta(1/\sqrt{L})$. Assuming the updates are faithful (i.e.,
input to gradient processing functions are $\Theta(1)$ entrywise),
the update size can be at most $1/L$ for the hidden layers, by an
(Jacobian) operator-norm argument, but potentially much less. Naively
speaking, there can be a trade-off between update size and initialization:
if initialization is large, then the update may need to be small so
as not to blow up the other parts of the network; likewise if the
initialization is small, then the update size can be larger. But one
may be surprised that a careful calculation shows that there is no
trade-off: we can maximize both initialization and update size at
the same time.

Before delving into the details, let us first define the notions of training routine, stability, faithfulness, and non-triviality. Hereafter, all the aymptotic notations such as $\cO$, $\Omega$ and $o$ should be understood in the limit ``$n \to \infty, \textrm{ then } L \to \infty$''. For random variables, such notations should be understood in the sense of weak convergence (convergence in distribution). When we use the notation $x = \cO(1)$ for some vector $x = (x_1, \dots, x_n) \in \mathbb{R}^n$, it should understood in the sense that for all $i \in [n], x_i = \cO(1)$. Lastly, we will use bold characters (e.g. $\hh$ instead of $h$) to denote `batched' versions of the quantities. This is just to emphasize that the following claims should hold for batched quantities as well.

\emph{Remark:} in this section, we state the results as ``claims'' instead of theorems. In Appendix~\ref{sec:claim_justifications}, we provide ``heuristic'' proofs that can be made rigorous under non-trivial technical conditions. We also showcase the correctness of the claims by proving them rigorously in our linear setting in \Cref{sec:linearsgd_classification}. We believe this additional layer of complexity is unneeded and does not serve the purpose of this paper. 

\begin{defn}[Training routine]\label{defn:training_routine}
    A training routine is the package of $\eta$, $\QQ$, and the input batches.
\end{defn}

\begin{defn}[Stability]\label{defn:stability} 
    We say a parametrization is
    \begin{enumerate}
    \item \emph{stable at initialization} if %
    \begin{equation}
    \hh_{0}^{l},\xx_{0}^{l}=\cO(1),\forall l\in[L],\quad\text{and}\quad \ff_{0}=\cO(1).\label{eq:initstable}
    \end{equation}
    \item \emph{stable during training} if for any training routine, any time $t\ge0$, $l\in[L]$, we have 
    \[
    \Delta \hh_{t}^{l},\Delta \xx_{t}^{l}=\cO(1),\forall l\in[L],\quad\text{and}\quad \Delta\ff_{t}=\cO(1),
    \]
     where the symbol `$\Delta$' refers to the change after one gradient step.\\
    \end{enumerate}
   
    We say the parametrization is \emph{stable} if it is stable both at initialization and during training.
\end{defn}

\begin{defn}[Faithful]\label{defn:faithful} 
    We say a parametrization is \emph{faithful at step $t$} if $\hh_{t}^{l}=\Theta(1)$ for all $l\in[L]$.
    We say the parametrization is \emph{faithful} if it is faithful for all $t$.
    We also say it is \emph{faithful at initialization} (resp.\ faithful during training) if this is true at $t = 0$ (resp.\ for $t \ge 1$).
\end{defn}
Note faithfulness here refers to ``faithfulness to $\phi$'', meaning the input to $\phi$ is $\Theta(1)$. This is different from the definition of faithfulness in \citet{yang2023tensor}, where faithfulness refers to ``faithfulness to $Q$'' meaning the input to $Q$ is $\Theta(1)$. ``faithfulness to $Q$'' is already assumed in this work as mentioned in \Cref{sec:unified_scaling}.

\begin{defn}[Nontriviality]
    We say a parametrization is \emph{trivial} if for every training routine and any time $t\ge1$, $\ff_{t}-\ff_{0}\asto0$ in the limit ``$n\to\infty,$ then $L \to \infty$'' (i.e., the function does not evolve in the infinite-width-then-depth limit). We say the parametrization is \emph{nontrivial} otherwise.
  \end{defn}

\begin{defn}[Feature Learning]
    We say a parametrization induces \emph{feature learning} in the limit ``$n \to \infty$, then $L \to \infty$'', if there exist a training routine, and $t\ge1$, and any $\lambda > 0$, we have $\Delta\hh_{t}^{\lfloor\lambda L\rfloor} = \Theta(1)$.
  \end{defn}

\subsection{Main Claims}
 We are now ready to state the main results. The next claim provides a necessary and sufficient condition under which a parametrization is stable at initialization.

\begin{claim}\label{clm:stability_init}
    A parametrization is stable at initialization iff $\alpha\ge1/2$.
\end{claim}
\Cref{clm:stability_init} is not new and similar results were reported by \citet{hayou21stable}. However, \citet{hayou21stable} focuses on initialization and lacks a similar stability analysis during training. In the next result, we identify two different behaviours depending on the scaling of the learning rate.
\begin{claim}\label{clm:stable_non_trivial}
Consider a parametrization that is stable at initialization. Then the following hold (separately from each other).
\begin{itemize}
    \item It is stable during training as well iff $\alpha + \gamma \ge 1$.
    \item It is nontrivial iff $\alpha + \gamma \le 1$.
\end{itemize}
Therefore, it is both stable and nontrivial iff $\alpha + \gamma = 1$.
\end{claim}

From \Cref{clm:stability_init} and \Cref{clm:stable_non_trivial}, having $\alpha+\gamma=1$ and $\alpha\ge1/2$ is a necessary and sufficient condition for a parametrization to be stable and nontrivial throughout training. In the next result, we therefore restrict our analysis to such parametrizations and study their faithfulness. 
\begin{claim}\label{clm:faithful}
    Consider a stable and nontrivial parametrization. The following hold (separately from each other).
    \begin{itemize}
        \item It is faithful at initialization iff $\alpha \geq 1/2$. As a result, $\alpha=1/2$ is the minimal choice of $\alpha$ that guarantees faithfulness.
        \item It is faithful during training iff $\alpha \le 1$.
    \end{itemize}
Therefore, a stable and nontrivial parametrization is faithful iff $\alpha \in [1/2,1]$.
\end{claim}
The first claim follows from well-known calculations of randomly initialized residual networks \cite{hayou21stable}.
For the second claim, the intuition here is just that if $\alpha+\gamma=1$ and $\alpha>1$
then $\gamma<0$, i.e., the update size blows up with depth. This
would then cause the input to the nonlinearities to blow up with size.

One might argue that faithfulness at initialization is not important (e.g. features at initialization could converge to zero without any stability or triviality issues) and what matters is faithfulness throughout training. It turns out that faithfulness at initialization plays a crucial role in the optimal use of network capacity. To see this, we first define the notion of feature diversity exponent, which relates to the similarity in the features of adjacent layers.
\begin{defn}[Feature Diversity Exponent]
We say a parametrization has feature diversity exponent $\kappa\geq 0$ if $\kappa$ is the maixmal value such that for all $\lambda \in [0, 1]$ and sufficiently small $\epsilon > 0$, and all time $t$, 
\[\frac{1}{\sqrt{n}} \left\| \xx^{\lfloor (\lambda + \epsilon) L \rfloor}_t - \xx_t^{\lfloor \lambda L \rfloor} \right\| = \Omega(\epsilon^{1-\kappa}),\]
where $\Omega(1)$ should be interpreted in the limit ``$n \to \infty$, then $L \to \infty$, then $\epsilon \to 0$''. We say a parametrization is \emph{redundant} if $\kappa=0$.
\end{defn}
In other words, the feature diversity exponent $\kappa$ is a measure of how different the outputs are in layers that are close to each other. With $\kappa=0$, the output of each layer is essentially the same as the output of the previous layer in the sense that the rate of change from one layer to the next is bounded (at least locally), and hence the network is intuitively ``wasting'' parameters.

\begin{claim}\label{clm:redundant}
    Consider a stable and nontrivial parametrization that is furthermore faithful during training (but not necessarily at initialization).
    Then it is redundant if $\alpha \in (1/2, 1]$.
\end{claim}
To understand the intuition behind \Cref{clm:redundant}, let us see what happens when $\alpha > 1/2$. In this case, the randomness of the initialization weights will have no impact on training trajectory as depth increases. To see this, consider some layer index $\lfloor \lambda L\rfloor$. The blocks are divided by $L^{\alpha}$ which is larger than the magnitude of accumulated randomness (of order $(\lambda L)^{1/2}$). This basically destroys all the randomness from initialization and therefore the randomness in the learned features will consist only of that coming from $U$ and $V$ (input and output matrices). When depth goes to infinity, the contribution of the randomness in two adjacent layers becomes less important, we end up with adjacent layers becoming very similar because the gradients to these layers are highly correlated. 

In contrast, we have the following result, which defines Depth-$\mu$P.

\begin{claim}[Depth-$\mu$P]\label{clm:depth_mup}
    $\alpha = \gamma = 1/2$ is the unique parametrization that is stable, nontrivial, faithful, induces feature learning, and achieves maximal feature diversity with $\kappa =1/2$.
\end{claim}

In terms of feature diversity, a phase transition phenomenon occurs when $\alpha = 1/2$. More precisely, for Depth-$\mu$P, we can show that $n^{-1/2} \left\| {\xx^{\lfloor (\lambda + \epsilon) L \rfloor}_t - \xx_t^{\lfloor \lambda L \rfloor}} \right\| = \cO(\epsilon^{1/2})$ while the same quantity is $\cO(\epsilon)$ for all $\alpha \in (1/2,1]$, which suggests that Depth-$\mu$P yields \emph{rough} path for $\xx_t$. This allows the features to change significantly from one layer to the next, hence efficiently using the parameters. For readers who are familiar with rough path theory, the $1/2$ continuity exponent is a result of Brownian increments in the path.\footnote{The reader might ask whether we can obtain an exponent smaller than $1/2$. This is indeed possible, but it will entail using correlated weights. We leave this question for future work.}

Moreover, with $\alpha = 1$, there is a phenomenon of feature collapse in the sense that the features will be contained in the $\sigma$-algebra generated by the input and output layers, but contains no randomness from the hidden layers (see \Cref{heuristics:inv-depth}). Intuitively, the case of $\alpha=1$ is analogous to width situation, where deep mean field collapses to a
single neuron (all neurons become essentially the same). For depth, the features (layers) are still relatively different but the redundancy does not allow significant variety in these features.

\subsection{Sublety: Layerwise (local) linearization but not global linearization}

\begin{defn}\label{defn:layerwise-linearization}
    We say a parametrization induces layerwise linearization iff each layer can be linearized without changing the network output when $L\to \infty$, that is, $\forall l\in [L]$, 
    \[L^{-\alpha}G\left(\phi(W^l_t\xx_t^{l-1})-\phi( W^l_0\xx_t^{l-1}) - \phi'(W_0^l\xx_t^{l-1})\odot ((W_t^l-W_0^l)\xx_t^{l-1})\right) = o(L^{-1})\]
\end{defn}

\begin{claim}\label{clm:layerwise-linearization}
    A stable and nontrivial parametrization induces layerwise linearization iff $\alpha \in [1/2, 1)$.
\end{claim}

However, note that this does not imply the entire network is linearized (w.r.t. all the parameters in the sense of Neural Tangent Kernel).
In our setup, where the input and output layers are initialized at a constant scale (w.r.t. $L$), it is actually not possible to have a kernel limit. Even in our linear case in \Cref{sec:linearsgd}, one can see the learned model is not linear. 

If the initialization of the output layer is $L$ times larger than our setup (assuming $L\ll n$ so the widthwise scaling still follows $\mu$P), it may induce a parametrization that can linearize the entire network. In that situation, the learning rate has to be $L$ times smaller than Depth-$\mu$P to obtain stability during training, so the change of parameters is also $L$ times smaller, which can lead to the linearization of the entire network. Since we focus on maximal feature learning, the rigorous argument is beyond the scope of this paper.

\section{Feature Diversity}
In this section, we show that the choice of nonlinearity and placement of nonlinearities can affect feature diversity greatly.

\subsection{Gradient Diversity}
\emph{Gradient diversity} is an important factor toward feature diversity. Observe that the gradient $\del x^l$ at $x^l$  is continuous in $l$ in the limit $L \to \infty$.  In a linear model (or the pre-nonlin model, where nonlinearity is put before the weights), this causes $\del h^l = L^{-\alpha}\del x^l$ to be very similar between neighboring blocks. As a result (because the weights $W^l$  receives an update proportional to $\del h^l \otimes x^{l-1}$), in the next forward pass, neighboring blocks contribute very similarly to the  main branch $x^l$. This leads to a waste of model capacity.

\subsection{Pre-Nonlin Leads to Poor Performance}
\begin{figure}
    \centering
    \includegraphics[width=0.4\linewidth]{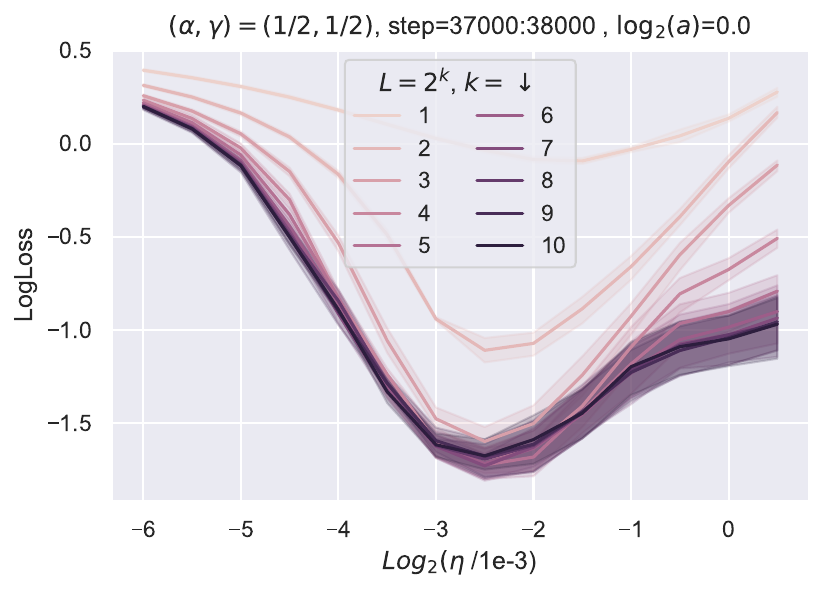}    \includegraphics[width=0.4\linewidth]{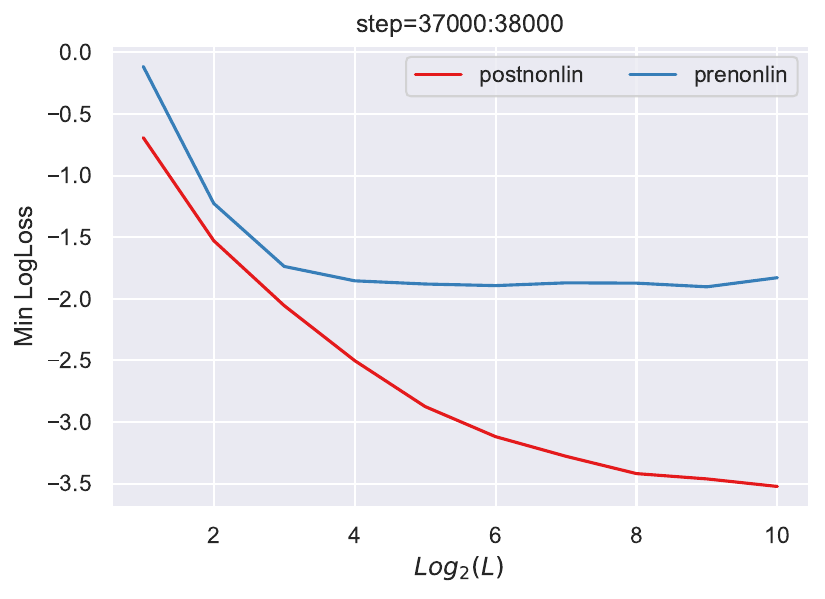}
    \caption{\textbf{Pre-Nonlin Leads to Poor Performance} Although Depth-$\mu$P for prenonlin resnet indeed transfers hyperparameters (Left), depth gives no performance gains beyond 8 layers and the performance is dramatically worse than the post-nonlin resnet (Right). In right plot, the "Min LogLoss" is minimal log loss over all block multiplier and learning rate.  Networks are trained on CIFAR-10 with Adam. See \Cref{fig:lr_transfer} for more details about the setup.}
    \label{fig:relu-bd1-ln0-mup-prenonlin-bm0-sweep-lr-seed}
\end{figure}

For example, in \Cref{fig:relu-bd1-ln0-mup-prenonlin-bm0-sweep-lr-seed}, for a relu pre-nonlin resnet (i.e. blocks are given by $W^l\phi(x^{l-1})$ instead of $\phi(W^l x^{l-1})$), we see that although Depth-$\mu$P indeed transfers hyperparameters (as predicted by our theory), the performance is dramatically worse than the post-nonlin resnet in \Cref{fig:lr_transfer}, and depth gives no performance gains beyond 8 layers. Specifically, it is because $\del h^l=L^{-\alpha}\del x^l$ like the linear case, and $\phi(x^{l-1})$ is also similar between neighboring blocks. As a result, the gradient of the weights $W^{l}$, proportional to $\del h^l\otimes \phi(x^{l-1})$, has little diversity compared to nearby blocks.

\subsection{Maximizing Feature Diversity with Absolute Value Nonlinearity}

In a nonlinear model,  we have $\delta h^l = \delta x^l \odot \phi'(h^l)$.  Because $h^l$ is almost independent from all other $h^m, m \ne l$ in the Depth-$\mu$P limit, $\phi'(h^l)$ can serve to decorrelate the $\delta h^l$, depending on what $\phi$  is.
For example, if $\phi$ is relu, then $\phi'$ is the step function. $h^l$ is approximately a zero-mean Gaussian in the Depth $\mu$P limit, so that $\phi'(h^l)$ is approximately 0 or 1 with half probability each.
This decorrelates $\delta h^l$ much better than the linear case.
But of course, this line of reasoning naturally leads to the conclusion that $\phi' = \mathrm{sign}$ would be the best decorrelator of $\delta h^l$ and the maximizer of feature diversity (with $\phi$ among the class of positively 1-homogeneous functions) --- then $\delta h^l$ and $\delta h^m$ are completely decorrelated for $l \ne m$.

\begin{figure}
    \centering
    \includegraphics[width=0.4\linewidth]{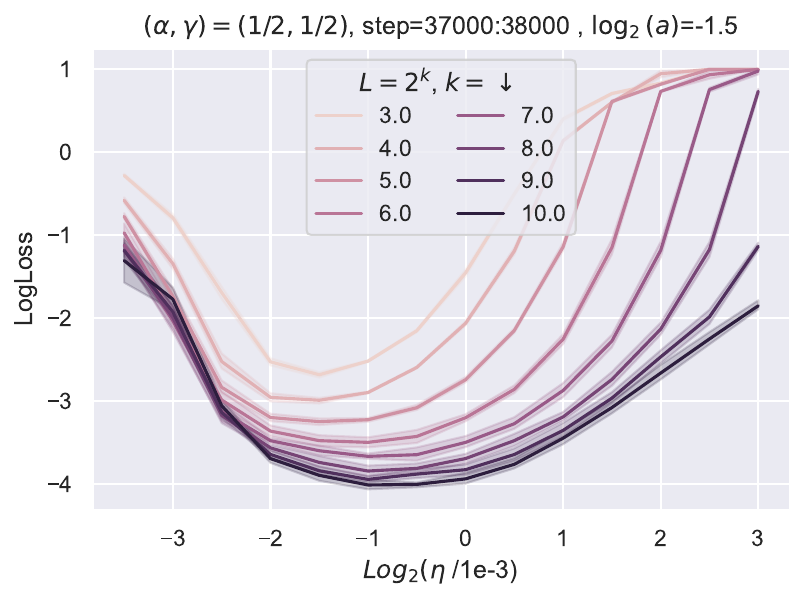}
    \includegraphics[width=0.4\linewidth]{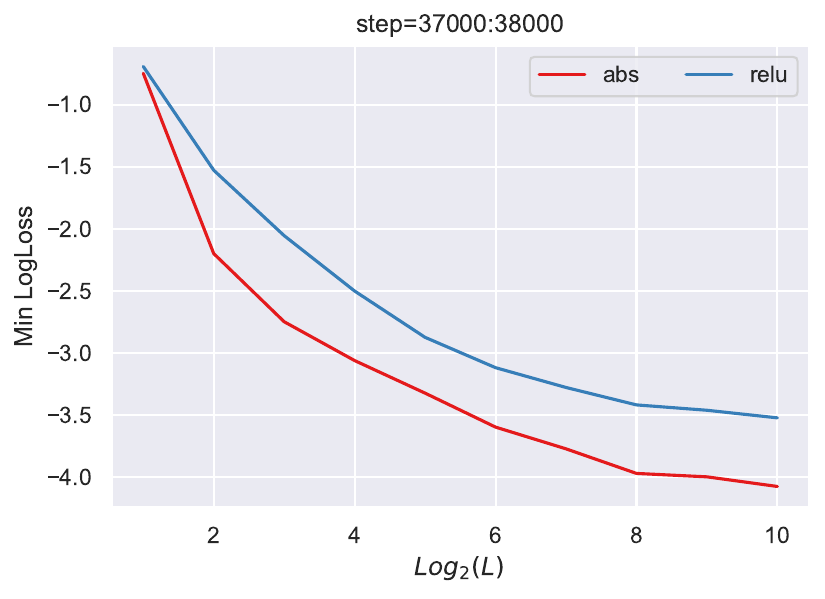}

    \caption{\textbf{Improving performance with absolute value non-linearity}, which maximizes feature diversity. (Networks are trained on CIFAR-10 with Adam.). See \Cref{fig:lr_transfer} for more details about the setup.}
    \label{fig:abs-bd1-ln0-mup-bmneg1p5-sweep-lr-seed}
\end{figure}

Indeed, as shown in \Cref{fig:abs-bd1-ln0-mup-bmneg1p5-sweep-lr-seed}, swapping in absolute value for $\phi$ dramatically improves the training performance of deep (block depth 1) resnets.

In general, in lieu of absolute value,  any even nonlinearity would suffice.

\subsection{Feature Diversity is in Tension with Layerwise Linearization}

The reason that $\phi'(h^l)$ can decorrelate  $\delta h^l$ is very much related to layerwise linearization.
Recall that in Depth-$\mu$P, $h^l$ can be decomposed to a zero-mean Gaussian part $\hat h^l$ of size $\Theta(1)$ and a correction term $\dot h^l$  of size $\Theta(L^{-1/2})$ (corresponding to the decomposition $\ket{h^l} = \hatket{h^l} + \dotket{h^l}$).
$\hat h^l$ is independent from $\hat h^{m}$ for $m \ne l$ but $\dot h^l$ can be very strongly correlated to all other $\dot h^m$.
Thus, $\phi'(h^l)$ can decorrelate $\delta h^l$ precisely because $\hat h^l$ dominates $\dot h^l$, and this is also precisely the reason we have layerwise linearization.

In the $1/L$ scaling $(\alpha, \gamma) = (1, 0)$, $\hat h^l$ is on the same order as $\dot h^l$ and layerwise linearization does not occur, but also $\phi'(h^l)$ can no longer effectively decorrelated $\delta h^l$.

Once again, we remind the reader that layerwise linearization in this case is not detrimental (in this block depth 1 case) because $\hat h^l$ in fact accumulate contributions from the learned features of all previous blocks and thus strongly depends on the learning trajectory (in contrast to the (widthwise) NTK case where $\hat h^l$ is already determined at initialization).

\clearpage{}%
\section{Block Depth 2 and Above}

\label{sec:bd2}

\emph{Remark on notation:} Here and in the next section, all big-O notation is in $L$ only; the scaling in width is assumed to be in $\mu$P.

In most of this work, we have considered depth-1 MLP for $g^l$ in \cref{eqn:resmlp_defn}, it's straightforward to derive and classify the infinite-width-then-infinite-depth limits for larger depths in each block.
In particular, the following $1/\sqrt{L}$ scaling still makes sense in this more general setting with block depth $k$ and leads to a well defined limit:

    \begin{align}
        x^{l}&=x^{l-1} +\frac a {\sqrt L} \cdot g^l(x^{l-1}; W^{l1}, \ldots, W^{lk}),\quad\text{$\Theta(1)$ initialization scale,}\quad\text{$\Theta(1/\sqrt L)$ learning rate}
    \end{align}
This is what we call Depth-$\mu$P in the block depth 1 case, but we shall not use this name in the general block depth case because \emph{this parametrization is no longer optimal}.\footnote{What we exactly mean by \emph{optimal} will be explained below.}

\subsection{Block Depth $\ge 2$ is Defective}

\begin{figure}
    \centering
    \includegraphics[width=0.4\linewidth]{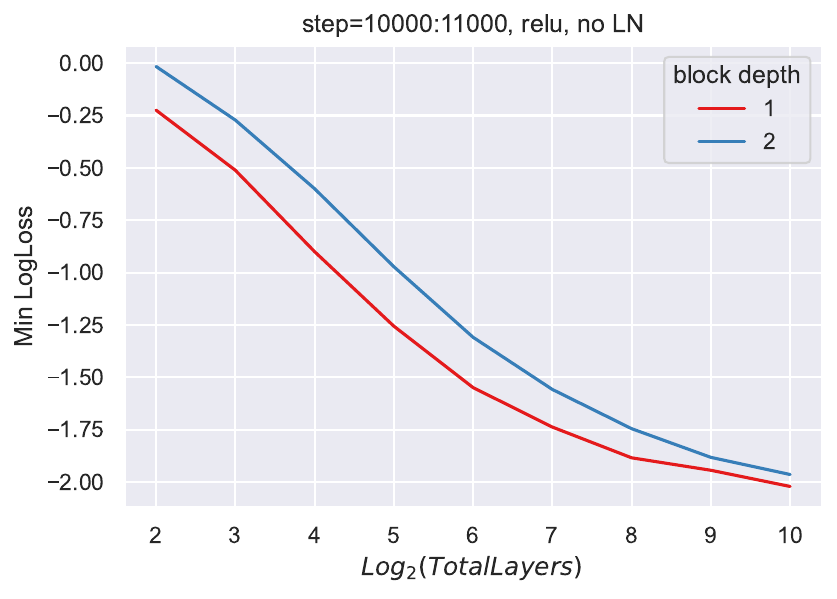}
    \includegraphics[width=0.4\linewidth]{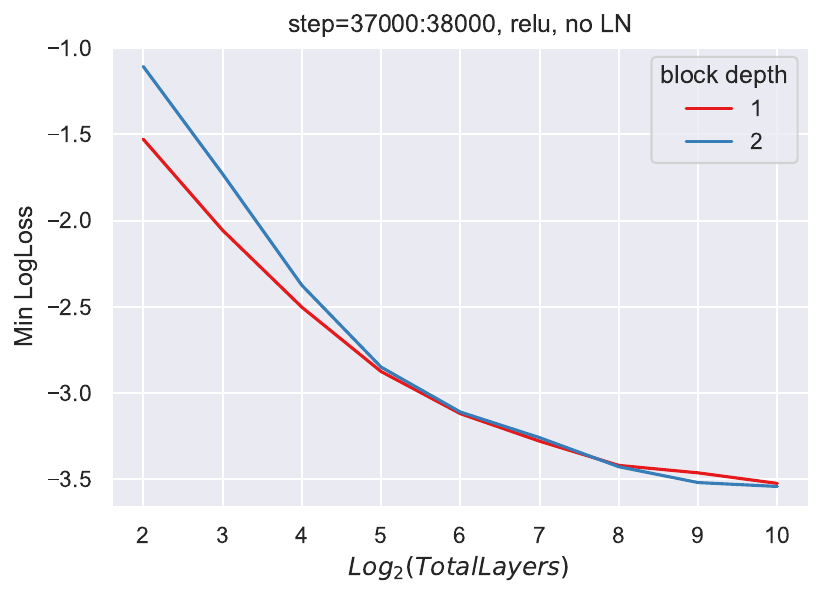}
    \caption{\textbf{Block Depth 2 < Block Depth 1, Relu}. In relu resnet with no LN, block depth 2 does worse than block depth 1 when matching total number of layers (and thus parameter count). However, training longer (38000 steps, Right) helps it catch up (compared to 11000 steps, Left). The y-axis is minimal log loss over all block multiplier and learning rate}
    \label{fig:relu-ln0-compare-bd}
\end{figure}

\begin{figure}
    \centering
    \includegraphics[width=0.4\linewidth]{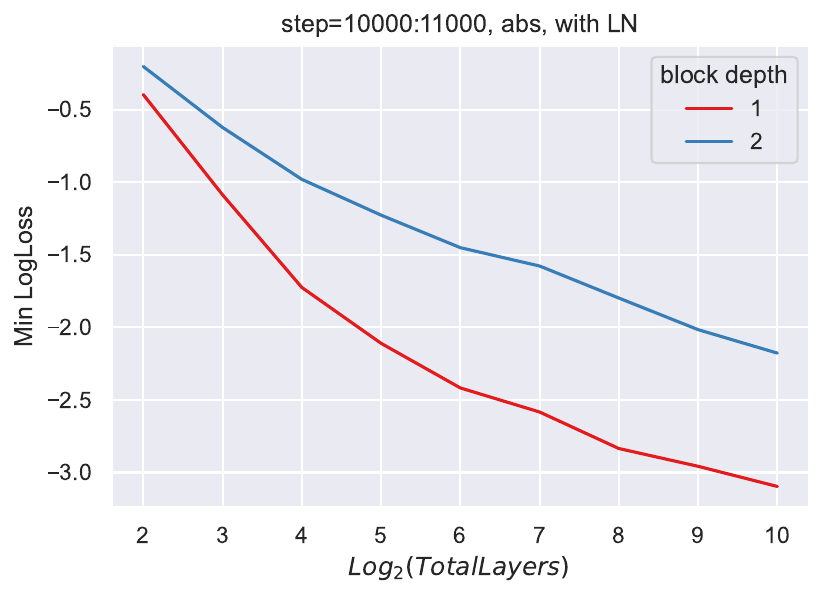}
    \includegraphics[width=0.4\linewidth]{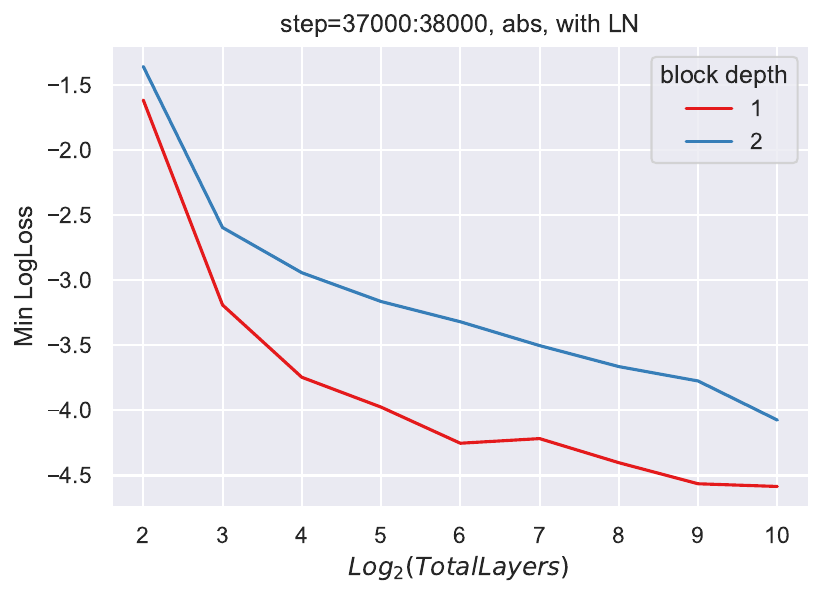}
  \caption{\textbf{Block Depth 2 < Block Depth 1, Abs}. In abs resnet with LN, block depth 2 does significantly worse than block depth 1 when matching total number of layers (and thus parameter count). Training longer (38000 steps, Right) does not close the performance gap (compared to 11000 steps, Left). The y-axis  is minimal log loss over all block multiplier and learning rate}
    \label{fig:abs-ln1-bd1-vd-bd2}
\end{figure}

\begin{figure}
    \centering
    \includegraphics[width=0.4\linewidth]{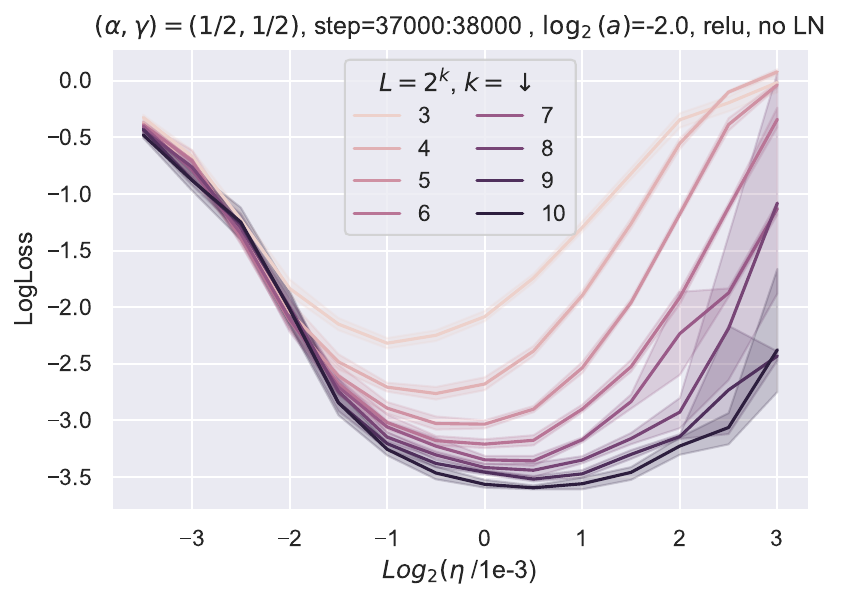}
    \includegraphics[width=0.4\linewidth]{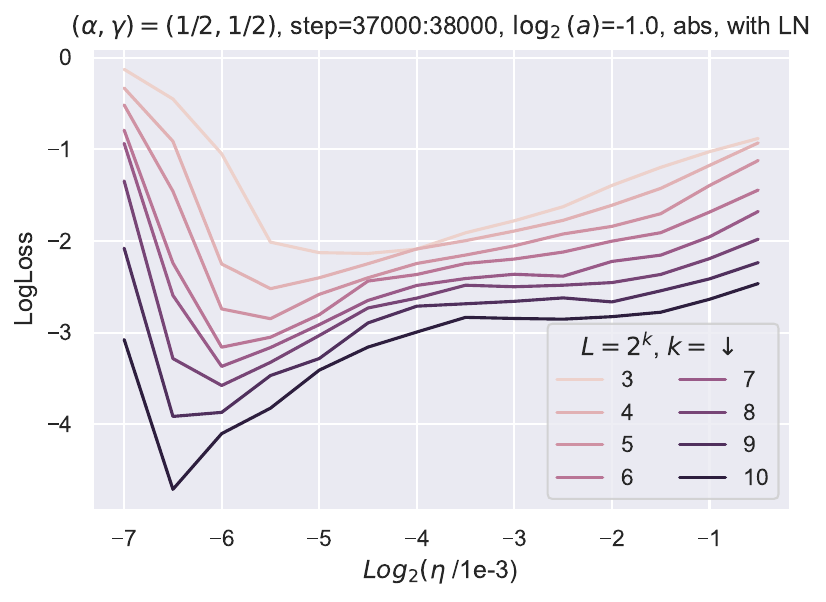}
    \caption{\textbf{Block Depth 2 Hyperparameter Shift} in relu resnet with no LN (Left) and abs resnet with LN (Right).}
    \label{fig:bd2-hp-shift}
\end{figure}

A very clear symptom of this is that the \emph{performance of block-depth-2 resnets is worse than that of block-depth-1 networks}, when matching parameter count, although they can (but not always) catch up after training for a long time (\cref{fig:relu-ln0-compare-bd,fig:abs-ln1-bd1-vd-bd2}).
Simultaneously, we are seeing nontrivial or even significant hyperparameter shifts as the total number of blocks increases (\cref{fig:bd2-hp-shift}).

\subsection{Defect of $1/\sqrt L$  Scaling in Block Depth 2}

The reason that the $1/\sqrt L$ scaling is no longer fine in the block depth $\ge 2$ case is the \emph{linearization of the multiplicative interaction} between the layers in the block.
Indeed, just like the block depth 1 case, the $1/\sqrt L$ scaling forces the weight updates $\Delta W$ of each weight matrix to be $\Theta(\sqrt{L})$ smaller than the initialization $W_0$.
Thus, within the block, the training dynamics when depth $L$ is large is in the kernel regime, where the contribution to the block output $g(x; W^\bullet)$ is only a \emph{summation}, instead of \emph{product}, of individual contributions from each layer's weights updates.

When aggregated over all $L$ blocks, the result is that there is only multiplicative interaction of $\Delta W$ across blocks but not within layers.
In other words, the network output is dominated, for example in the linear case, by the contributions of the form
$M^L \cdots M^1$
where each $M^l$ can be one of $I, W_0^{l2} W_0^{l1}, W_0^{l2} \Delta W^{l1},$ or $\Delta W^{l2} W_0^{l1}$, but NOT $\Delta W^{l2} \Delta W^{l1}$.
All other contributions (which all involve within-block interactions like $\Delta W^{l2} \Delta W^{l1}$) are subleading. 
In the general nonlinear case, replacing the block 
\begin{equation*}
    \phi(W^{l2}\phi(W^{l1} x^{l-1}))    
\end{equation*}
with the linearized version 
\begin{equation*}
    \phi(h^l_\wedge) + \phi'(h^l_\wedge) \odot [\Delta W^{l2}\phi(h^l_\vee)] + \phi'(h^l_\wedge) \odot [W_0^{l2} (\phi'(h^l_\vee) \odot [\Delta W^{l1} x^{l-1}])]    
\end{equation*}
will achieve the same performance as depth $L \to \infty$,
where $h^l_\wedge = W_0^{l2}\phi(h^l_\vee)$ and $h^l_\vee = W_0^{l1} x^{l-1}$.

When block depth $k=1$ (our main subject of study in this work), \emph{all} interactions are included but this is no longer true when $k>1$.

\begin{figure}
    \centering
    \includegraphics[width=0.9\linewidth]{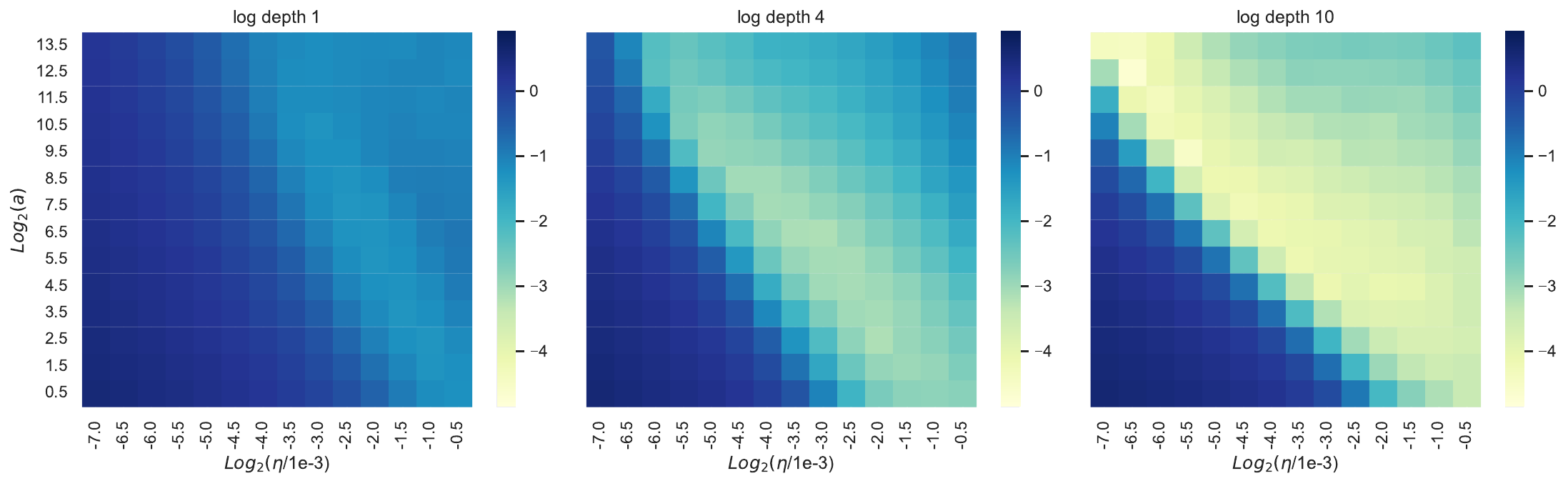}
    \caption{The "slope" of the optimal sublevel set in the (learning rate, block multiplier) space changes from $-2$ to $-1$ as depth goes from $2^1$ to $2^{10}$. Here we use absolute value nonlinearity with layer normalization, block depth 2, and networks are trained for 50 epochs with Adam on CIFAR-10.} 
    \label{fig:bd2_shifting_slope}
\end{figure}

In \cref{fig:bd2_shifting_slope}, the heatmap of loss as a function of block multiplier and learning rate demonstrates this vividly for block depth 2.

\paragraph{Small depth}
The optimal sublevel set of (learning rate, block multiplier) has slope $\approx -2$ when the number of blocks is $2^1$. In other words, around the optimum, double the learning rate while dividing the block multiplier by 4 has similar performance. This is because $\Delta W^{l1}$ and $\Delta W^{l2}$ interact \emph{multiplicatively}, so that doubling their sizes leads to quadrupling their contribution to the block output. The simultaneous decrease of block multiplier by 4 then roughly keep their contribution invariant in size.

\paragraph{Large depth}
On the other hand, the optimal sublevel set has slope $\approx -1$ when the depth is $2^{10}$: Doubling the learning rate while halving the block multiplier has similar performance. This reflects the fact that  $\Delta W^{l1}$ and $\Delta W^{l2}$ now interact \emph{additively}. 

Intermediate depths interpolate this phenomenon, as seen in the plot for depth $2^5$.

In the same heatmaps, one can see the optimal (learning rate, block multiplier) (in the $1/\sqrt L$ parametrization) shifts from the middle of the grid to the upper left as depth goes from $2^5$ to $2^{10}$, demonstrating the lack of hyperparameter transfer.

This change in slope is seen in relu networks as well, with or without layernorm.

Finally, we note that the $1/\sqrt{L}$ scaling still yields a $L \to \infty$ limit where the network still learns features as a whole, even though within each block this is no longer true.
Thus, this is another reminder that mere "feature learning" does not imply "hyperparameter transfer"!

\subsection{Classification of Parametrizations}

These heatmaps already demonstrate that no parametrization of (global learning rate\footnote{meaning, the learning tied across all layers in a block}, block multiplier) can transfer hyperparameters robustly, because any such parametrization can only \emph{shift} the heatmaps but not \emph{stretch} them, so one cannot "transfer" a sublevel set of one slope into a sublevel set of another slope.

But even if we allow learning rate to vary between layers in a block, no stable, faithful, nontrivial parametrization can avoid the linearization problem described above.

For simplicity, fix a positive-homogeneous nonlinearity and block depth 2.\footnote{but our arguments generalize trivially to arbitrary block depth $\ge 2$}
We consider the space of hyperparameters consisting of the learning rate for each of the layers in a block, as well as the block multiplier (one for each block); WLOG all weights are initialized $\Theta(1)$.\footnote{This is WLOG because the nonlinearities are homogeneous} This yields a space of dimension $\mathrm{blockdepth} + 1 = 3$.

Indeed, for this to happen, the weight update $\Delta W^{li}$ must be at least of order $\Omega(1)$ (size of initialization) for some $i$.  But this would contribute a drift term to the block output $g^l = g^l(x^{l-1}; W^{\bullet})$ that is as large as the noise term.
This then implies that either the parametrization is unstable (if the block multiplier $L^{-\alpha}$ is  $\Omega(1/{L})$) or lacks feature diversity (if the block multiplier $L^{-\alpha}$ is $O(1/L)$).

For example, in a linear model, $$L^{\alpha}\ket{g^l} = \ket{W^{l2} W^{l1} x^{l-1}} = \hatket{W_0^{l2} W^{l1} x^{l-1}} + \dotket{W_0^{l2} W^{l1} x^{l-1}} + \ket{\Delta W^{l2} W^{l1} x^{l-1}}.$$ 
$\hatket{W_0^{l2} W^{l1} x^{l-1}}$ is independent and zero-mean across $l$ (the noise term), while $\dotket{W_0^{l2} W^{l1} x^{l-1}} + \ket{\Delta W^{l2} W^{l1} x^{l-1}}$ is correlated across $l$ (the drift term).
$\hatket{W_0^{l2} W^{l1} x^{l-1}}$ is always $\Theta(1)$ because the $W^{l2}_0, W^{l1}_0$ are.
If $\Delta W^{l2}$ is $\Omega(1)$, then $\ket{\Delta W^{l2} W^{l1} x^{l-1}} = \Omega(1)$ as well, making the drift term as large as the noise term.
If $\Delta W^{l1}$ is $\Omega(1)$, then $\dotket{W_0^{l2} \Delta W^{l1} x^{l-1}} = \Omega(1)$, causing $\dotket{W_0^{l2} W^{l1} x^{l-1}} = \dotket{W_0^{l2} W_0^{l1} x^{l-1}} + \dotket{W_0^{l2} \Delta W^{l1} x^{l-1}}$ to be $\Omega(1)$.\footnote{One can also observe that if $\Delta W^{l1} = \Omega(1)$, then by symmetry the backward pass suffers the same problem. But for general block depth, this argument does not say anything about the middle layers, while the argument presented above implies that $\Delta W^{li}$ cannot be $\Omega(1)$ for any $i$.}

The same argument can be straightforwardly adapted to nonlinear MLPs (with mean subtraction) and arbitrary block depth $\ge 2$, and as well to general nonlinearities that are not necessarily positive-homogeneous, with hyperparameter space enlarged to include initialization.

\subsection{So What is the Optimal Parametrization?}

All of the above considerations suggest that \emph{we are missing crucial hyperparameters in our consideration} when increasing the complexity of each block.
Our study right now is akin to the naive study of the 1-dimensional hyperparameter space of the global learning rate in SP.
Discovering these missing hyperparameters will be an important question for future work.
\clearpage{}%

\begin{figure}
    \centering
    \includegraphics[width=\textwidth]{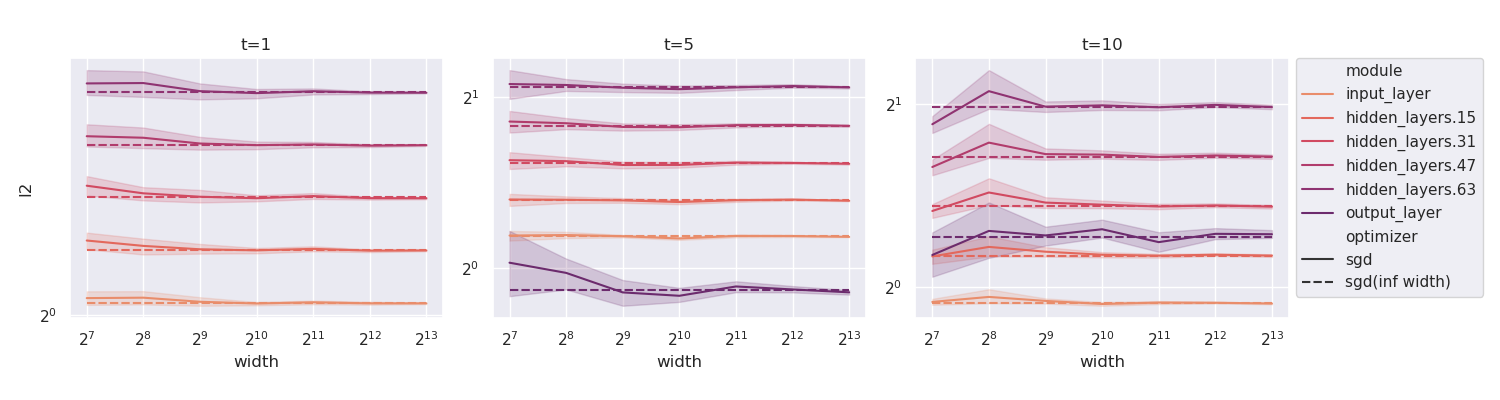}
    \caption{Trained linear network  converges to its infinite width limit which is computed recursively based on $\Gamma$ and $C$. Depth is fixed at 64,width varies between $2^7, 2^8, \ldots, 2^{13}$. Networks are trained with SGD for 10 steps. The root mean square statistics ($y$-axis) at 1st, 5th and 10th steps are plotted using solid lines where the $x$-axis is the width. The root mean square values are computed on the outputs of some of the layers (including the input layer, output layer, and hidden layers at each quarter). The corresponding value for the infinite width is indicated with dashed lines.  }
    \label{fig:verify-linear-finite-d64}
\end{figure}

\begin{figure}
    \centering
    \includegraphics[width=\textwidth]{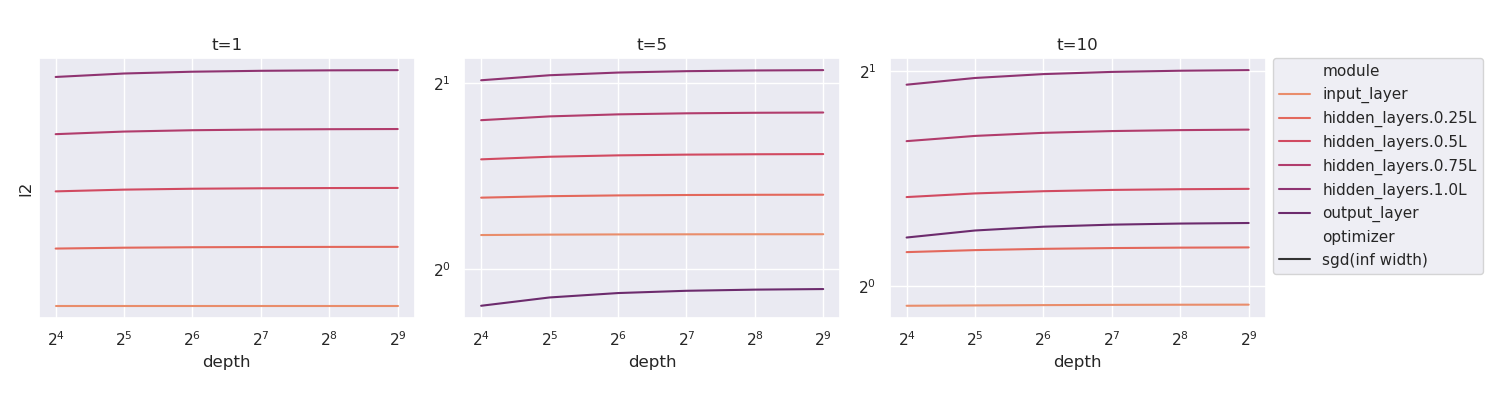}
    \caption{Under Depth-$\mu$P, infinite wide linear network training converges when increasing the depth. Infinite wide linear networks of depth $2^4, 2^5, \ldots, 2^{9}$  are computed recursively based on $\Gamma$ and $C$. The root mean square statistics ($y$-axis) at 1st, 5th and 10th steps are plotted across the depth ($x$-axis). }
    \label{fig:verify-linear-infinite}
\end{figure}

\section{Experiments}\label{sec:exp}

\subsection{Verifying the Theory in the Linear Case}\label{sec:verify_linear}

In \Cref{sec:linearsgd}, we showed that a complete description of the training dynamics of linear networks can be formulated in terms of  $\Gamma$ and $C$. In this section, we provide empirical results supporting our theoretical findings. We first verify the finite-depth recursive formula for $\Gamma$ in \Cref{lemma:finite_depth_gamma} is the correct limit when the width goes to infinity, then proceed to show that the infinite-depth limit is the correct one.

\paragraph{Infinite-width limit.} In \Cref{fig:verify-linear-finite-d64}, we train a series of $64$-layer linear networks of width $2^7, 2^8, \ldots, 2^{13}$ with $1, 5, 10$ steps on MNIST, and plot the root mean square\footnote{The root mean square of a vector $x=(x_1,\ldots, x_n)$ is $\sqrt{\frac{\sum_{i=1}^n x_i^2}{n}}$, which is denoted as ``l2'' in \Cref{fig:verify-linear-finite-d64,fig:verify-linear-infinite}.} of the layer outputs using solid lines. We also compute the infinite width limit of the corresponding statistics using the recursive formula for $\Gamma$ and plot them as dashed horizontal lines. For clarity of the figure, we only plot the statistics of the input layer, output layer, and hidden layers of index 16, 32, 48, and 64. It is clear that as the width grows, the solid lines converge to the dashed lines consistently across the training steps. It indicates that our computation of the infinite width limit is correct.

\paragraph{Infinite-depth limit.} We verify that the infinite \emph{width} limit above converges when the \emph{depth} grows. We consider linear networks of the same architecture but vary the depth from $2^4$ to $2^9$. We again compute the root mean square values of the layer outputs using the recursive formula for $\Gamma$, and plot them in \Cref{fig:verify-linear-infinite} with depth being $x$-axis. For clarity of the figure, we only plot the statistics of the input layer, output layer, and hidden layers of index $L/4$, $L/2$, $3L/4$, and $L$. One can observe that the statistics of the layer outputs converge quickly when the depth grows from $2^4$ to $2^9$, which verifies our convergence result.

\begin{figure}[h]
    \centering
    \includegraphics[width=0.45\linewidth]{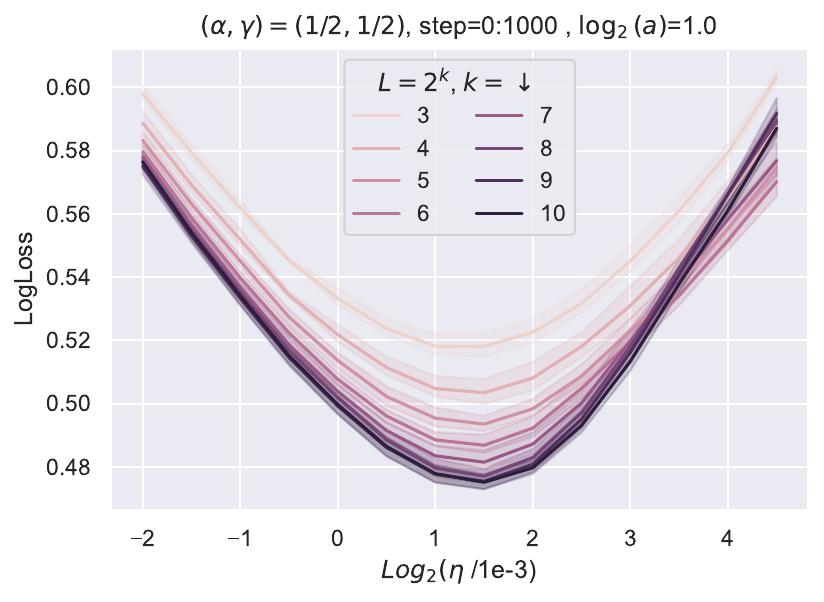}
    \includegraphics[width=0.45\linewidth]{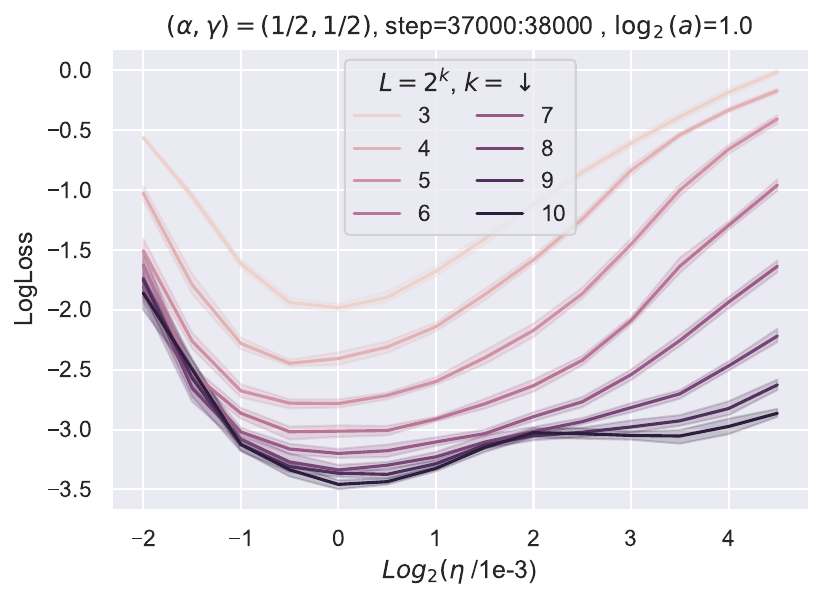}

    \centering
    \includegraphics[width=0.45\linewidth]{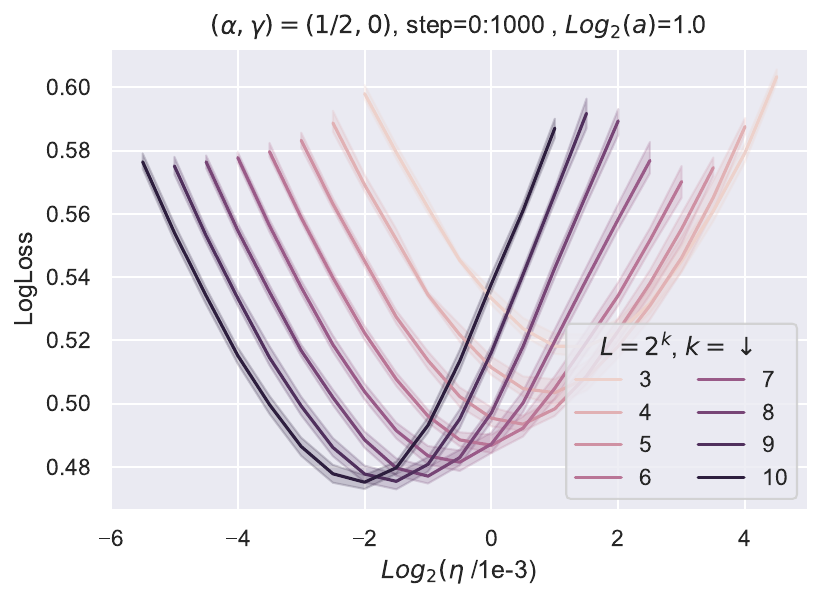}
    \includegraphics[width=0.45\linewidth]{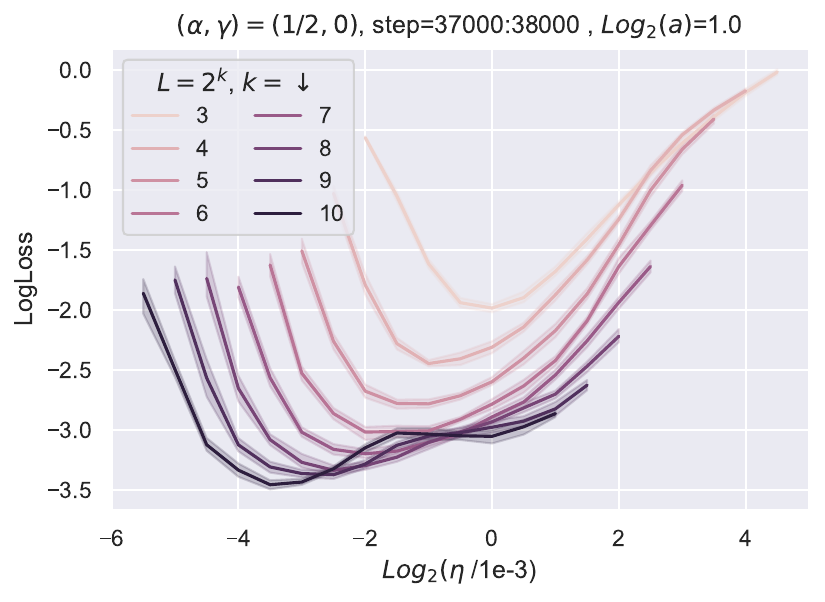}

    \includegraphics[width=0.45\linewidth]{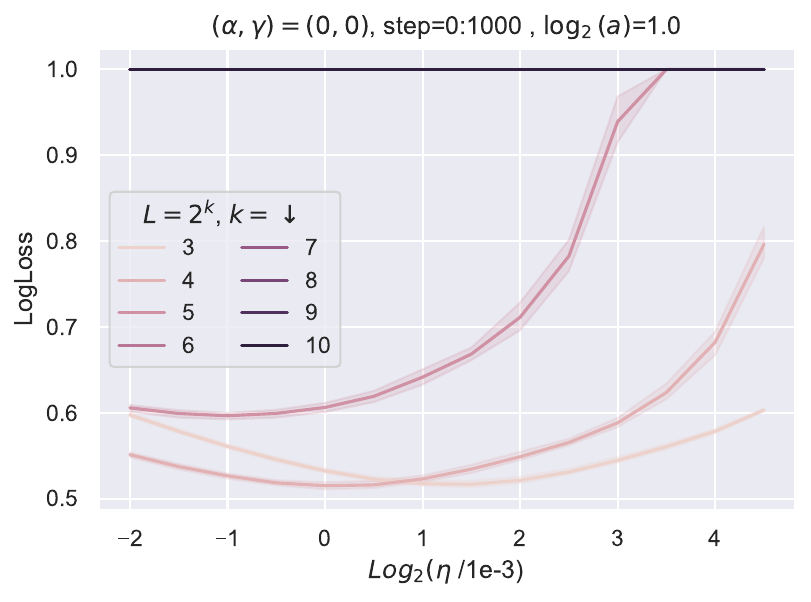}
        \includegraphics[width=0.45\linewidth]{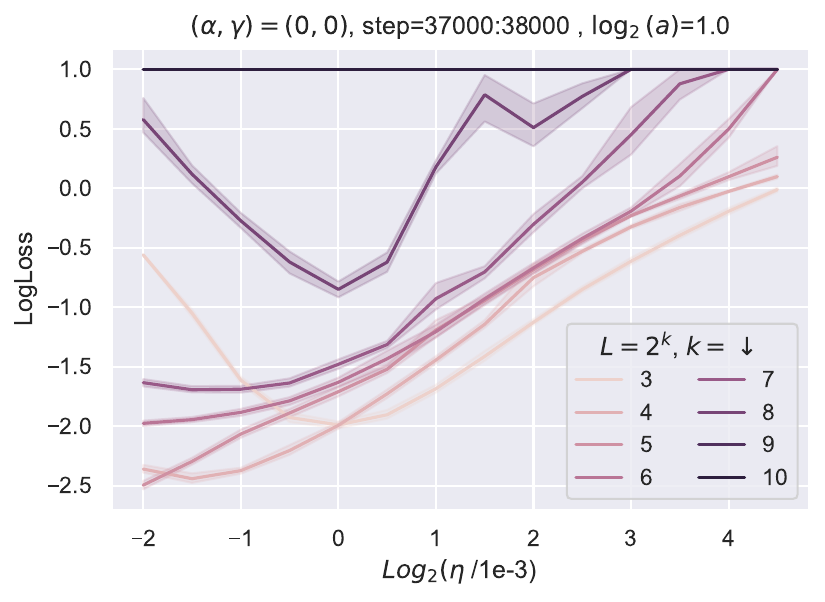}
    
    \caption{Train logloss versus learning rate for width $n=256$ and varying depths. The network consists of MLP blocks (with block depth 1), trained for 50 epochs on CIFAR10 dataset using Adam. The batch size is fixed to $64$. We tune the depth $2^3$ network to obtain the optimal $(\log_2(a), \log_2(\eta/1e-3)) = (1, 0)$, and scale all deeper networks using $2^3$ as base depth. The reader can check that the $L=2^3$ curves in each columns are the same. We show the logloss versus the learning rate of the hidden layers (input/output layers fixed) for three parametrizations: Depth-$\mu$P (\textbf{Top}), Scaling only the blocks (no LR scaling), i.e. $\gamma=0$ (\textbf{Middle}), and Standard Parametrization without any scaling ($\alpha=\gamma=0$) (\textbf{Bottom}). Each curve represents the average training loss over a time slice of 1000 steps for depths $2^k$ for $k\in \{1,2,\dots,10\}$. Confidence intervals are based on 5 seeds. The results show that Depth-$\mu$P preserves the optimal learning rate while consistently improving the training loss as depth increases. If we only scale the blocks without scaling the LR ($\alpha=1/2, \gamma=0$) when training with Adam, the optimal learning rate shifts significantly with depth. With standard parametrization without any depth scaling (common practice), the results show a significant shift in the optimal learning rate as well.
    For SP, we cap the log loss at 1, which is why for depth $2^9, 2^{10}$, we have a black horizontal line at $LogLoss=1$.
    }
    \label{fig:lr_transfer}
    \vspace{-0.5cm}
\end{figure}

\subsection{Hyperparameter Transfer}
In this section, we provide empirical evidence to show the optimality of Depth-$\mu$P scaling and the transferability of some quantities across depth. We train vanilla residual network with block depth 1 (1 MLP layer in each residual block) on CIFAR-10 dataset using Adam optimizer, batch size $64$, for $50$ epochs (input and output layers are fixed). The network is parameterized as 
follows
$$
x^l = x^{l-1}+ a \times L^{-\alpha} \textrm{MS}(\phi(W^l x^{l-1})),
$$
and the weights are trained with the rule
$$
  W^{l}\gets W^{l}-\eta \times n^{-1} L^{-\gamma}Q^l_t(n L^{\delta} g_{0},\ldots,n L^{\delta} g_{t}),
$$
where the learning rate $\eta$ and the block multiplier $a$ are the \emph{hyperparameters}.\footnote{Note that $\eta$ here is the constant, and the effective learning rate is given by $\eta n^{-1} L^{-\gamma}$.} The values of $\alpha, \gamma$ depend on the parametrization of choice. For Depth-$\mu$P, we have $\alpha=\gamma=1/2$, and for standard parametrization, we have $\alpha=0,\gamma=1$.\footnote{In standard parametrization, there is generally no rule to scale the learning rate with depth, and the optimal learning rate is typically found by grid search. Here, we assume that in standard parametrization, the learning rate is scaled by $L^{-1}$ to preserve faithfulness.} In our experiments, we assume base depth $8$, meaning that we replace $L$ by $L/8$ in the parametrization above.

\paragraph{Learning rate transfer ($\eta$). }In \Cref{fig:lr_transfer}, we show the training loss versus learning rate for depths $2^k$, for $k\in \{3,4\dots,10\}$. For Depth-$\mu$P, a convergence pattern can be observed for the optimal learning rate as depth grows. Optimal learning rates for small depths (e.g. $L=2^3$) exhibit a mild shift which should be expected, as our theory shows convergence in the large depth limit.  However, starting from depth $L=2^6$, the optimal learning rate is concentrated around $10^{-3}$. 
For parametrization that only scales the multiplier but not LR ($\alpha=1/2$, $\gamma=0$), we observe the optimal learning rate shifts significantly. 
For standard parametrization without any depth scaling  ($\alpha=\gamma=0$), the optimal learning rate exhibits a more significant shift as depth grows. Moreover, even if one picks the optimal learning rate for each depth, the performance still degrades when the depth is very large, suggesting that standard parametrization is not suitable for depth scaling. Additional figures with multiple time slices are provided in \Cref{sec:additional_exps}.

\begin{figure}[h]
    \centering
    \includegraphics[width=0.45\linewidth]{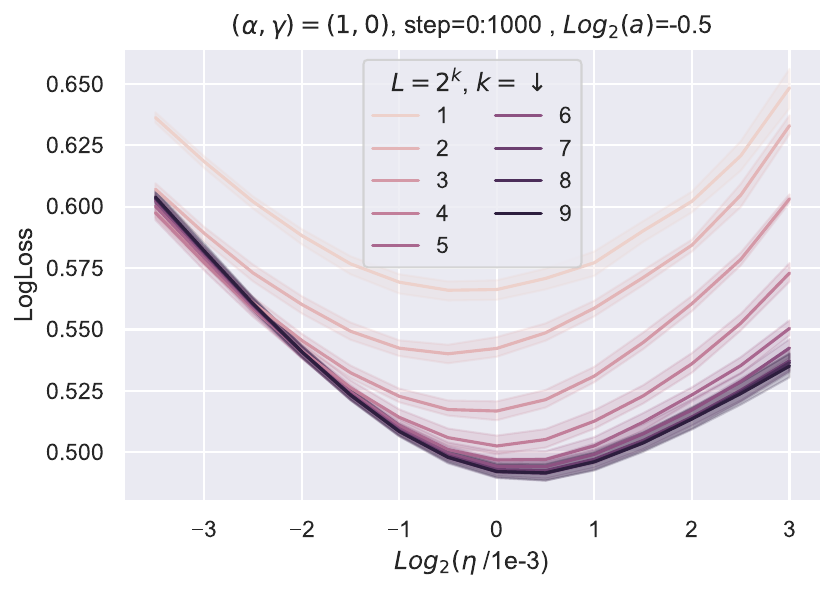}
    \includegraphics[width=0.45\linewidth]{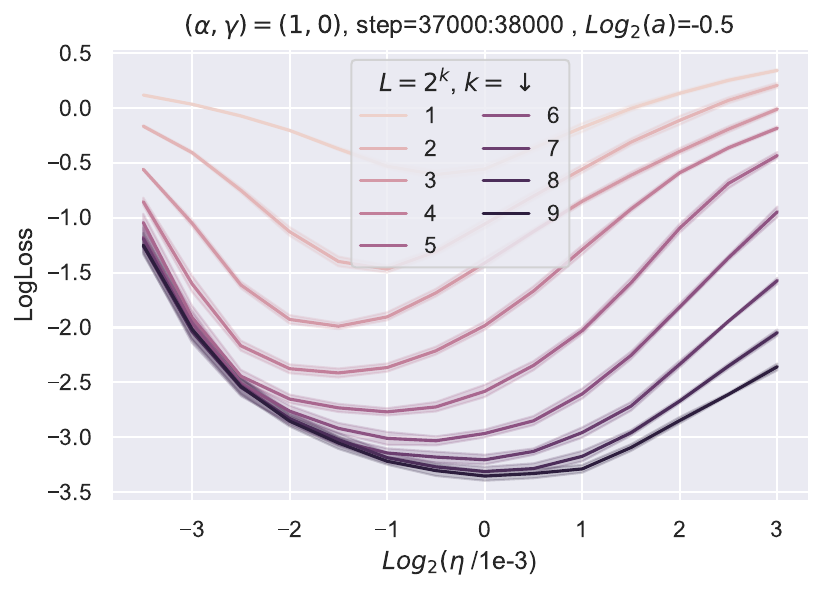}
    \caption{Same setup as \cref{fig:lr_transfer} for the parametrization $(\alpha,\gamma)=(1,0)$ (the ODE limit).}
    \label{fig:ode_limit}
\end{figure}

\paragraph{Is feature learning sufficient for HP transfer?} In \Cref{sec:cause_hp}, we explained when and why hyperparameter transfer occurs. Precisely, to obtain HP transfer, one needs to classify all feature learning limits and choose the optimal one. We introduced the notion of feature diversity and showed that Depth-$\mu$P is optimal in the sense that it maximizes feature diversity. To show that optimality is needed for HP transfer, we train a resnet with $(\alpha, \gamma)=(1,0)$ which is also a feature learning limit. \Cref{fig:ode_limit} shows that in this case the learning rate exhibits a significant shift with depth. Interestingly, the constant $\eta$ in this case seems to increase with depth, suggesting that the network is trying to break from the \emph{ODE} limit, which is sub-optimal. Note that in \Cref{fig:lr_transfer}, with Depth-$\mu$P we obtain better training loss compared to the ODE parametrization in \Cref{fig:ode_limit}.

\begin{figure}[h]
    \centering
    \includegraphics[width=0.45\linewidth]{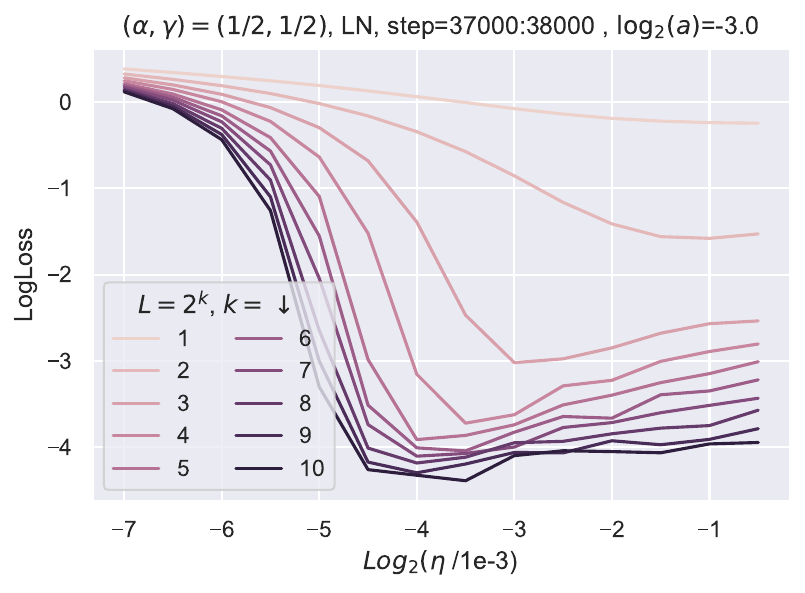}
    \includegraphics[width=0.45\linewidth]{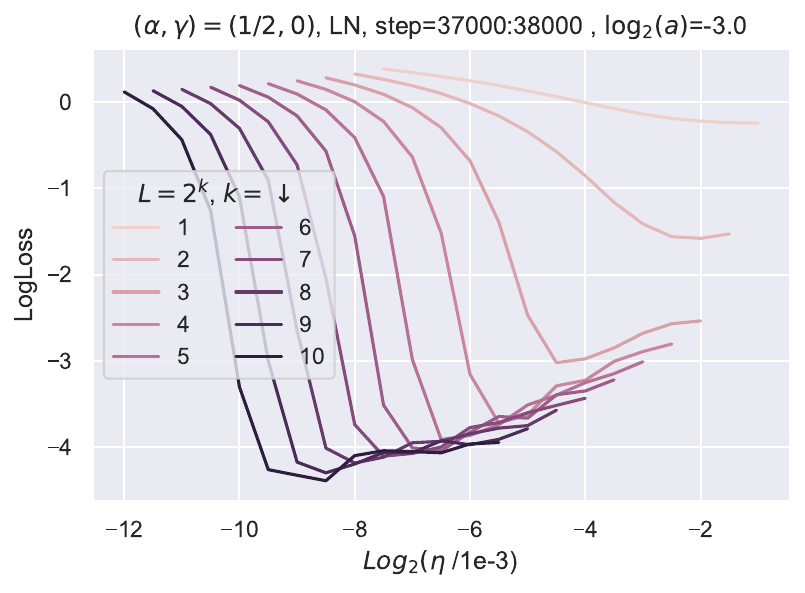}

    \caption{Same setup as \Cref{fig:lr_transfer} with Abs non-linearity instead of ReLU and LayerNorm applied to $x^{l-1}$ before matrix multiplication with $W^l$. We show the logloss versus the learning rate of the hidden layers (input/output layers fixed) for two parametrizations: Depth-$\mu$P (\textbf{Left}) and scaling only the blocks without LR scaling ($(\alpha,\gamma)=(1/2,0)$) (\textbf{Right}). The results show that Depth-$\mu$P preserves the optimal learning rate while consistently improving the training loss as depth increases. If we only scale the blocks without scaling the LR ($\alpha=1/2, \gamma=0$) when training with Adam, the optimal learning rate shifts significantly with depth. }
    \label{fig:lr_transfer_ln}
\end{figure}

\paragraph{Do we still have transfer with LayerNorm (LN)?} Our theory considers only Mean Substraction (MS), and \Cref{fig:lr_transfer} shows the results with MS. To see wether LN affects HP transfer, we train resnets with the same setup as \Cref{fig:lr_transfer} with absolute value non-linearity and LN applied to $x^{l-1}$ before matrix multiplication with $W^l$ (preLN). We keep MS after non-linearity although it can be removed since LN is applied in the next layer. Our results, reported in \Cref{fig:lr_transfer_ln} suggest that Depth-$\mu$P guarantees learning rate transfer with LN as well.

\begin{figure}[h]
    \centering
    \includegraphics[width=0.45\linewidth]{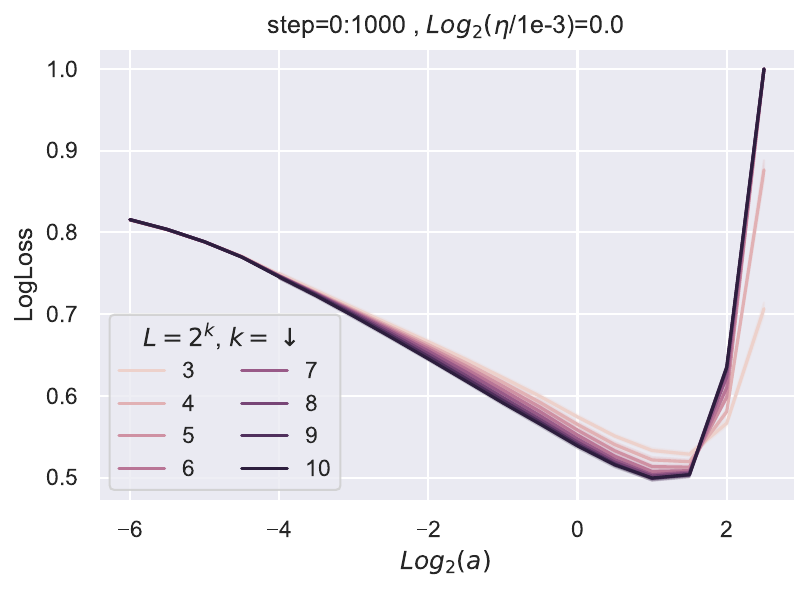}
    \includegraphics[width=0.45\linewidth]{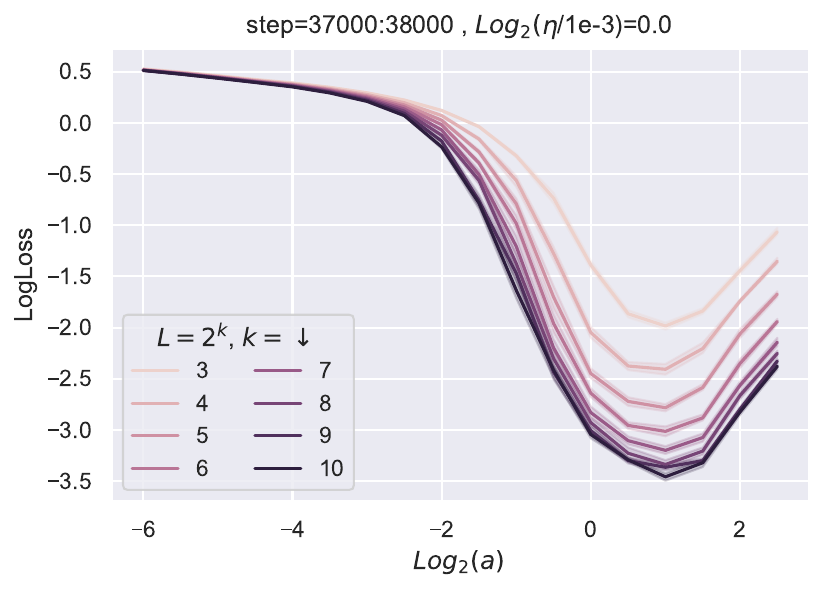}

    \caption{Train logloss versus block multiplier $a$ for varying depths. Same training setup as in \cref{fig:lr_transfer}. The results suggest that Depth-$\mu$P stabilizes the hyperparameter $a$ as depth increases.}
    \label{fig:blockmult_transfer}
\end{figure}

\paragraph{Block multiplier transfer ($a$).}
In \Cref{fig:blockmult_transfer}, we investigate the stability of the hyperparameter $a$ in Depth-$\mu$P as depth increases. The results suggest that the optimal value of this constant converges as depth grows, which suggest transferability. Additional experiments with multiple time slices are provided in \Cref{sec:additional_exps}.

\subsection{What Happens in a Transformer?}

\begin{figure}[h]
    \centering
    \includegraphics[width=0.45\linewidth]{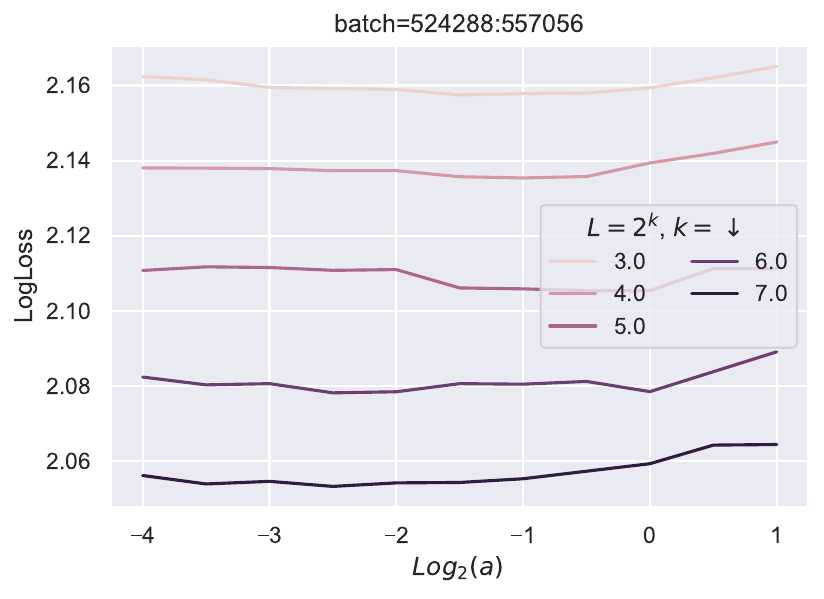}
    \caption{Modern transformers are insensitive to block multiplier $a$.}
    \label{fig:megatron_bm_insensitive}
\end{figure}

\begin{figure}[h]
    \centering
    \includegraphics[width=0.4\linewidth]{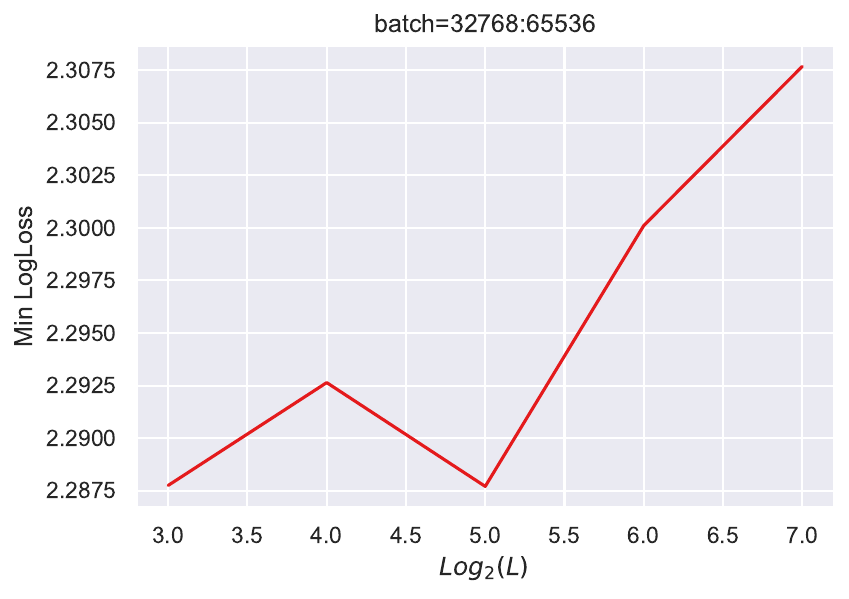}
    \includegraphics[width=0.4\linewidth]{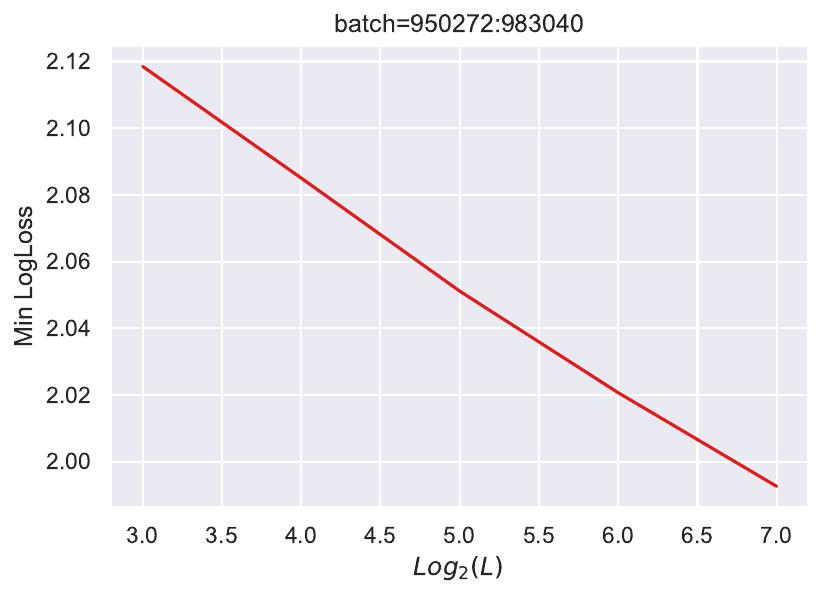}
    \caption{In (Megatron) Transformer trained on Common Crawl, deeper does worse initially (Left) but eventually does better (Right).}
    \label{fig:megatron-deeper-worse}
\end{figure}

\begin{figure}[h]
    \centering
    \includegraphics[width=0.4\linewidth]{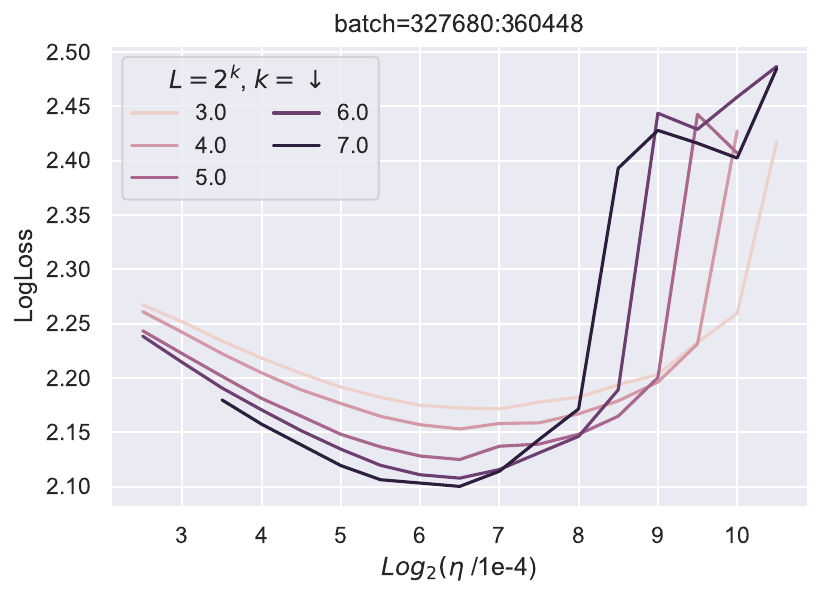}
    \includegraphics[width=0.4\linewidth]{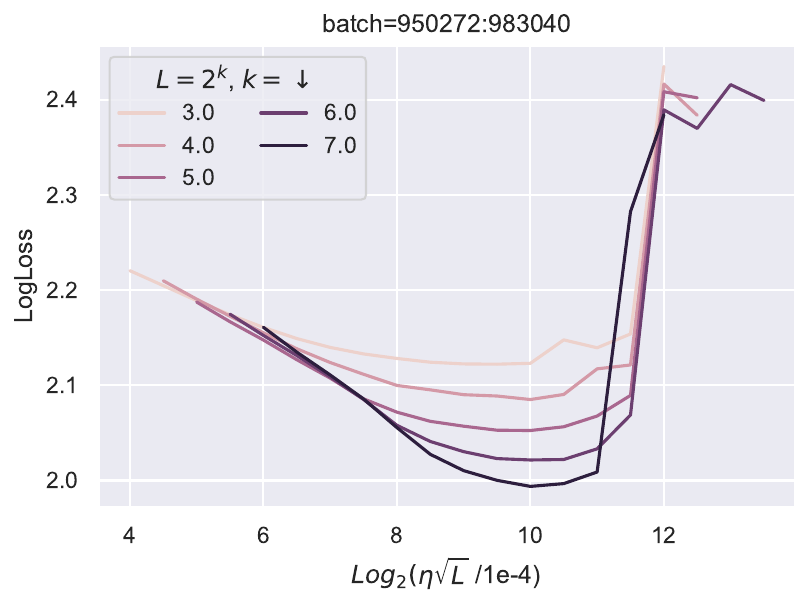}
    \caption{In the middle of (Megatron) transformer training, optimal learning rate is approximately invariant (Left), while at the end of training, it approximately scales like $1/\sqrt{L}$. However, the $1/\sqrt{L}$ scaling transfers the maximum viable learning rate better in either case. }
    \label{fig:megatron-scaling-shifts}
\end{figure}

Because transformers have block depth 2, as discussed in \cref{sec:bd2}, we have plenty of reasons to suspect that no parametrization of (learning rate, block multiplier) will be able to robustly transfer hyperparameters across depth for transformers.

Here we do a large scale experiment using Megatron trained on Common Crawl and catalogue our observations.\footnote{We train the models for 3900 steps, using cosine decay schedule with 500 warmup steps. We use a sequence length of 4096, batch size 256, resulting in approximately 4B tokens per training run.}
In summary, in our particular setup (which should be close to most large language model pretraining), we see that the $1/\sqrt L$ scaling seems to transfer hyperparameters at the end of training (\Cref{{fig:megatron-scaling-shifts}}(Right)). However, we also see that 1) deeper does worse in initial training (\Cref{{fig:megatron-deeper-worse}}(Left)), and 2) optimal hyperparameters scale like $\Theta(1)$ in the middle of training (\Cref{fig:megatron-scaling-shifts}(Left)).
Combined with the theoretical insights of \Cref{sec:bd2}, this leads us to conclude that while the $1/\sqrt L$ scaling can potentially be practically useful in transformer training, it is likely to be brittle to architectural  and algorithmic changes, or even simple things like training time.

In fact, we observe that transformers are insensitive to the block multiplier $a$ (\Cref{fig:megatron_bm_insensitive}), so that the only relevant hyperparameter is really just learning rate. Thus, empirically measuring the scaling trend of the optimal learning rate, as done in modern large scale pretraining, can be a practically more robust way to transfer hyperparameters.

Here $L$ is the number of transformer layers, each of which consists of an attention layer and an MLP layer (each of which has depth 2).

\subsection{Feature Diversity}

\begin{figure}[t]
    \centering
    \includegraphics[width=\textwidth]{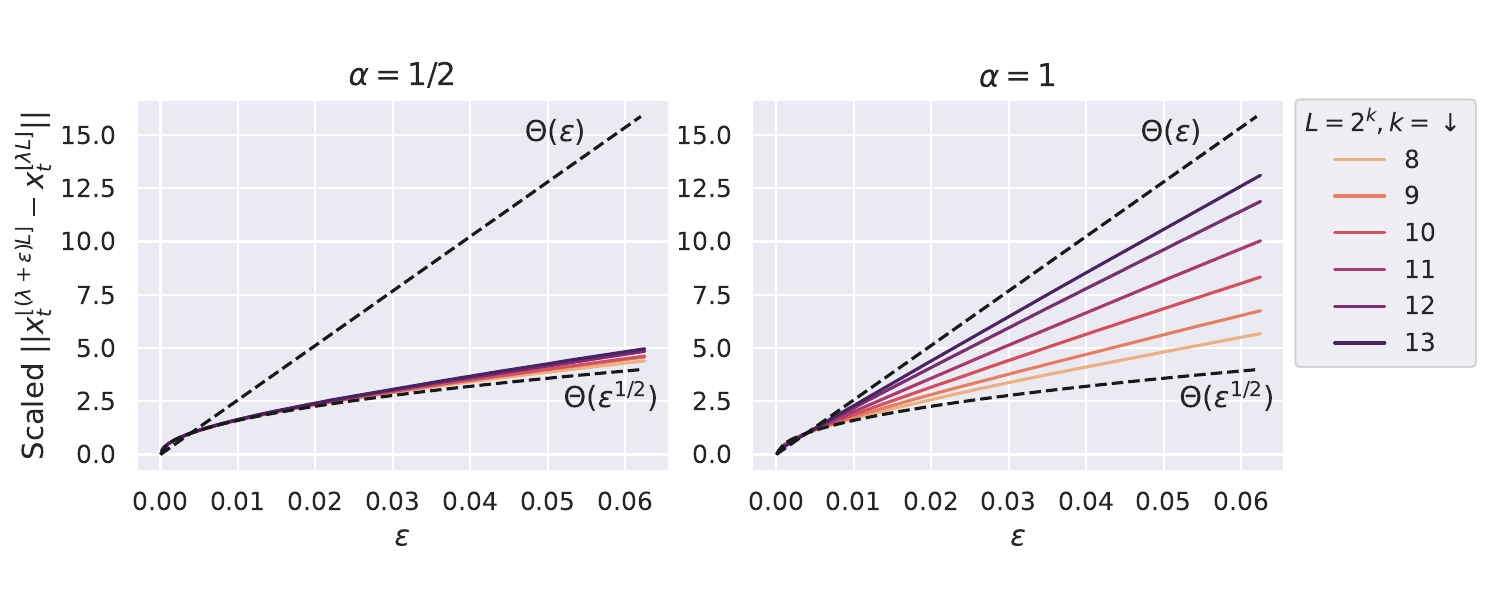}
    \caption{Difference between feature at layer $\lfloor\lambda L\rfloor$ and feature at layer $\lfloor(\lambda+\epsilon)L\rfloor$ as a curve of $\epsilon$ for width $n=256$ and varying depths. For a clean presentation, each curve is scaled by a constant so it always passes $(1/256, 1)$. The feature diversity exponent $\kappa$ depends on the growth of the curve when $L\to\infty$. For Depth-$\mu$P (left), the curve is always close to $\epsilon^{1/2}$, meaning $\kappa=1/2$. For ODE parametrization (right), the curve shifts from $\epsilon^{1/2}$ to $\epsilon$ when $L$ grows, indicating its $\kappa$ goes to $0$ in the infinite depth limit. }
    \label{fig:feature-diversity}
\end{figure}

In this section, we empirically verify our claims about feature diversity exponent (\Cref{clm:redundant,clm:depth_mup}). We use the same setup as in the last section, i.e., we train deep residual networks of width $n=256$ on CIFAR-10 dataset with Adam and batch size $64$. In \Cref{fig:feature-diversity}, we compare two parametrizations, Depth-$\mu$P ($\alpha=\gamma=1/2$) and the ODE parametrization $(\alpha,\gamma)=(1,0)$. We measure $\left\| \xx^{\lfloor (\lambda + \epsilon) L \rfloor}_t - \xx_t^{\lfloor \lambda L \rfloor} \right\|\odefeq d(\epsilon)$ at $t=1000$ for the two parametrizations and varying depth. For each parametrization and depth $L$, we rescale function $d$ by multiplying a constant $c$ such that $c\cdot d(1/256)=1$, and then plot the rescaled function $c\cdot d$ for a clean presentation. One can observe clearly that Depth-$\mu$P has feature diversity exponent (almost) $1/2$ for any $L$, while the curves for ODE parametrization move from $\epsilon^{1/2}$ to $\epsilon$ when $L$ grows.   This exactly fits our theory that Depth-$\mu$P maximizes the feature diversity, while other parametrizations (even with feature learning) have smaller feature diversity exponents that should go to $0$ in the infinite depth limit.

\begin{figure}[t]
    \centering
    \includegraphics[width=\textwidth]{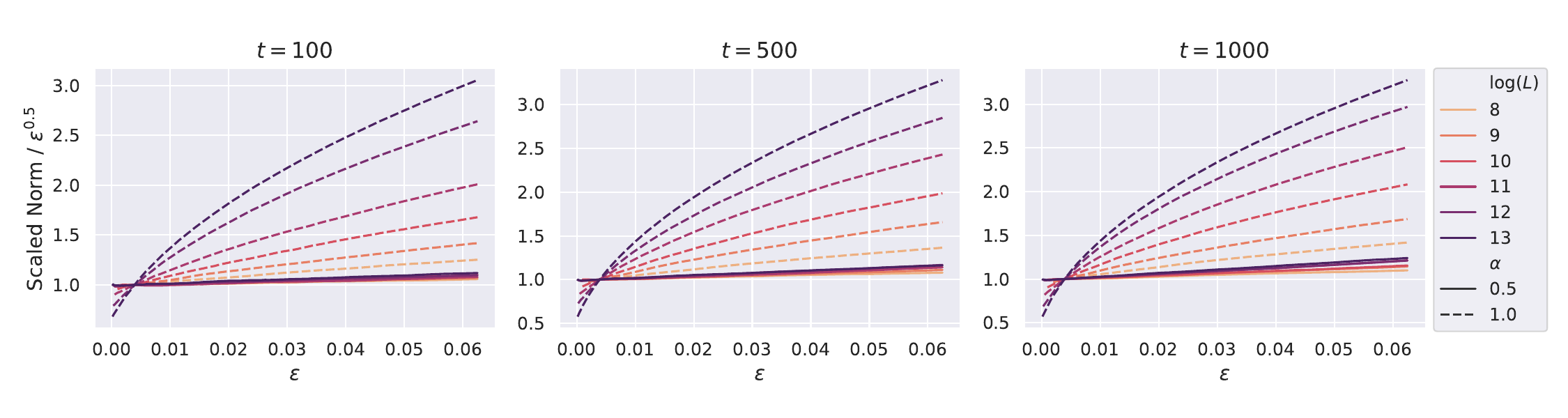}
    \caption{Same setup as \Cref{fig:feature-diversity} but at step $t=100, 500, 1000$, and each curve is scaled by dividing a constant and \emph{additional} $\epsilon^{1/2}$ so it always passes $(1/256, 1)$. The curve indicating feature diversity exponent $\kappa$ exactly $1/2$ should be a horizontal line at $1$. For Depth-$\mu$P ($\alpha=0.5$), the curves are almost horizontal. For ODE parametrization ($\alpha=1$), slopes of the curves are larger with larger $L$ and larger $t$.}
    \label{fig:feature-diversity-ode}
\end{figure}

\paragraph{Growth along with $L$ and $t$.} In \Cref{fig:feature-diversity-ode}, we measure $d(\epsilon)$ at $t=100, 500, 1000$, and rescale it by \emph{dividing} \emph{additional} $\epsilon^{0.5}$ and a constant $c$ such that $\frac{d(1/256)}{c\cdot \epsilon^{0.5}}=1$, and then plot the rescaled function $d/(c\cdot \epsilon^{0.5})$ for a clean comparison between $d$ and $\epsilon^{0.5}$. We observe that for both Depth-$\mu$P and ODE parametrization, the slopes of the curves grow along with $L$ and $t$. The growth along $t$ can be explained by the cumulative correlation between layers. The growth along $L$ for ODE parametrization is because the independent components between nearby layers decrease when $L$ grows. We do not have a clear understanding for the growth along $L$ for Depth-$\mu$P and we leave it as a future work.

\begin{figure}[t]
    \centering
    \includegraphics[width=\textwidth]{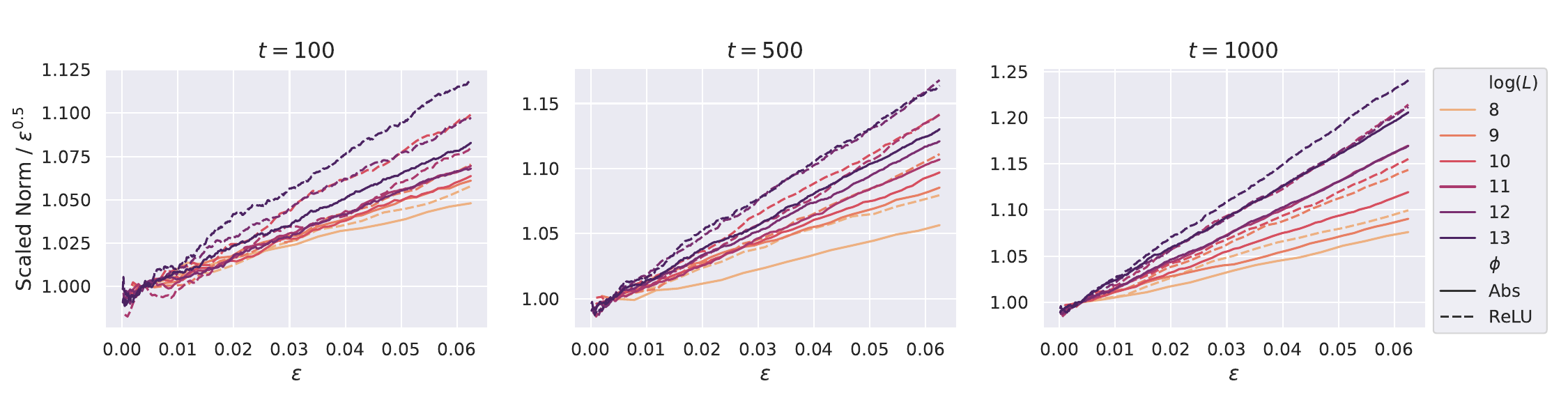}
    \caption{Same setup as \Cref{fig:feature-diversity-ode}, but comparing Depth-$\mu$P with ReLU activation and absolute value activation. Each curve is scaled by dividing a constant and $\epsilon^{1/2}$ so it always passes $(1/256, 1)$. The curve indicating feature diversity exponent $\kappa$ exactly $1/2$ should be a horizontal line at $1$. For both activations, slopes of curves are small, but growing along with $L$ and $t$. The slopes with absolute value activation 
    ($\phi=\mathrm{Abs}$) are slower than the slopes with ReLU activation ($\phi=\mathrm{ReLU}$), indicating feature diversity is higher with absolute value activation.}
    \label{fig:feature-diversity-abs}
\end{figure}

\paragraph{Absolute value activation increases feature diversity.} In \Cref{fig:feature-diversity-abs}, we plot the same curves as in \Cref{fig:feature-diversity-ode} but comparing ReLU activation and absolute value activation under Depth-$\mu$P. We observe that the slope of the curves for absolute value activation is smaller than ReLU activation. It matches our theory that absolute value activation increases feature diversity.

\section*{Acknowledgement}
We thank Huishuai Zhang, Jeremy Bernstein, Edward Hu, Michael Santacroce, Lucas Liu for their helpful comments and discussion. 
D. Yu was supported by NSF and ONR. Part of this work was done during D. Yu's internship at Microsoft. 

\section*{Author Contributions}

GY developed the core theory and ran experiments in early part of the exploratory stage and most experiments in the final draft.
DY worked on and proved key claims for linear resnets (including the limiting equations, convergence, and classification of parametrization), drafted the very first version of the paper, and ran experiments verifying the theoretical claims (including the convergence of linear case and feature diversity separation).
CZ ran experiments in later part of the exploratory stage. They revealed the viability of Depth-$\mu$P in the block depth 1 case, in contrast to the general block depth case.
CZ also ran the Megatron experiments in the final version of the paper.
SH contributed to brainstorming since the beginning of the project, wrote the warm-up section on linear networks, formalized the notion of feature diversity exponent, and helped transforming experimental results into plots and visualizations.
 
\newpage

\bibliography{reference}

\begin{thebibliography}{28}
\providecommand{\natexlab}[1]{#1}
\providecommand{\url}[1]{\texttt{#1}}
\expandafter\ifx\csname urlstyle\endcsname\relax
  \providecommand{\doi}[1]{doi: #1}\else
  \providecommand{\doi}{doi: \begingroup \urlstyle{rm}\Url}\fi

\bibitem[Allen-Zhu et~al.(2019)Allen-Zhu, Li, and Song]{allenzhu2019convergence}
Z.~Allen-Zhu, Y.~Li, and Z.~Song.
\newblock A convergence theory for deep learning via over-parameterization, 2019.

\bibitem[Chizat and Bach(2018)]{chizat2018global}
L.~Chizat and F.~Bach.
\newblock On the global convergence of gradient descent for over-parameterized models using optimal transport, 2018.

\bibitem[Chizat et~al.(2020)Chizat, Oyallon, and Bach]{chizat2020lazy}
L.~Chizat, E.~Oyallon, and F.~Bach.
\newblock On lazy training in differentiable programming, 2020.

\bibitem[Hanin and Rolnick(2018)]{hanin2018start}
B.~Hanin and D.~Rolnick.
\newblock How to start training: The effect of initialization and architecture, 2018.

\bibitem[Hayou(2023)]{hayou2023infinitedepth}
S.~Hayou.
\newblock On the infinite-depth limit of finite-width neural networks.
\newblock \emph{Transactions on Machine Learning Research}, 2023.

\bibitem[Hayou and Yang(2023)]{hayou2023width}
S.~Hayou and G.~Yang.
\newblock Width and depth limits commute in residual networks.
\newblock In A.~Krause, E.~Brunskill, K.~Cho, B.~Engelhardt, S.~Sabato, and J.~Scarlett, editors, \emph{Proceedings of the 40th International Conference on Machine Learning}, volume 202 of \emph{Proceedings of Machine Learning Research}, pages 12700--12723. PMLR, 23--29 Jul 2023.
\newblock URL \url{https://proceedings.mlr.press/v202/hayou23a.html}.

\bibitem[Hayou et~al.(2021)Hayou, Clerico, He, Deligiannidis, Doucet, and Rousseau]{hayou21stable}
S.~Hayou, E.~Clerico, B.~He, G.~Deligiannidis, A.~Doucet, and J.~Rousseau.
\newblock Stable resnet.
\newblock In A.~Banerjee and K.~Fukumizu, editors, \emph{Proceedings of The 24th International Conference on Artificial Intelligence and Statistics}, volume 130 of \emph{Proceedings of Machine Learning Research}, pages 1324--1332. PMLR, 13--15 Apr 2021.
\newblock URL \url{https://proceedings.mlr.press/v130/hayou21a.html}.

\bibitem[He et~al.(2016{\natexlab{a}})He, Zhang, Ren, and Sun]{he2016deep}
K.~He, X.~Zhang, S.~Ren, and J.~Sun.
\newblock Deep residual learning for image recognition.
\newblock In \emph{Proceedings of the IEEE conference on computer vision and pattern recognition}, pages 770--778, 2016{\natexlab{a}}.

\bibitem[He et~al.(2016{\natexlab{b}})He, Zhang, Ren, and Sun]{he2016identity}
K.~He, X.~Zhang, S.~Ren, and J.~Sun.
\newblock Identity mappings in deep residual networks.
\newblock In \emph{Computer Vision--ECCV 2016: 14th European Conference, Amsterdam, The Netherlands, October 11--14, 2016, Proceedings, Part IV 14}, pages 630--645. Springer, 2016{\natexlab{b}}.

\bibitem[Jacot et~al.(2020)Jacot, Gabriel, and Hongler]{jacot2020neural}
A.~Jacot, F.~Gabriel, and C.~Hongler.
\newblock Neural tangent kernel: Convergence and generalization in neural networks, 2020.

\bibitem[Jelassi et~al.(2023)Jelassi, Hanin, Ji, Reddi, Bhojanapalli, and Kumar]{jelassi2023depth}
S.~Jelassi, B.~Hanin, Z.~Ji, S.~J. Reddi, S.~Bhojanapalli, and S.~Kumar.
\newblock Depth dependence of $\mu$p learning rates in relu mlps, 2023.

\bibitem[Liu et~al.(2020)Liu, Liu, Gao, Chen, and Han]{liu2020understanding}
L.~Liu, X.~Liu, J.~Gao, W.~Chen, and J.~Han.
\newblock Understanding the difficulty of training transformers.
\newblock \emph{arXiv preprint arXiv:2004.08249}, 2020.

\bibitem[Noci et~al.(2022)Noci, Anagnostidis, Biggio, Orvieto, Singh, and Lucchi]{noci2022signal}
L.~Noci, S.~Anagnostidis, L.~Biggio, A.~Orvieto, S.~P. Singh, and A.~Lucchi.
\newblock Signal propagation in transformers: Theoretical perspectives and the role of rank collapse, 2022.

\bibitem[Noci et~al.(2023)Noci, Li, Li, He, Hofmann, Maddison, and Roy]{noci2023shaped}
L.~Noci, C.~Li, M.~B. Li, B.~He, T.~Hofmann, C.~Maddison, and D.~M. Roy.
\newblock The shaped transformer: Attention models in the infinite depth-and-width limit, 2023.

\bibitem[OpenAI(2023)]{openai2023gpt4}
OpenAI.
\newblock Gpt-4 technical report, 2023.

\bibitem[Shoeybi et~al.(2019)Shoeybi, Patwary, Puri, LeGresley, Casper, and Catanzaro]{shoeybi2019megatron}
M.~Shoeybi, M.~Patwary, R.~Puri, P.~LeGresley, J.~Casper, and B.~Catanzaro.
\newblock Megatron-lm: Training multi-billion parameter language models using model parallelism.
\newblock \emph{arXiv preprint arXiv:1909.08053}, 2019.

\bibitem[Silver et~al.(2016)Silver, Huang, Maddison, Guez, Sifre, van~den Driessche, Schrittwieser, Antonoglou, Panneershelvam, Lanctot, Dieleman, Grewe, Nham, Kalchbrenner, Sutskever, Lillicrap, Leach, Kavukcuoglu, Graepel, and Hassabis]{alphago}
D.~Silver, A.~Huang, C.~J. Maddison, A.~Guez, L.~Sifre, G.~van~den Driessche, J.~Schrittwieser, I.~Antonoglou, V.~Panneershelvam, M.~Lanctot, S.~Dieleman, D.~Grewe, J.~Nham, N.~Kalchbrenner, I.~Sutskever, T.~P. Lillicrap, M.~Leach, K.~Kavukcuoglu, T.~Graepel, and D.~Hassabis.
\newblock Mastering the game of go with deep neural networks and tree search.
\newblock \emph{Nature}, 529:\penalty0 484--489, 2016.

\bibitem[Srivastava et~al.(2015)Srivastava, Greff, and Schmidhuber]{srivastava2015highway}
R.~K. Srivastava, K.~Greff, and J.~Schmidhuber.
\newblock Highway networks, 2015.

\bibitem[Vaswani et~al.(2017)Vaswani, Shazeer, Parmar, Uszkoreit, Jones, Gomez, Kaiser, and Polosukhin]{vaswani2017attention}
A.~Vaswani, N.~Shazeer, N.~Parmar, J.~Uszkoreit, L.~Jones, A.~N. Gomez, L.~Kaiser, and I.~Polosukhin.
\newblock Attention is all you need, 2017.

\bibitem[Yang(2020{\natexlab{a}})]{yang2020scaling}
G.~Yang.
\newblock Scaling limits of wide neural networks with weight sharing: Gaussian process behavior, gradient independence, and neural tangent kernel derivation, 2020{\natexlab{a}}.

\bibitem[Yang(2020{\natexlab{b}})]{yang2020tensor}
G.~Yang.
\newblock Tensor programs ii: Neural tangent kernel for any architecture, 2020{\natexlab{b}}.

\bibitem[Yang(2021)]{yang2021tensor_i}
G.~Yang.
\newblock Tensor programs i: Wide feedforward or recurrent neural networks of any architecture are gaussian processes, 2021.

\bibitem[Yang and Hu(2021)]{yang2021tensor}
G.~Yang and E.~J. Hu.
\newblock Tensor programs iv: Feature learning in infinite-width neural networks.
\newblock In \emph{International Conference on Machine Learning}, pages 11727--11737. PMLR, 2021.

\bibitem[Yang and Littwin(2023)]{yang2023tensor}
G.~Yang and E.~Littwin.
\newblock Tensor programs ivb: Adaptive optimization in the infinite-width limit, 2023.

\bibitem[Yang et~al.(2022)Yang, Hu, Babuschkin, Sidor, Liu, Farhi, Ryder, Pachocki, Chen, and Gao]{yang2022tensor}
G.~Yang, E.~J. Hu, I.~Babuschkin, S.~Sidor, X.~Liu, D.~Farhi, N.~Ryder, J.~Pachocki, W.~Chen, and J.~Gao.
\newblock Tensor programs v: Tuning large neural networks via zero-shot hyperparameter transfer.
\newblock \emph{arXiv preprint arXiv:2203.03466}, 2022.

\bibitem[Zhang et~al.(2019)Zhang, Dauphin, and Ma]{zhang2019fixup}
H.~Zhang, Y.~N. Dauphin, and T.~Ma.
\newblock Fixup initialization: Residual learning without normalization, 2019.

\bibitem[Zhang et~al.(2023)Zhang, Yu, Yi, Chen, and Liu]{zhang2023stabilize}
H.~Zhang, D.~Yu, M.~Yi, W.~Chen, and T.-Y. Liu.
\newblock Stabilize deep resnet with a sharp scaling factor $\tau$, 2023.

\bibitem[Zou et~al.(2018)Zou, Cao, Zhou, and Gu]{zou2018stochastic}
D.~Zou, Y.~Cao, D.~Zhou, and Q.~Gu.
\newblock Stochastic gradient descent optimizes over-parameterized deep relu networks, 2018.

\end{thebibliography}
\bibliographystyle{abbrvnat}

\newpage

\appendix

\section{Notations}
\label{sec:notations}

This section provides an introduction to the new TP notations from \cite{yang2023tensor}. We only require the definition of the inner and outer products in this paper. 

\paragraph{Averaging over $n$}
When $x \in \R^n$, we always use greek subscript $\alpha, \beta, \ldots \in [n]$ to index its entries.
Then $\la x_\alpha \ra_\alpha$ denotes its average entry.
This notation will only be used to average over $n$-dimensions, but not over constant dimensions.

\subsection{The Tensor Program Ansatz: Representing Vectors via Random Variables}
\label{sec:TPansatz}

From the Tensor Programs framework \citep{yang2022tensor}, we know that as width becomes large, the entries of the (pre-)activation
vectors and their gradients will become roughly iid, both at initialization and training. Hence any such vector's behavior can be
tracked via a random variable that reflects the distribution of its
entries. While we call this the ``Tensor Program Ansatz'', it is a completely rigorous calculus.

\subsubsection{Ket Notation}

Concretely, if $x\in\R^{n}$ is one such vector, then we write $\ket x \in \R$ (called a \emph{ket})
for such a random variable, such that $x$'s entries look like iid
samples from $\ket{x}$.%
For any two such vectors $x, y \in \R^n$, $(x_\alpha, y_\alpha) \in \R^2$ for each $\alpha$ will look like iid samples from the random vector $(\ket x, \ket y)$, such that, for example, $\lim_{n\to\infty}\frac{x^{\top}y}{n}=\EV \ket{x}\cdot \ket{y}$, which we write succinctly as just $\braket x y$.
Here $\bra x$ is called a \emph{bra}, interpreted as a sort of ``transpose'' to $\ket x$.
In our convention, $\ket x$ is always a random variable independent of $n$ and $x$ always has $\Theta(1)$ typical entry size.\footnote{i.e., $\|x\|^{2}/n=\Theta(1)$ as $n\to\infty$}.

This notation can be generalized to the case where $\xx \in \mathbb{R}^{n \times k}, \yy \in \mathbb{R}^{n \times j}$. In this case, we can think of $\braket{\xx}{\yy}$ as the $k \times j$ matrix given by $(\braket{x_a}{y_b})_{\substack{1 \leq a \leq k \\ 1 \leq b \leq j}}$.

Because we will often need to multiply a ket with a diagonal matrix, we introduce a shorthand:
\begin{equation}
  \ket\xx _{\cchi} = \ket\xx \Diag(\cchi),\label{eqn:multdiagnotation}
\end{equation}
if $\xx$ is $n\times k$ and $\cchi$ is a $k$-dimensional vector.

\subsubsection{Outer Product}
\label{sec:ket_outer}

Likewise, if both $\xx$ and $\yy$ have shape $n\times k$, the expression
\[
\ket{\xx}\bra{\yy}\ \text{ represents the limit of $\xx\yy^{\trsp}\in\R^{n\times n}$.}
\]
More formally,
$\ket{\xx}\bra{\yy}$ is defined as an operator that takes a ket $\ket{\zz}\in\R^{j}$
and return the ket
\[
(\ket{\xx}\bra{\yy})\ket{\zz}=\ket{\xx}(\braket{\yy}{\zz})\in\R^{j}
\]
i.e., it returns the random vector $\ket{\xx}\in\R^{k}$ multiplied
by the deterministic matrix $\braket{\yy}{\zz}\in\R^{k\times j}$
on the right. 
This corresponds to the limit of $\xx \yy^\trsp \zz/n$.
Likewise, $\ket{\xx}\bra{\yy}$ acts on a bra $\bra{\ww}\in\R^{j}$
by
\[
\bra{\ww}(\ket{\xx}\bra{\yy})=(\braket{\ww}{\xx})\bra{\yy}\in\R^{j}.
\]
which corresponds to the limit of $\frac 1 n \ww^\trsp \xx \yy^\trsp$.
This definition of $\ket{\xx}\bra{\yy}$ makes the expressions
\[
\ket{\xx}\braket{\yy}{\zz},\quad\braket{\ww}{\xx}\bra{\yy},\quad\braket{\ww}{\xx}\braket{\yy}{\zz}
\]
unambiguous (since any way of ordering the operations give the same
answer).

\begin{rem}[Potential Confusion]
One should \emph{not} interpret $\ketdbra{\xx}{}\yy$ as the scalar random variable $\ket\xx \cdot \ket \yy  = \sum_{i=1}^k \ket {x^i}\ket{y^i}$, which would act on a ket $\ket{\zz}$ to produce $(\bra{\xx} \cdot \bra\yy)\ket{\zz} = \EV (\ket\xx \cdot \ket \yy)\ket{\zz} $, which is deterministic.
On the other hand, $\ketdbra{\xx}{}\yy \zz \ra$ is always a linear combination of $\ket \xx$, a nondeterministic random variable in general.
In particular, any correlation between $\ket \xx$ and $\ket \yy$ does not directly play a role in their outer product $\ketdbra\xx{}\yy$: we always have $\ketdbra{\xx}{}\yy \zz \ra = \ketdbra{\xx}{}\yy^{\bx1} \ket\zz^{\bx1}$, where $(\ket \yy^{\bx1}, \ket \zz^{\bx1})$ is an iid copy of $(\ket \yy, \ket \zz)$ independent from $\ket \xx$.
\end{rem}

\paragraph{Outer Product with Diagonal Inserted}
Finally, if $\cchi\in\R^{k}$ is deterministic, then (consistent with \cref{eqn:multdiagnotation}) we define $\ketdbra{\xx}{\cchi}{\yy}$
as the operator that acts on kets $\ket{\zz}\in\R^{j}$ by
\[
(\ketdbra{\xx}{\cchi}{\yy})\ket{\zz}=\ketdbra{\xx}{\cchi}{\yy}{\zz}\ra=\ket{\xx}\Diag(\cchi)(\braket{\yy}{\zz})\in\R^{j}.
\]
Morally, $\ketdbra{\xx}{\cchi}{\yy}$ is just a shorter way of writing
$\ket{\xx}\Diag(\cchi)\bra{\yy}$ and represents the limit of $\xx\Diag(\cchi)\yy^{\trsp}$.
In particular, $\ketdbra{\xx}{\boldsymbol{1}}{\yy}=\ket{\xx}\bra{\yy}$.

\subsubsection{Nonlinear Outer Product}

If $x y^\trsp \in \R^{n\times n}$ is the (linear) outer product of two vectors $x \in \R^n$ and $y \in \R^n$, then $\phi(xy^\trsp)$, the entrywise application of nonlinear $\phi: \R \to \R$ to $xy^\trsp$, is a kind of \emph{nonlinear outer product}.%
Passing to the ket notation, in general we define $\phi\left(\ket{\xx}\bra{\yy}\right)$
as the operator that acts on kets as
\[
\phi\left(\ket{\xx}\bra{\yy}\right)\ket{\zz}\defeq\EV_{\bx{1}}\phi\left(\sum_{i=1}^{k}\ket{x^{i}}\ket{y^{i}}^{\bx{1}}\right)\ket{\zz}^{\bx{1}}
\]
where $\left(\ket{y^{1}}^{\bx{1}},\ldots,\ket{y^{k}}^{\bx{1}},\ket{\zz}^{\bx{1}}\right)$
is an iid copy of $\left(\ket{y^{1}},\ldots,\ket{y^{k}},\ket{\zz}\right)$
independent from $\ket{\xx}$ and the expectation is taken only over
the former. 
This is just like, in the finite $n$ case,
\[
\phi\left(\xx \yy^\trsp \right)\zz/n = \phi\left(\sum_{i=1}^{k} x^{i}{y^{i\trsp}}\right) {\zz}/n.
\]

Moreover, if $\ket{\ww}\in\R^{j},\ket{\zz}\in\R^{k}$,
then
\begin{align*}
\bra{\ww}\phi\left(\ket{\xx}\bra{\yy}\right)\ket{\zz} & =\bra{\ww}\phi\left(\ket{\xx}\bra{\yy}^{\bx{1}}\right)\ket{\zz}^{\bx{1}}\in\R^{j\times k}\\
 & =\EV\phi\left(\sum_{i=1}^{k}\ket{x^{i}}\ket{y^{i}}^{\bx{1}}\right)\left(\ket{\ww}\otimes\ket{\zz}^{\bx{1}}\right)
\end{align*}
where $\otimes$ denotes outer product of vectors and expectation
is taken over everything.

More generally, if $\phi: \R^t \to \R$, then $\phi\Big(\ket{\xx_1}\bra{\yy_1}, \ldots, \ket{\xx_t}\bra{\yy_t}\Big)$ is an operator taking kets to kets, defined by
\begin{align*}
  \phi\Big(\ket{\xx_1}\bra{\yy_1}, \ldots, \ket{\xx_t}\bra{\yy_t}\Big) \ket \zz \defeq
  \EV_{\bx{1}}\phi\left(\sum_{i=1}^{k}\ket{x_1^{i}}\ket{y_1^{i}}^{\bx{1}}, \ldots, \sum_{i=1}^{k}\ket{x_t^{i}}\ket{y_t^{i}}^{\bx{1}}\right)\ket{\zz}^{\bx{1}}
\end{align*}

\begin{rem}[Potential Confusion]
Note $\phi(\ket \xx \bra \yy)$ is not the image of the operator $\ket \xx \bra \yy$ under $\phi$ in the continuous function calculus of operators, but rather a ``coordinatewise application'' of $\phi$.
For example, if $\phi(t) =t^2$, then $\phi(\ket x \bra y)$ \emph{is not} $\ket x \bra y x\ra \bra y$, the latter being what typically ``squaring an operator'' means, but rather $\ket{x}^2 \bra{y}^2 = \ket{x\odot x}\bra{y\odot y}$.
\end{rem}

\subsubsection{Comparison with Previous $Z^\bullet$ Notation}
For readers familiar with the \emph{Tensor Programs} papers, this new ``bra-ket'' notation
(aka Dirac notation) relates to the old $Z^\bullet$ notation by
\[
\ket x= Z^{x},\quad\braket xy=\EV Z^{x}Z^{y}.
\]
The new notation's succinctness of expectation inner product should already be apparent.
Furthermore, the old notation is not very compatible with multi-vectors whereas $\ket x$ makes it clear that $\ra$ represents the constant dimension side.
Consequently, (nonlinear) outer product is awkward to express in it, especially when its contraction with random variables requires an explicit expectation symbol $\EV$.

\section{Infinite-Width Limit with the Bra-ket notation}\label{sec:tp}
As before, when the width $n$ of the program goes to infinity, one can infer how the program behaves via a calculus of random variables.
We define them below via the new ket notation instead of the earlier $Z$ notation.
\paragraph{Ket Construction.}
We recursively define the random variable $\ket x$ (called a \emph{ket}) for each vector $x$ and deterministic number $\mathring{\theta}$ for each scalar $\theta$ in the program.
For a vector $Wx$ in the program, we also define random variables $\hatket{Wx}$ and $\dotket{Wx}$ (called \emph{hat-ket} and \emph{dot-ket} respectively) such that $\ket{Wx} = \hatket{Wx} + \dotket{Wx}$. These are the same as $\hat{Z}$ and $\dot{Z}$ in the old TP notation \citep{yang2022tensor} and they satisfy
\begin{itemize}

\item[\texttt{Hat}]
All hat-kets
are jointly Gaussian with zero-mean and covariance%
\footnote{In \cref{eqn:hatketcovar}, $\ind(W=U)$ is the deterministic number that is 1 iff $W$ and $U$ are the same matrix (as symbols in the program) and 0 otherwise. This should \emph{not} be interpreted as a random variable that is 1 precisely when $W$ and $U$ take the same values.}
\begin{align}
  \Cov(\hatket{Wx}, \hatket{Uy}) &= \ind(W = U)\braket x y \label{eqn:hatketcovar}
\end{align}

\item[\texttt{Dot}] Every dot-ket is a linear combination of previous kets, expressed by the following equation
\begin{equation}
  \dotket{Wx}\defeq \sum_{y \in \xx} \ket{y}\EV \frac{\partial \ket x}{\partial \hatket{W^\trsp y}}\label{eqn:dotket}
\end{equation}
\end{itemize}
\cref{eqn:dotket} is the same equation as in\citep[Zdot]{yang2022tensor} 
but formulated much more succinctly in the bra-ket notation:
\begin{align*}
  \text{\citep[Zdot]{yang2022tensor}},\quad
  \Zdot^{Wx}
  &= \sum_{y \in \xx } Z^{y}\EV \f{\partial Z^x} {\partial \hat Z^{W^\trsp y}}.
\end{align*}

\newcommand\lhatbra[1]{\mkern+3mu\check{\mkern-3mu\langle}#1 \ob}
\newcommand\lhatbraket[2]{\mkern+3mu\check{\mkern-3mu\langle}#1 \ob #2 \rangle}

There is an alternative notion for $\dotket{Wx}$ in \citet{yang2023tensor} that write 
\[\dotket{Wx}=\ket{\xx}\dbra{W^\top \xx}x\ra.\] This is more convenient to write as we introduce the operator view.

We can see the ket $\ket{W x}$ as the result of the action of an operator on the ket $\ket{x}$.
\begin{defn}\label{defn:oplim}
  Let $W$ be an initial matrix in a Tensor Program.
  We define $\oplim{W}, \hatoplim{W}, \dotoplim{W}$ to be the linear operators on kets
  \footnote{\label{footnote:hilbertspace}
    To be rigorous, we need to specify the ``Hilbert space'' of kets.
    This is somewhat pedantic and not crucial to the key points of this paper, but the Hilbert space can be constructed as follows:
    Let $\sigma(\pi)$ be the $\sigma$-algebra generated by the kets of the program $\pi$.
    Let $\Sigma(\pi) \defeq \bigcup_{\pi' \supseteq \pi} \sigma(\pi)$ be the union (more precisely, the direct limit) of $\sigma(\pi')$ over all programs $\pi'$ extending $\pi$.
    Then the Hilbert space in question is the $L^2$ space of random variables over the $\Sigma$ of our program.
    }
  that act by 
  \begin{align*}
      \hatoplim{W} x \ra &\defeq \hatket{Wx}\\
      \dotoplim{W} x \ra &\defeq \dotket{Wx}\\
      \opket{W} x &\defeq \hatoplim{W} x \ra + \dotoplim{W} x \ra.
  \end{align*}
  Any linear operator that is equal to $\oplim{W}$ for some initial matrix $W$ is called an \emph{initial operator}.
\end{defn}

We also define the adjoint relations between the operators:
\begin{align*}
    \hatoplim{W}^\dagger &= \dotoplim{W^\trsp},\\
    \dotoplim{W}^\dagger &= \hatoplim{W^\trsp},\\
    \oplim{W}^\dagger &= \oplim{W^\trsp}.
\end{align*}

\paragraph{Parameter Update}

In the SGD case, the parameter update of $W^l$ is simple. With the operator notation and outer product notation, we can write
\[\oplim{W_{t+1}^l}=\oplim{W_t^l}-\eta \ketdbra{\tilde \del h^l_t}{\chi_t}{x^{l-1}_t}. \]

In this work, $\Delta$ denotes change for one step, i.e., \[\oplim{\Delta W_{t+1}^l}=-\eta \ketdbra{\tilde \del h^l_t}{\chi_t}{x^{l-1}_t};\] $\bar \Delta$ denotes total change, i.e., \[
\oplim{\bar \Delta W_t^l} =-\sum_{\tau=0}^{t-1}\eta \ketdbra{\tilde \del h^l_\tau}{\chi_\tau}{x^{l-1}_\tau}, 
\] which we write succinctly $\oplim{\bar \Delta W_t^l}=-\eta \ketdbra{\tilde \del \hh^l_{<t}}{\cchi}{\xx^{l-1}_{<t}}$. (Compared to~\citet{yang2023tensor}, $\Delta$ and $\bar\Delta$ are changed from $\delta$ and $\Delta$ because we want to use $\delta$ for gradients instead of $d$, which is now used for depth differentiation).

Note in the general case, 
\[\oplim{\Delta W_{t+1}^l}=-\eta \Qketdbra{\tilde \del \hh^l_{\leq t}}{\cchi_{\leq t}}{\xx^{l-1}_{\leq t}}\]
where \[\Qketdbra{\tilde \del \hh^l_{\leq t}}{\cchi_{\leq t}}{\xx^{l-1}_{\leq t}}\odefeq Q_t^l(\ketdbra{\tilde \del h^l_0}{\chi_0}{x^{l-1}_0}, \ldots, \ketdbra{\tilde \del h^l_t}{\chi_t}{x^{l-1}_t}).\]
So 
\begin{equation}\label{eq:qbar}
    \oplim{\bar\Delta W_t^l}=-\eta \sum_{\tau =0}^{t-1} \Qketdbra{\tilde \del \hh^l_{\leq \tau}}{\cchi_{\leq  \tau}}{\xx^{l-1}_{\leq \tau}}.
\end{equation}

For the rest of the paper, we write $\oplim{\bar \Delta W_t^l}=-\eta \ketdbra{\tilde \del \hh^l_{<t}}{\cchi}{\xx^{l-1}_{<t}}$ for convenience. The generalization to \cref{eq:qbar} follows \citet{yang2023tensor}.

\section{Details of the linear case}\label{app:proof_linear}

\subsection{Proof sketch of \Cref{prop:derivatives}}
Here we provide a proof sketch of \Cref{prop:derivatives}, the formal prove is implied by the existence of $\Gamma$ and $C$ in the infinite depth limit.
\begin{proof}[Proof sketch]
The claims can be reasoned by induction on $t$ and $l$. Let us take $\ket{x_t^l}$ as an example, since $\ket{\tilde \del x_t^{l-1}}$ is symmetric with $\ket{x_t^l}$. 
By expanding the definition of $\ket{x_t^l}$, we have 
\[\ket{x_t^l} = \ket{x_t^{l-1}} + \frac1{\sqrt L} \hatket{W_0^l x_t^{l-1}} + \frac1{\sqrt L} \sum_{s=1}^{t-1} \ket{\tilde \del x_s^{l}} \left(\frac{\partial \ket{x_t^{l-1}}}{\partial \hatket{W_0^{l\top} \tilde \del x_s^{l}}} - \frac1{\sqrt L} \braket{x_s^{l-1}}{x_t^{l-1}} \right).\]
Note by induction, $\braket{x_s^{l-1}}{x_t^{l-1}}=\cO(1)$ and $\frac{\partial x_t^{l-1}}{\partial \hatket{W_0^{l\top} \tilde \del x_s^{l}}}=\cO(1/\sqrt L)$,
so
\begin{align*}
    \ket{x_t^l} & = \ket{x_t^{l-1}} + \frac1{\sqrt L} \hatket{W_0^l x_t^{l-1}} + \cO\left(\frac1L\right)\sum_{s=1}^{t-1} \ket{\tilde \del x_s^{l}} \\
    & = \xi_t \ket{U} + \sum_{m=1}^l \frac1{\sqrt L} \hatket{W_0^m x_t^{m-1}}+\cO\left(\frac1L\right)\sum_{m'=1}^l\sum_{s'=1}^{t-1}\ket{\tilde \del x_{s'}^{m'}}.
\end{align*}
Then by unwinding $\ket{\tilde \del x_{s'}^{m'}}$ and noting that by induction, $\forall s <t$, $
    \frac{\partial \ket{\tilde \del x_{s'}^{m'}}}{\partial \hatket{W_0^mx_s^{m-1}}} = \cO\left(\frac1{\sqrt L}\right)$,
    $\frac{\partial \ket{\tilde \del x_{s'}^{m'}}}{\partial \hatket{W_0^{m\top}\tilde \del x_s^m}} = \cO\left(\frac1{\sqrt L}\right)$,
    $\frac{\partial \ket{\tilde \del x_{s'}^{m'}}}{\partial \ket{U}} = \cO\left(1\right)$,
    $\frac{\partial \ket{\tilde \del x_{s'}^{m'}}}{\partial \ket{nV}} = \cO\left(1\right)$,
we have 
\[\frac{\partial \ket{x_t^l}}{\partial \hatket{W_0^mx_s^{m-1}}} = \cO\left(\frac1{\sqrt L}\right),
     \frac{\partial \ket{x_t^l}}{\partial \hatket{W_0^{m\top}\tilde \del x_s^m}} = \cO\left(\frac1{\sqrt L}\right),
     \frac{\partial \ket{x_t^l}}{\partial \ket{U}} = \cO\left(1\right),
     \frac{\partial \ket{x_t^l}}{\partial \ket{nV}} = \cO\left(1\right).\]

Also by unwinding, $\forall \ket{y}\in \{\ket{x_s^m}, \ket{\tilde \del x_s^m}\}$,
\begin{align*}
    \braket{y}{x_t^l}= & \sum_{m'}\sum_{s'}\sum_{t'} \frac{\partial \ket{x_t^l}}{\partial \hatket{W_0^{m'}x_{t'}^{m'-1}}}\cdot \frac{\partial \ket{y}}{\partial \hatket{W_0^{m'}x_{s'}^{m'-1}}}\cdot \braket{x_{t'}^{m'-1}}{x_{s'}^{m'-1}} \\
    & + \sum_{m'}\sum_{s'}\sum_{t'}  \frac{\partial \ket{x_t^l}}{\partial \hatket{W_0^{m'\top}\tilde \del x_{t'}^{m'}}}\cdot \frac{\partial \ket{y}}{\partial \hatket{W_0^{m'\top}\tilde \del x_{s'}^{m'}}}\cdot  \braket{\tilde \del x_{t'}^{m'}}{\tilde \del x_{s'}^{m'}}  \\
    & + 
    \frac{\partial \ket{x_t^l}}{\partial \ket{U}} \cdot 
    \frac{\partial \ket{y}}{\partial \ket{U}} + 
    \frac{\partial \ket{x_t^l}}{\partial \ket{nV}}
    \frac{\partial \ket{y}}{\partial \ket{nV}} \\
    = & \cO(1). \qedhere
\end{align*}
\end{proof}

\subsection{Formal recursive formula of $\Gamma$ and $C$}
By the same way of expanding $\ket{x_t^l}$ and $\braket{y}{x_t^l}$, we formally derive the recursive formula for $\Gamma$ and $C$ below.

\begin{lemma}[Finite depth recursive formula for $\Gamma$ and $C$]\label{lemma:finite_depth_gamma_formal}
$\Gamma$ can be computed recursively as follows:

For $t=0, \ldots, T-1$,
\begin{itemize}
    \item $\forall q\in (0, 1], \Gamma_{t,-1,0,q}(0,q)  = \xi_t$,
    \item For $l=1,\ldots,L$, 
    $\forall r \leq t$, 
    $\forall p \in \left(\frac{l-1}L,\frac lL\right]$,
    $\forall q\in (0,1]$, $\forall b\in \{0, 1\}$,
    \begin{align*} 
     C_{t,s,0}(p)=&~\sum_{t'=-1}^{t}\sum_{s'=-1}^{s}\sum_{b\in\{0,1\}}\int_0^1 \Gamma_{t,t',0,b}\left(\frac{l-1}L,q\right) C_{t',s',b}(q)\Gamma_{s,s',0,b}\left(\frac{l-1}L,q\right) \dd q;\\
     \Gamma_{t,r,0,b}\left(p, q\right) = &~\Gamma_{t,r,0,b}\left(\frac{l-1}L, q\right) + \ind_{[(t=r) \wedge (b=0) \wedge (l=\lceil L q\rceil)]} \\
     & + \frac1L \sum_{s=0}^{t-1} \Gamma_{s,r,1,b}\left(\frac l L, q\right)\left(\Gamma_{t,s,0,1}\left(\frac{l-1}L, \frac l L\right)-C_{t,s,0}\left(\frac l L\right)\right).
    \end{align*}
    \item $\mathring f_t = \Gamma_{t,-1,0,1}(1, 1)$,
    \item $\mathring \chi_t = \ell'_t(\mathring f_t)$,
    \item $\forall q\in (0, 1], \Gamma_{t, -1, 1, 1}(1, q)=\mathring \chi_t$,
    \item For $l=L,\ldots,1$, 
    $\forall r \leq t$, 
    $\forall p \in \left(\frac{l-2}L,\frac{l-1} L\right]$,
    $\forall q\in (0,1]$, $\forall b\in \{0, 1\}$,
    \begin{align*}
        C_{t,s,1}\left(p+\frac1L\right)=&~\sum_{t'=-1}^{t}\sum_{s'=-1}^{s}\sum_{b\in\{0,1\}}\int_0^1 \Gamma_{t,t',1,b}(l/L,q) C_{t',s',b}(q)\Gamma_{s,s',1,b}(l/L,q) \dd q;\\
     \Gamma_{t,r,1,b}\left(p, q\right) = &~\Gamma_{t,r,1,b}\left(\frac{l}L, q\right) + \ind_{[(t=r) \wedge (b=1) \wedge (l=\lceil L q\rceil)]} \\
     & + \frac1L \sum_{s=0}^{t-1} \Gamma_{s,r,0,b}\left(\frac {l-1} L, q\right)\left(\Gamma_{t,s,1,0}\left(\frac{l}L, \frac l L\right)-C_{t,s,1}\left(\frac l L\right)\right).\\
     \end{align*}
\end{itemize}
\end{lemma}

The proof is straightforward from Program \ref{alg:tp-linear}. The recursive nature of $\Gamma$ and $C$ yields the following infinite-depth behavior.

\begin{prop}[Infinite depth limit of $\Gamma$ and $C$] \label{prop:infinite_depth_linear_formal}
In the limit $L\to \infty$, we  have $\forall p\in [0, 1], q\in (0,1], b\in\{0, 1\}$:
\begin{align*}
    & \Gamma_{t,-1,0,0}(0, q)=\xi_t; \\
  & \Gamma_{t,r,0,b}(p, q)=\ind_{[(t=r) \wedge (b=0) \wedge (p\geq q)]} + \int_0^p \sum_{s=0}^{t-1}\Gamma_{s,r,1,b}(p',q)\cdot (\Gamma_{t,s,0,1}(p', p')- C_{t,s,0}(p'))\dd p'; \\
     & \mathring f_t = \Gamma_{t,-1,0,1}(1, 1);\\
     & \mathring \chi_t = \ell'_t(\mathring f_t); \\
     & \Gamma_{t,-1,1,1}(1,q)=\mathring \chi_t; \\
     & \Gamma_{t,r,1,b}(p, q) = \ind_{[(t=r) \wedge (b=1) \wedge (p\leq q)]} + \int_p^1 \sum_{s=0}^{t-1} \Gamma_{s,r,0,b}(p',q) \cdot (\Gamma_{t,s,1,0}(p',p') - C_{t,s,1}(p'))\dd p';\\
     & C_{t,s,a}(p)=\sum_{t'=-1}^{t}\sum_{s'=-1}^{s}\sum_{b\in\{0,1\}}\int_0^1 \Gamma_{t,t',a,b}(p,q) C_{t',s',b}(q)\Gamma_{s,s',a,b}(p,q) \dd q.
\end{align*}
\end{prop}

\subsection{Convergence of $\Gamma$ and $C$ when $L=2^k$}

In this section, we prove $\Gamma$ and $C$ will converge when $L\to\infty$. For convenience, we will only consider the case when $L=2^k$ for some integer $k$. To distinguish $\Gamma$ and $C$ corresponding to different $L$, we add the depth as the superscript, i.e., $\Gamma^L$ and $C^L$.

\begin{thm}
    $\forall t\leq T,s<t, a\in \{0,1\}, b\in \{0, 1\}$, $\forall p\in [0, 1], q\in (0,1]$, 
    \begin{itemize}
        \item $\{\Gamma^{2^k}_{t,s,a,b}(p,q)\}_{k\in \N}$ is a Cauchy sequence, 
        \item $\{C^{2^k}_{t,s,a}(p)\}_{k\in \N}$ is a Cauchy sequence.
    \end{itemize}
\end{thm}

The proof is by induction on $t$. We will prove the following claims (A) (B) (C) (D) on $t > 0$ given they are satisfied for any $s<t$. For $t=0$, (A) (B) (C) (D) are trivial. 
\paragraph{Assumption on $s<t$} Assume $\exists c>1$ such that $\forall L > L'$ and $L=2^k$ for $k\in \N$, $\forall s < t$, $\forall r < s$,
\begin{enumerate}[label=(\Alph*)]
    \item $\forall p \in \{0, \frac1L, \ldots, 1\}, q\in (0,1]$,
    \[|\Gamma^{L/2}_{s,r,a,b}(p,q) - \Gamma^L_{s,r,a,b}(p,q)| \leq c/L,\qquad |C^{L/2}_{s,r,a}(p,q) - C^{L}_{s,r,a}(p,q)|\leq c / L.\]
    \item $|\Gamma^{L}_{s,r,a,b}(p,q)|\leq c$, $|C^L_{s,r,a}(p)|\leq c$.
    \item $C^L_{s, r, a}(p)$ is $c$-Lipschitz w.r.t. $p$, and $\Gamma^L_{s,r,a,b}(p,q)$ is $c$-Lipschitz w.r.t. $p$. 
    \item $|\Gamma^L_{s,r,0,1}(p-\frac1L,p+\frac1L)-\Gamma^L_{s,r,0,1}(p-\frac1L,p)|\leq c/L$,
    $|\Gamma^L_{s,r,1,0}(p,p) - \Gamma^L_{s,r,1,0}(p,p-\frac1L)|\leq c/L$.
\end{enumerate}

\paragraph{Remark}
    (A) indicates that $\{\Gamma^{2^k}_{s,r,a,b}\}_k$ and $\{C^{2^k}_{s,r,a}\}_k$ converge.
    We only care about $r<s$ because $C^L_{s,s,a}$ will never be used, and $\Gamma^L_{s,s,a,b}$ is known: for $p\in \{0, \frac1L, \ldots, 1\}$, \[\Gamma^L_{s,s,a,b}(p,q) = \ind[(a=0)\wedge (b=0)\wedge (p\geq q)]+ \ind[(a=1)\wedge (b=1)\wedge (p+1/L\leq q)].\]

\paragraph{Proof for $t$-th step (the forward pass)}
In the following subsections, we will prove inductively on increasing order of all $L > L'$ and $L=2^k$, and increasing order of $p\in \{0, 1/L, \ldots, 1\}$ that $\forall s< t$, 
\begin{enumerate}
    \item[(D0)] $|\Gamma^L_{t,s,0,1}(p,p+\frac2L)-\Gamma^L_{t,s,0,1}(p,p+\frac1L)|\leq c_2\exp(c_1 p)/L$;
    \item[(C0)] For $s < t$,
    $|\Gamma^{L}_{t,s,0,b}(p,q) - \Gamma^{L}_{t,s,0,b}(p-\frac1L,q) |\leq tcc_2 \exp(c_1 (p-\frac1L)) / L$;
    \item[(B0)] $|\Gamma^{L}_{t,s,0,b}(p,q)|\leq c_2 \exp(c_1 (p-\frac{1}{2L}))$;
    \item[(A0)] $|\Gamma^{L/2}_{t,s,0,b}(p,q) - \Gamma^L_{t,s,0,b}(p,q)|\leq c_3c_2 \exp(c_1 (p-\frac{1}{2L})) / L$;
    \item[(C1)] $|C^{L}_{t,s,0}(p+\frac1L) - C^{L}_{t,s,0}(p)|\leq c_4  c_2\exp(c_1 (p-\frac1L))/L$;
    \item[(B1)] $|C^L_{t,s,0}(p+\frac1L)|\leq c_2 \exp(c_1 p)$;
    \item[(A1)] $|C^{L/2}_{t,s,0}(p+\frac1L) - C^L_{t,s,0}(p+\frac1L)|\leq c_5 c_2 \exp(c_1 p) / L,$
\end{enumerate}
where $c_2 =\max\{\xi_t^2 , |\xi_t|\} \exp(c_1/{2L'})$, $c_3=3ct$, $c_4=4t(t+1)c^2+2tc$, $c_5=c_4+1$, $c_1=c^3t(4ct+2c_4+29) + tc(3c_4+14)+c(2c_4+2c)$.

\paragraph{Proof for $t$-th step (the backward pass)} Similar bounds also apply to $\Gamma_{t,s,1,b}$ and $C_{t,s,1}$ by induction on decreasing order of $p$. 

\paragraph{Conclusion} Combining both backward pass and forward pass at time $t$ shows (A)(B)(C)(D) also hold for $s=t$ with a larger (but constant) $c$. Thus, (A)(B)(C)(D) hold for any constant $s$ by induction on training steps. 

\subsubsection{$\Gamma^{L}_{t,s,0,b}(p,q)$ in forward pass (Proof for D0, C0, B0, A0)}
We first consider
\begin{align*} 
     \Gamma^L_{t,r,0,b}\left(p, q\right) = &~\Gamma^L_{t,r,0,b}\left(p-\frac1L, q\right) + \ind[(t=r) \wedge (b=0) \wedge (Lp=\lceil L q\rceil)] \\
     & + \frac1L \sum_{s=0}^{t-1} \Gamma^L_{s,r,1,b}\left(p, q\right)\left(\Gamma^L_{t,s,0,1}\left(p-\frac1L, p\right)-C^L_{t,s,0}\left(p\right)\right).
\end{align*}
\paragraph{(D0) Difference between $\Gamma^L_{t,s,0,1}(p,p+\frac2L)$ and $\Gamma^L_{t,s,0,1}(p,p+\frac1L)$ }

Assume $p\geq 1/L$ ($p=0$ is trivial), 
let $q=p+1/L, q'=p+2/L$, note that $\Gamma_{s,s,1,b}^L(p,q)=\Gamma_{s,s,1,b}^L(p,q')$ since $p+1/L\leq q\leq q'$, so for $r < t$, 
\begin{align*}
    & |\Gamma^L_{t,r,0,b}\left(p, q\right) - \Gamma^L_{t,r,0,b}\left(p, q'\right)|\\
    \leq &~|\Gamma^L_{t,r,0,b}\left(p-\frac1L, q\right) -\Gamma^L_{t,r,0,b}\left(p-\frac1L, q'\right)|  \\
     & + \frac1L \sum_{s=0}^{t-1} |\Gamma^L_{s,r,1,b}\left(p, q\right)-\Gamma^L_{s,r,1,b}\left(p, q'\right)|\left|\Gamma^L_{t,s,0,1}\left(p-\frac1L, p\right)-C^L_{t,s,0}\left(p\right)\right|\\
    \leq &~ c_2\exp(c_1(p-\frac1L)) / L + \frac1L \cdot t \cdot c/L \cdot 2c_2\exp(c_1(p-\frac1L)) \\
    = &~ (1 + 2ct/L) c_2\exp(c_1(p-\frac1L)) /L \leq c_2\exp(c_1p) /L,
\end{align*}
as $c_1\geq 2ct$.

\paragraph{(C0) Lipschitz w.r.t. $p$ } For $r < t$, 
\begin{align*}
  & |\Gamma^{L}_{t,r,0,b}(p,q) - \Gamma^{L}_{t,r,0,b}(p-\frac1L,q)|\\
  = &~ |\frac1L \sum_{s=0}^{t-1} \Gamma^L_{s,r,1,b}\left(p, q\right)\left(\Gamma^L_{t,s,0,1}\left(p-\frac1L, p\right)-C^L_{t,s,0}\left(p\right)\right)|\\
  \leq &~ \frac1L  \sum_{s=0}^{t-1} c\left(c_2\exp(c_1(p-\frac1L))+c_2\exp(c_1(p-\frac1L))\right) \\
  = &~ ctc_2\exp(c_1(p-\frac1L))/L.
\end{align*}

\paragraph{(B0) Bounded} Again assume $p\geq 1/L$ ($p=0$ is trivial because $c_2\geq |\xi_t|\exp(c_1/2L)$), since $|\Gamma^{L}_{t,r,0,b}(p-\frac1L,q)|\leq c_2\exp(c_1(p-\frac1L))$,
we can bound $|\Gamma^{L}_{t,r,0,b}(p,q)|$:
\begin{align*}
    |\Gamma^{L}_{t,r,0,b}(p,q)|\leq&~ c_2\exp(c_1(p-\frac1L))+ ctc_2\exp(c_1(p-\frac1L))/L\\
    = &~ c_2\exp(c_1(p-\frac1L))(1+ct/L)\\
    \leq &~ c_2\exp(c_1(p-\frac{1}{2L})),
\end{align*}
as long as $c_1 \geq 2ct$.

\paragraph{(A0) Difference between $L$ and $L/2$ bounded}

When $p=0$, it is trivial.
When $p=1/L$, it is also trivial by Lipschitz w.r.t. $p$, which results  
\[|\Gamma_{t,r,0,b}^{L/2}\left(p, q\right) - \Gamma_{t,r,0,b}^{L}\left(p, q\right)|\leq 3ctc_2/L\leq c_3 c_2\exp(c_1/2L)/L.\]
When $p\geq 2/L$, since 
\begin{align*}
    \Gamma_{t,r,0,b}^{L/2}\left(p, q\right) = \Gamma_{t,r,0,b}^{L/2}\left(p-\frac2L, q\right) + \frac2L \sum_{s=0}^{t-1} \Gamma_{s,r,1,b}^{L/2}\left(p, q\right)\left(\Gamma_{t,s,0,1}^{L/2}\left(p-\frac2L, p\right)-C_{t,s,0}^{L/2}(p)\right),
\end{align*}
we compare it with $\Gamma_{t,r,0,b}^{L}\left(p, q\right)$ expanded based on its previous two steps
\begin{align*} 
     \Gamma^L_{t,r,0,b}\left(p, q\right) = &~\Gamma^L_{t,r,0,b}\left(p-\frac2L, q\right) + \frac1L \sum_{s=0}^{t-1} \Gamma^L_{s,r,1,b}\left(p, q\right)\left(\Gamma^L_{t,s,0,1}\left(p-\frac1L, p\right)-C^L_{t,s,0}\left(p\right)\right) \\
     & + \frac1L \sum_{s=0}^{t-1} \Gamma^L_{s,r,1,b}\left(p-\frac1L, q\right)\left(\Gamma^L_{t,s,0,1}\left(p-\frac2L, p-\frac1L\right)-C^L_{t,s,0}\left(p-\frac1L\right)\right).
\end{align*}
In order to bridge the two above, namely matching the inputs for $\Gamma$ and $C$, we need a middle term 
\begin{align*} 
    \tilde\Gamma^L_{t,r,0,b}\left(p, q\right) = &~\Gamma^L_{t,r,0,b}\left(p-\frac2L, q\right) + \frac2L \sum_{s=0}^{t-1} \Gamma^L_{s,r,1,b}\left(p, q\right)\left(\Gamma^L_{t,s,0,1}\left(p-\frac2L, p\right)-C^L_{t,s,0}\left(p\right)\right).
\end{align*}

Now we can bound $|\Gamma^L_{t,r,0,b}\left(p, q\right)-\tilde\Gamma^L_{t,r,0,b}\left(p, q\right)|$, and $|\tilde\Gamma^L_{t,r,0,b}\left(p, q\right)-\Gamma^{L/2}_{t,r,0,b}\left(p, q\right)|$ separately, which add up to be the bound for $|\Gamma^L_{t,r,0,b}\left(p, q\right)-\Gamma^{L/2}_{t,r,0,b}\left(p, q\right)|$.
\begin{align*} 
    &~ |\Gamma^L_{t,r,0,b}\left(p, q\right)-\tilde\Gamma^L_{t,r,0,b}\left(p, q\right)| \\
    \leq  &~\frac1L \sum_{s=0}^{t-1} |\Gamma^L_{s,r,1,b}\left(p, q\right)|\left|\Gamma^L_{t,s,0,1}\left(p-\frac1L, p\right)-\Gamma^L_{t,s,0,1}\left(p-\frac2L, p\right)\right|  \\
    & + \frac1L \sum_{s=0}^{t-1} \left|\Gamma^L_{s,r,1,b}\left(p-\frac1L, q\right)\left(\Gamma^L_{t,s,0,1}\left(p-\frac2L, p-\frac1L\right)-C^L_{t,s,0}\left(p-\frac1L\right)\right)\right.\\
    & \qquad\qquad \left.- \Gamma^L_{s,r,1,b}\left(p, q\right)\left(\Gamma^L_{t,s,0,1}\left(p-\frac2L, p\right)-C^L_{t,s,0}\left(p\right)\right)\right|\\
    \leq &~ \frac 1L \cdot  ct\cdot ctc_2\exp(c_1(p-\frac2L)) / L + \frac1L \cdot 2t\cdot c / L \cdot c_2\exp(c_1(p-\frac2L)) \\
    & + \frac1L \cdot ct \cdot c_2\exp(c_1(p-\frac2L)) / L + \frac1L \cdot ct \cdot c_4 c_2\exp(c_1(p-\frac2L))/L\\
    = &~ \frac{c^2t^2+3ct+c_4ct}{L^2}
    \cdot c_2\exp(c_1(p-\frac2L)).
\end{align*}

and
\begin{align*} 
     &~ |\Gamma_{t,r,0,b}^{L/2}\left(p, q\right) - \tilde\Gamma_{t,r,0,b}^{L}\left(p, q\right)| \\
     \leq  &~|\Gamma_{t,r,0,b}^{L/2}\left(p-\frac2L, q\right) - \Gamma_{t,r,0,b}^L\left(p-\frac2L, q\right)|  \\
     & + \frac1L \sum_{s=0}^{t-1} c\left|\Gamma_{t,s,0,1}^{L/2}\left(p-\frac2L, p\right) - \Gamma_{t,s,0,1}^L\left(p-\frac2L, p\right)-C_{t,s,0}^{L/2}\left(p\right)+C_{t,s,0}^L\left(p\right)\right|\\
     & + \frac1L \sum_{s=0}^{t-1} \frac cL\left(|\Gamma^L_{t,s,0,1}\left(p-\frac2L, p\right)|+|C^L_{t,s,0}\left(p\right)|\right) \\
     \leq &~ \frac 1L \cdot (c_3c_2 \exp(c_1(p-\frac1L)) +   ct (c_3+c_5) c_2 \exp(c_1(p-\frac1L)) / L + 2t \cdot \frac cL \cdot  c_2 \exp(c_1(p-\frac1L)) )\\
     \leq &~ \frac {c_3+ct(c_3+c_5+2)/L}L \cdot c_2 \exp(c_1(p-\frac1L)).
\end{align*}
In sum, as $c_1\geq \frac{2(c_3+c_5+ct+c_4+5)}3$,
\begin{align*}
    |\Gamma_{t,r,0,b}^{L/2}\left(p, q\right) - \Gamma_{t,r,0,b}^{L}\left(p, q\right)|\leq &~ \frac{c_3+ct(c_3+c_5+ct+c_4+5)/L}{L} c_2 \exp(c_1(p-\frac1L))\\
    \leq &~ c_2 \exp(c_1(p-\frac1{2L})) /L.
\end{align*} 

\subsubsection{$C_{t,s,0}(p+\frac1L)$ in forward pass (Proof for C1, B1, A1)}

Now consider $C^L_{t,s,0}$. By expanding 
\[C^L_{t,s,0}(p+\frac1L)=\sum_{t'=-1}^{t}\sum_{s'=-1}^{s}\sum_{b\in\{0,1\}}\int_0^1 \Gamma^L_{t,t',0,b}(p,q) C^L_{t',s',b}(q)\Gamma^L_{s,s',0,b}(p,q) \dd q,\]
we will have 
\begin{align*}
    C^L_{t,s,0}(p+\frac1L)=& \sum_{t'=-1}^{t-1}\sum_{s'=-1}^{s}\sum_{b\in\{0,1\}}\int_0^1 \Gamma^L_{t,t',0,b}(p,q) C^L_{t',s',b}(q)\Gamma^L_{s,s',0,b}(p,q) \dd q \\
    & + \sum_{s'=0}^{s}\int_0^{p} C^L_{t,s',0}(q)\Gamma^L_{s,s',0,0}(p,q) \dd q.
\end{align*}
\paragraph{(C1) Lipschitz} Since $C^L_{t',s',b}$ and $\Gamma^L_{s,s',0,b}$ are bounded and Lipschitz,
\begin{align*}
    & |C^L_{t,s,0}(p+\frac1L) - C^L_{t,s,0}(p)|\\
    \leq &~ \sum_{t'=-1}^{t-1}\sum_{s'=-1}^{s}\sum_{b\in\{0,1\}}\int_0^1 |\Gamma^L_{t,t',0,b}(p,q) - \Gamma^L_{t,t',0,b}(p-\frac1L,q)| \cdot c^2 \dd q \\
    & + \sum_{t'=-1}^{t-1}\sum_{s'=-1}^{s}\sum_{b\in\{0,1\}}\int_0^1 |\Gamma^L_{t,t',0,b}(p-\frac1L,q)| \cdot c\cdot \frac cL \dd q \\
    & + \sum_{s'=0}^{s} \frac1L \cdot |C^L_{t,s',0}(p)\Gamma^L_{s,s',0,0}(p,p)| \\
    & + \sum_{s'=0}^{s}\int_0^{p-\frac1L} |C^L_{t,s',0}(q)|\cdot \frac cL \dd q.\\
    \leq &~ 1/L \cdot (2t(s+1)  \cdot ctc_2\exp(c_1 (p-\frac1L))\cdot c^2 + 2t(s+1) c_2\exp(c_1 (p-\frac1L)) \cdot c^2\\
    & +   s \cdot  c_2\exp(c_1 (p-\frac1L))\cdot c + s \cdot c_2\exp(c_1 (p-\frac1L)) \cdot c) \\
    = &~ (4t(s+1)c^2+2sc)/L\cdot  c_2\exp(c_1 (p-\frac1L))\\
    \leq &~ c_4 c_2\exp(c_1 (p-\frac1L)) / L.
\end{align*}
\paragraph{(B1) Bounded} Since $|C^L_{t,s,0}(p)|\leq c_2\exp(c_1 (p-\frac1L))$, we bound $C^L_{t,s,0}(p+\frac1L)$ as:
\[|C^L_{t,s,0}(p+\frac1L)|\leq c_2\exp(c_1 (p-\frac1L)) \cdot (1 + c_4/L)\leq c_2\exp(c_1 p),\]
as long as $c_1 \geq c_4$.
\paragraph{(A1) Difference between $L$ and $L/2$ bounded} It is easy to see that for $p=0$, \[C^L_{t,s,0}(p+\frac1L) - C^{L/2}_{t,s,0}(p+\frac1L)=0,\]
we will prove that for $p\in \{2/L, 4/L, \ldots, 1\}$, 
\[|C^L_{t,s,0}(p+\frac1L) - C^{L/2}_{t,s,0}(p+\frac1L)|\leq c_2\exp(c_1p)/L.\]
Then by (C1), for $p\in \{1/L, 3/L, \ldots, 1-1/L\}$,
\begin{align*}
|C^L_{t,s,0}(p+\frac1L) - C^{L/2}_{t,s,0}(p+\frac1L)|\leq &~c_2\exp(c_1(p-\frac1L))/L+c_4c_2\exp(c_1(p-\frac1L))/L\\
\leq &~(c_4+1)c_2\exp(c_1p)/L \\
=&~ c_5c_2\exp(c_1p)/L.
\end{align*}
Suppose $p\in \{2/L, 4/L, \ldots, 1\}$,
we compare $C^L_{t,s,0}(p+\frac1L) - C^L_{t,s,0}(p-\frac1L)$ and $C^{L/2}_{t,s,0}(p+\frac1L) - C^{L/2}_{t,s,0}(p-\frac1L)$. Intuitively, both of them are $\cO(1/L)$, and their difference is $\cO(1/L^2)$.
In particular, both of them can be written into four parts:
\begin{align*}
    &~C^L_{t,s,0}(p+\frac1L) - C^L_{t,s,0}(p-\frac1L) \\
    =&~\sum_{t'=-1}^{t-1}\sum_{s'=-1}^s \sum_{b\in \{0, 1\}} \int_0^1 \left(\Gamma^L_{t,t',0,b}(p,q) - \Gamma^L_{t,t',0,b}(p-\frac2L,q)\right) C^L_{t',s',b}(q)\Gamma^L_{s,s',0,b}(p,q) \dd q & (\cE_1^L) \\
    & + \sum_{t'=-1}^{t-1}\sum_{s'=-1}^s \sum_{b\in \{0, 1\}} \int_0^1 \Gamma^L_{t,t',0,b}(p-\frac2L,q) C^L_{t',s',b}(q)\left(\Gamma^L_{s,s',0,b}(p,q) -\Gamma^L_{s,s',0,b}(p-\frac2L,q)\right) \dd q & (\cE_2^L) \\
    & + \sum_{s'=0}^s \int_{p-\frac2L}^p C^L_{t,s',0}(q) \Gamma_{s,s',0,0}^L(p,q)\dd q & (\cE_3^L)\\
    & + \sum_{s'=0}^s \int_0^{p-\frac2L} C^L_{t,s',0}(q) (\Gamma_{s,s',0,0}^L(p,q)-\Gamma_{s,s',0,0}^L(p-\frac2L,q))\dd q & (\cE_4^L)
\end{align*}
and $C^{L/2}_{t,s,0}(p+\frac1L) - C^{L/2}_{t,s,0}(p-\frac1L) = \cE_1^{L/2} + \cE_2^{L/2}+\cE_3^{L/2}+\cE_4^{L/2}$ where 
$\cE_i^{L/2}$ is defined in the same way as $\cE_i^L$ but with $C^{L/2}$ and $\Gamma^{L/2}$ instead of $C^L$ and $\Gamma^L$. Next we bound $|\cE_i^L-\cE_i^{L/2}|$ one by one:

\begin{enumerate}
    \item The only hard part to bound in $|\cE_i^L-\cE_i^{L/2}|$ is 
    \[|\Gamma^L_{t,t',0,b}(p,q) - \Gamma^L_{t,t',0,b}(p-\frac2L,q) - (\Gamma^{L/2}_{t,t',0,b}(p,q) - \Gamma^{L/2}_{t,t',0,b}(p-\frac2L,q))|.\]
    By almost the same proof of (A0), 
    \begin{align*}
        &|\Gamma^L_{t,t',0,b}(p,q) - \Gamma^L_{t,t',0,b}(p-\frac2L,q) - (\Gamma^{L/2}_{t,t',0,b}(p,q) - \Gamma^{L/2}_{t,t',0,b}(p-\frac2L,q))|\\
        \leq &~ \frac{ct(c_3+c_5+ct+c_4+5)}{L^2} c_2 \exp(c_1(p-\frac1L)).
    \end{align*}
    Then we have 
    \begin{align*}
        & |\cE_1^L-\cE_1^{L/2}| / (2t(s+1))\\
        \leq &~ \frac{ct(c_3+c_5+ct+c_4+5)}{L^2} c_2 \exp(c_1(p-\frac1L))\cdot c\cdot c\\
        & + 4ctc_2\exp(c_1(p-\frac1L))/L\cdot c/L \cdot c \\
        & + 4ctc_2\exp(c_1(p-\frac1L))/L\cdot c\cdot c/L\\
        \leq &~ \frac{c^3t(c_3+c_5+ct+c_4+13)}{L^2} c_2 \exp(c_1(p-\frac1L))
    \end{align*}
    \item Bounding $|\cE_2^L-\cE_2^{L/2}|$ is similar to $|\cE_1^L-\cE_1^{L/2}|$, where we first bound
    \[|\Gamma^L_{s,s',0,b}(p,q) - \Gamma^L_{s,s',0,b}(p-\frac2L,q) - (\Gamma^{L/2}_{s,s',0,b}(p,q) - \Gamma^{L/2}_{s,s',0,b}(p-\frac2L,q))|        \leq 9c^2t/L^2.\]
    Then we have 
    \begin{align*}
        & |\cE_2^L-\cE_2^{L/2}| / (2t(s+1))\\
        \leq &~ c_3c_2\exp(c_1(p-\frac2L))/L\cdot c\cdot 2c/L\\
        & + c_2\exp(c_1(p-\frac2L)) \cdot c / L \cdot 2c/L \\
        & + c_2\exp(c_1(p-\frac2L)) \cdot c \cdot 9c^2t/L^2\\
        \leq &~ \frac{c^2(2c_3+2+9ct)}{L^2} c_2 \exp(c_1(p-\frac2L)).
    \end{align*}
    \item For $|\cE_3^L-\cE_3^{L/2}|$, we first simplify \[\cE_3^{L/2}=\frac2L\sum_{s'=0}^s C_{t,s',0}^{L/2}(p)\Gamma_{s,s',0,0}^{L/2}(p,p), \] and 
    \[\cE_3^{L}=\frac1L\sum_{s'=0}^s C_{t,s',0}^{L}(p)\Gamma_{s,s',0,0}^{L}(p,p) + C_{t,s',0}^{L}(p-\frac1L)\Gamma_{s,s',0,0}^{L}(p,p-\frac1L).\]
    Again, we introduce an intermediate term 
    \[\tilde\cE_3^L=\frac2L\sum_{s'=0}^s C_{t,s',0}^{L}(p)\Gamma_{s,s',0,0}^{L}(p,p).\]
    Then we can bound
    \begin{align*}
        & |\cE_3^L-\cE_3^{L/2}|\\
        \leq &~ |\cE_3^L-\tilde\cE_3^{L}|+|\tilde\cE_3^L-\cE_3^{L/2}|\\
        \leq &~ \frac tL (c_4c_2\exp(c_1(p-\frac2L))/L \cdot c + c_2\exp(c_1(p-\frac2L)) \cdot c/L)\\
        & + \frac{2t}L (c_5c_2\exp(c_1(p-\frac1L))/L c + c_2\exp(c_1(p-\frac1L)) \cdot c/L)\\
        \leq &~ \frac{tc(c_4+1+2c_5+2)}{L^2}c_2\exp(c_1(p-\frac1L)).
    \end{align*}
    \item For $|\cE_4^L-\cE_4^{L/2}|$, we use \[|\Gamma^L_{s,s',0,b}(p,q) - \Gamma^L_{s,s',0,b}(p-\frac2L,q) - (\Gamma^{L/2}_{s,s',0,b}(p,q) - \Gamma^{L/2}_{s,s',0,b}(p-\frac2L,q))|        \leq 9c^2t/L^2,\] which is used in $|\cE_2^L-\cE_2^{L/2}|$. Finally,
    \begin{align*}
        & |\cE_4^L-\cE_4^{L/2}| / t\\
        \leq &~ c_4c_2\exp(c_1(p-\frac2L))/L\cdot 2c/L\\
        & + c_2\exp(c_1(p-\frac2L)) \cdot 9c^2t/L^2\\
        \leq &~ \frac{c(2c_4+9ct)}{L^2} c_2 \exp(c_1(p-\frac2L)).
    \end{align*}
\end{enumerate}

In sum, 
\begin{align*}
    & |C^L_{t,s,0}(p+\frac1L) - C^L_{t,s,0}(p-\frac1L) -C^{L/2}_{t,s,0}(p+\frac1L) + C^{L/2}_{t,s,0}(p-\frac1L)| \\
    \leq &~ \frac{c^3t(4ct+2c_4+14) + c^2(2+15ct) + tc(3c_4+5)+c(2c_4+9ct)}{L^2}\cdot c_2 \exp(c_1(p-\frac1L))\\
    = &~ \frac{c^3t(4ct+2c_4+29) + tc(3c_4+14)+c(2c_4+2c)}{L^2}\cdot c_2 \exp(c_1(p-\frac1L)).
\end{align*}
Therefore, since $c_1= c^3t(4ct+2c_4+29) + tc(3c_4+14)+c(2c_4+2c)$,
\begin{align*}
    & |C^L_{t,s,0}(p+\frac1L) -C^{L/2}_{t,s,0}(p+\frac1L)| \\
    \leq &~ |C^L_{t,s,0}(p-\frac1L) - C^{L/2}_{t,s,0}(p-\frac1L)| + c_1 / L^2 \cdot c_2 \exp(c_1(p-\frac1L))\\
    \leq&~ (1+c_1/L)c_2 \exp(c_1(p-\frac1L)) /L\\
    \leq&~ c_2\exp(c_1p)/L.
\end{align*}

\section{Classification of Depthwise Parametrizations in Linear Case}\label{sec:linearsgd_classification}

We discuss the classification results on the linear residual networks with SGD training and give rigorous proofs for the claims in this simplified setting. Recall the linear residual networks:
\[
\forall l \in [L], x^l = x^{l-1} + a L^{-\alpha} h^l, 
\]
where $h^l=W^l x^l$, and the effective learning rate of $W^l$ is $\eta n^{-1} L^{-\gamma}$. Without loss of generality, we assume $\eta=a=1$. 

\subsection{Initialization}
At initialization, we have 
\[
\ket{x_0^l} = \ket{x_0^{l-1}} + L^{-\alpha} \ket{h^l_0},
\]
where \[\ket{h_0^l}=\ket{W_0^l x_0^{l-1}}=\hatket{W_0^l x_0^{l-1}}.\]
Since $\ket{x_0^{l-1}}$ is independent from $\hatket{W_0^l x_0^{l-1}}$,
we have 
\[\braket{x_0^l}{x_0^l} = \braket{x_0^{l-1}}{x_0^{l-1}} + L^{-2\alpha} \braket{h^l_0}{h^l_0} = \braket{x_0^{l-1}}{x_0^{l-1}} + L^{-2\alpha} \braket{x_0^{l-1}}{x_0^{l-1}} =  (1+L^{-2\alpha})\braket{x_0^{l-1}}{x_0^{l-1}}.\]
Using this recursion, we can write 
\[\braket{x_0^l}{x_0^l} = (1+L^{-2\alpha})^l \braket{x_0^0}{x_0^0}.\]
Therefore, $\braket{x_0^L}{x_0^L}=\Theta(1)$ iff $\alpha\geq 1/2$, otherwise $(1+L^{-2\alpha})^L\approx e^{L^{-2\alpha+1}}$ explodes with large $L$.

A similar argument stands for $h^l_0$ and $f_0$. Therefore, we have proved \Cref{clm:stability_init}.

Similarly, we can get the stability of the first backward pass, i.e., $\tilde \del x_0^l=\Theta(1)$ for $\alpha\geq 1/2$. Given $\alpha \geq 1/2$, we can also settle the size of $\tilde \del h_0$ that
\[\tilde \del h_0^l = \Theta(L^{-\alpha}), \]
which implies 
\[\Delta W_1^l = L^{-\gamma+\alpha}\cdot \tilde \del  h_0^l \otimes x_0^{l-1}.\]

\subsection{After the first step of gradient update}

Now we look at the second forward pass, and assume the input is the same, i.e., $\ket{x_1^0}=\ket{x_0^0}$, we have
\[\ket{x_1^l} = \ket{x_1^{l-1}} + L^{-\alpha} (\hatket{W_0^l x_1^{l-1}} + \dotket{W_0^l x_1^{l-1}}+ \oplim{\Delta W_1^l} x_1^{l-1}\ra)\]
where $\oplim{\Delta W_1^l}= -L^{-\gamma} \Tilde{\ket{\tilde \delta h_0^l}}\bra{x_0^{l-1}}=-L^{-\gamma} \ket{\tilde \delta x_0^l}\bra{x_0^{l-1}}$, and $\Tilde{\ket{\tilde \delta h_0^l}}\odefeq L^{\alpha} \ket{\tilde \del h_0^l}$ is the normalized version of $\ket{\tilde \del h_0^l}$, which happens to equal to $\ket{\tilde \del x_0^l}$. By the definition of $\hatket{W_0^l x_1^{l-1}}$ and $\dotket{W_0^l x_1^{l-1}}$, we get a similar formula to the Depth-$\mu$P case:
\[\ket{x_1^l} = \ket{x_1^{l-1}} + L^{-\alpha} \hatket{W_0^l x_1^{l-1}} + L^{-\alpha} \ket{\tilde \del x_0^{l}} \left(\frac{\partial \ket{x_1^{l-1}}}{\partial \hatket{W_0^{l\top} \tilde \del x_0^{l}}} - L^{-\gamma} \braket{x_0^{l-1}}{x_1^{l-1}} \right).\]
Now we write $b^l=L^{\gamma} \frac{\partial \ket{x_1^{l-1}}}{\partial \hatket{W_0^{l\top} \tilde \del x_0^{l}}}$ and $c^l=- \braket{x_0^{l-1}}{x_1^{l-1}}$, then
\[\ket{x_1^{l}}=\ket{x_1^{l-1}} + L^{-\alpha} \hatket{W_0^l x_1^{l-1}} + L^{-\alpha-\gamma}(b^l + c^l)\ket{\tilde \del x_0^l}.\]
By expanding $\ket{\tilde \del x_0^{l-1}}=\ket{\tilde \del x_0^{l}} + L^{-\alpha} \hatket{W_0^{l\top}\tilde \del x_t^l}=\ket{\tilde \del x_0^L} + \sum_{m=l}^L L^{-\alpha} \hatket{W_0^{m\top}\tilde \del x_t^m}$, 
we have 
\begin{align}
    \ket{x_1^{l}}=&~\ket{x_1^{l-1}} + L^{-\alpha} \hatket{W_0^l x_1^{l-1}} + L^{-\alpha-\gamma}(b^l + c^l)\left(\ket{\tilde \del x_0^L} + \sum_{m=l+1}^L L^{-\alpha} \hatket{W_0^{m\top}\tilde \del x_0^m}\right) \nonumber \\
    =&~ \ket{x_1^{0}} + \sum_{m=1}^l L^{-\alpha} \hatket{W_0^m x_1^{m-1}} + \sum_{m=1}^l L^{-\alpha-\gamma} (b^m +c^m)\ket{\tilde \del x_0^L} \nonumber\\
    & + \sum_{m=2}^L L^{-\alpha-\gamma} \sum_{l'=1}^{\min\{m-1,l\}} (b^{l'}+c^{l'}) L^{-\alpha} \hatket{W_0^{m\top}\tilde \del x_0^m}.\label{eq:x_1l}
\end{align}
Note the four terms in \cref{eq:x_1l} are independent of each other.

Now it is easy to compute $c^l$ because only the first two terms in \cref{eq:x_1l} have correlation with $x_0^l$:
\[c^l =c^{l-1} (1+ L^{-2\alpha})=\Theta(1)\]
with $\alpha \geq 1/2$.
For $b^l$, we have the following recursive formula:
\[b^{l+1}=L^{-2\alpha} \sum_{m=1}^l (b^l+c^l)=\Theta(l\cdot L^{-2\alpha}).\]
\paragraph{Stable during training and nontrivial.} Finally, we can reason about the $\mathring f_1$ (note $\mathring f_0=0$, so $\Delta \mathring f_1=\mathring f_1$), which indicates whether the parametrization is stable during the first step\footnote{We need $\Delta x$ and $\Delta h$ for stability, but they are similar to $\Delta \mathring f_1$.}, and whether the parametrization is nontrivial for the first step:
\[\mathring f_1 = \braket{nV}{x_1^L} = \sum_{m=1}^L L^{-\alpha-\gamma} (b^m+c^m) \chi_0=\Theta(L^{1-\alpha-\gamma}).\]
Therefore, we have proved \Cref{clm:stable_non_trivial} that the parametrization is stable during training iff $\alpha + \gamma \geq 1$, and is nontrivial iff $\alpha + \gamma \leq 1$.

\paragraph{Faithfulness.} Although there is no activation in the linear case, we still prove \Cref{clm:faithful} to enlighten the proof of the general case. 

At the initialization, $h_0^l$ and $x_0^{l-1}$ have the same size, therefore, faithfulness is equivalent to stability, which means it happens iff $\alpha \geq 1/2$.

During training, we can expand $\ket{h_1^l}$ in a similar way to \cref{eq:x_1l} as
\[\ket{h_1^l}=\hatket{W_0^l x_1^{l-1}} + L^{-\gamma} (b^l+c^l)\left(\ket{\tilde \del x_0^L} + \sum_{m=l+1}^L L^{-\alpha} \hatket{W_0^{m\top} \tilde \del x_0^m}\right)=\Theta(1+L^{-\gamma}).\]
Therefore, it is faithful iff $\gamma\geq 0$. It is equivalent to $\alpha\leq 1$ because we have $\alpha+\gamma=1$.

\paragraph{Feature diversity exponent.} To simplify the analysis, we assume that $\epsilon L$ is always an integer. We first expand $x_1^{l+\epsilon L}-x_1^l$ 
\begin{align}
    \ket{x_1^{l+\epsilon L}} - \ket{x_1^{l}}=&~ \sum_{m=l+1}^{l+\epsilon L} L^{-\alpha} \hatket{W_0^m x_1^{m-1}} + \sum_{m=l+1}^{l+\epsilon L} L^{-\alpha-\gamma} (b^m +c^m)\ket{\tilde \del x_0^L} \nonumber\\
    & + \sum_{m=2}^L L^{-\alpha-\gamma} \sum_{l'=\min\{m-1,l\}+1}^{\min\{m-1,l+\epsilon L\}} (b^{l'}+c^{l'}) L^{-\alpha} \hatket{W_0^{m\top}\tilde \del x_0^m}. \nonumber
\end{align}
With $\alpha+\gamma=1$, it is clear that the first term is $\Theta(L^{-\alpha} \sqrt{\epsilon L})=\Theta(\epsilon^{1/2}L^{-\alpha+1/2})$, 
the second term has size $\Theta(\epsilon)$, 
and the third term has size $\Theta(\sqrt L\cdot \epsilon L^{-\alpha})=\Theta(\epsilon L^{-\alpha+1/2})$. Therefore, there are only two cases here: if $\alpha = 1/2$, the overall size is $\Theta(\epsilon^{1/2}+\epsilon)=\Theta(\epsilon^{1/2})$; if $\alpha > 1/2$, the first and the third term vanish as $L\to\infty$, so the overall size is $\Theta(\epsilon)$. In sum, we have proved \Cref{clm:redundant,clm:depth_mup}.

\paragraph{Layerwise linearization.} \Cref{clm:layerwise-linearization} is trivial in this simplified setting because layerwise linearization is always true for linear nets. To enlighten the proof of the general case, we recap that $\oplim{\Delta W_1^l} x_1^{l-1} \ra = L^{-\gamma} c^l \ket{\tilde \del x_0^l}=\Theta(L^{-\gamma})$, which is much smaller than $\ket{W_0^l x_1^{l-1}}=\Theta(1)$ when $\gamma > 0$. If there were an activation function, the linearization would bring an error of $o(L^{-\gamma})$ in $h_1^l$, which means an error of $o(L^{-\gamma-\alpha})=o(L^{-1})$ to $x_1^l$.

\subsection{Beyond one step}
The argument above is in general tracking the derivatives and covariance, in other words, $\Gamma$ and $C$ in the Depth-$\mu$P case. 

Now we generalize \Cref{lemma:finite_depth_gamma}, and obtain the following recursion for $\Gamma$ and $C$
\begin{align*} 
     \Gamma_{t,r,0,b}\left(\frac l L, q\right) = &~\Gamma_{t,r,0,b}\left(\frac{l-1}L, q\right) + L^{1/2-\alpha}\ind_{[(t=r) \wedge (b=0) \wedge (l=\lceil L q\rceil)]} \\
     & + L^{-\alpha-\gamma} \sum_{s=0}^{t-1} \Gamma_{s,r,1,b}\left(\frac l L, q\right)\left(L^{\gamma-1/2}\Gamma_{t,s,0,1}\left(\frac{l-1}L, \frac l L\right)-C_{t,s,0}\left(\frac l L\right)\right).
    \end{align*}
    \begin{align*} 
     \Gamma_{t,r,1,b}\left(\frac {l-1} L, q\right) = &~\Gamma_{t,r,1,b}\left(\frac{l}L, q\right) + L^{1/2-\alpha} \ind_{[(t=r) \wedge (b=1) \wedge (l=\lceil L q\rceil)]} \\
     & + L^{-\alpha-\gamma} \sum_{s=0}^{t-1} \Gamma_{s,r,0,b}\left(\frac {l-1} L, q\right)\left(L^{\gamma -1/2}\Gamma_{t,s,1,0}\left(\frac{l}L, \frac l L\right)-C_{t,s,1}\left(\frac l L\right)\right).
     \end{align*}
     \[C_{t,s,a}(p)=\sum_{t'=-1}^{t}\sum_{s'=-1}^{s}\sum_{b\in\{0,1\}}\int_0^1 \Gamma_{t,t',a,b}(l/L,q) C_{t',s',b}(q)\Gamma_{s,s',a,b}(l/L,q) \dd q,\] where $l=\lceil Lp\rceil -1$ if $a=0$, and  $l=\lceil Lp\rceil$ if $a=1$.

Then all the claims can be reasoned by tracking the order of $\Gamma$ and $C$.
\paragraph{Distinguish parametrizations with $\alpha+\gamma=1$ and $\alpha\leq 1$. } The parametrizations with $\alpha+\gamma=1$ and $\alpha\leq 1$ are all nontrivial, stable, and faithful. However, there is a large gap between $\alpha=1/2$ (Depth-$\mu$P) and $\alpha > 1/2$ in terms of the difficulty of tracking $\Gamma$ and $C$. For $\alpha > 1/2$, we can see that
$C_{t,s,a}=\Theta(1)$, $\Gamma_{t,-1,a,b}=\Theta(1)$ and $\Gamma_{t,s,a,b}=o(1)$ for $s\geq 0$. In this case, we can simplify the recursion by ignoring $\Gamma_{t,s,a,b}$ with $s\geq 0$: 
\begin{align*} 
     \Gamma_{t,-1,0,b}\left(\frac l L\right) \approx &~\Gamma_{t,-1,0,b}\left(\frac{l-1}L\right) - \frac1L \sum_{s=0}^{t-1} \Gamma_{s,-1,1,b}\left(\frac l L\right)C_{t,s,0}\left(\frac l L\right).
    \end{align*}
    \begin{align*} 
     \Gamma_{t,-1,1,b}\left(\frac {l-1} L\right) \approx &~\Gamma_{t,-1,1,b}\left(\frac{l}L\right)  - \frac1L \sum_{s=0}^{t-1} \Gamma_{s,-1,0,b}\left(\frac {l-1} L\right)C_{t,s,1}\left(\frac l L\right).
     \end{align*}
     \[C_{t,s,a}(p)\approx\sum_{b\in\{0,1\}} \Gamma_{t,-1,a,b}(l/L) \Gamma_{s,-1,a,b}(l/L),\] where $l=\lceil Lp\rceil -1$ if $a=0$, and  $l=\lceil Lp\rceil$ if $a=1$. Note $\Gamma_{t,-1,a,b}(p,q)$ is simplified to a function that only depends on $p$ because $\Gamma_{t,-1, a,b}(p, q)$ is constant when fixing $p$. 

This simplification means the randomness in any $W_0^l$ does not have an effect on the dynamics in the infinite depth limit --- the complicated functional integrals for $\alpha=1/2$ in \Cref{prop:infinite_depth_linear} are simplified to be ODEs when $\alpha >1/2$. This ODE dynamic also directly implies that the feature diversity exponent is 0 for $\alpha > 1/2$.

\section{Nonlinear Depth-$\mu$P Limit}

When the nonlinearity $\phi$ is nontrivial, the distribution of the final representations $x^L_s$ may be highly non-Gaussian because of complex compositions of $\phi$ and $\phi'$, as one would suspect from known examples of large width limits.
This is indeed the case for finite depth $L$. But in fact, when $L \to \infty$, $\{x^L_s\}_{s\ge 0}$ becomes a Gaussian process again!

The kernel of the limiting GP can be computed in a similar way as in the linear case:
\begin{definition}\label{defn:nonlinearGammaC}
    Define $\Gamma$ and $C$ recursively by
    \begin{align*}
        \Gamma_{t,r,0,b}(p, q)&=\ind_{[(t=r) \wedge (b=0) \wedge (p\geq q)]} \\ 
        &+ \int_0^p \sum_{s=0}^{t-1} V_{\phi'}[C]_{t,s,0}(p')\Gamma_{s,r,1,b}(p',q)\cdot (\Gamma_{t,s,0,1}(p', p') - C_{t,s,0}(p'))\dd p'; \\
           \Gamma_{t,r,1,b}(p, q) &= \ind_{[(t=r) \wedge (b=1) \wedge (p\leq q)]} \\
           &+ \int_p^1 \sum_{s=0}^{t-1} \Gamma_{s,r,0,b}(p',q) \cdot (V_{\phi_c|\phi'}[C]_{t,s,0}\Gamma_{t,s,1,0}(p',p') - V_{\phi'}[C]_{t,s,0}(p') C_{t,s,1}(p'))\dd p';\\
           C_{t,s,a}(p)&=
           \sum_{b\in\{0,1\}}\int_0^1 \Gamma_{t,t',a,b}(p,q) V_{\phi_c} [C]_{t',s',b}(q)\Gamma_{s,s',a,b}(p,q) \dd q.
    \end{align*}
\end{definition}

Here
\begin{align*}
    V_{\phi_c}[C]_{t,s,b}(p) &\defeq \EV \MS(\phi(z)) \MS(\phi(y)) \\ 
    V_{\phi'}[C]_{t,s,b}(p) &\defeq \EV \phi'(z) \phi'(y) \\
    V_{\phi_c | \phi'}[C]_{t,s,b}(p) &\defeq \EV \MS(\phi'(z)) \phi'(y) \\
\end{align*}
where $(z, y) \sim \Gaus(0, C^b(p)|_{\{t,s\}})$.

Then 
\begin{claim}\label{claim:nonlineardepthmuplimit}
    For sufficiently smooth nonlinearity $\phi$, in the $L \to \infty$ limit, the kets $\ket{x^L_s}, s = 0, 1, \ldots,$ converge in distribution as a zero-mean Gaussian process with kernel
    \[\braket{x^L_s}{x^L_t} = C_{t,s,1}(1).\]
    as defined in \Cref{defn:nonlinearGammaC}.
    Thus, for each fixed neuron index $\alpha$, the collection $\{x^L_{\alpha s}\}_{s\ge 0}$ converges in distribution to a zero-mean Gaussian process with kernel $C_{t,s,1}(1)$ in the $n\to\infty$ then $L\to\infty$ limit.
\end{claim}
We frame this as a claim because we do not want to get into the details of what ``sufficiently smooth nonlinearity'' means here, nor do we give a proof. Instead, we give an intuitive justification.

\begin{proof}[Heuristic justification of \Cref{claim:nonlineardepthmuplimit}]
First, in Depth-$\mu$P, we can Taylor expand each block
$$\ket{\phi(W^l_t x^{l-1}_t)} = \phi(\hatket{W^l_0 x^{l-1}_t}) + \phi'(\hatket{W^l_0 x^{l-1}_t}) [\ket{\bar\Delta W^l_t x^{l-1}_t} + \dotket{W^l_0 x^{l-1}_t}] + \cO(L^{-1}).$$
The remainder term thus contributes $\cO(L^{-3/2})$ after accounting for the $L^{-1/2}$ block multiplier. Summing over depth $l \in [0, L]$, by Gronwall's lemma, the remainders of all layers sum up to $\cO(L^{-1/2})$, so we can ignore them.
From here on, we study the linearized blocks $x \mapsto \MS(\phi(W^l_0 x) + \phi'(W^l_0 x) \odot \bar\Delta W^l_t x)$.

Now the key observation is that each of $\ket{x^l_t}$ or $\ket{\delta x^l_t}$ is always equal, up to $\cO(L^{-1/2})$ factor, to a linear combination of $\{\MS \phi(\hatket{W^l_0 x^{l-1}_s}), \hatket{W^{l \trsp}_0 \delta h^l_s}\}_{l, s}$, where each coefficient in this linear combination is $\cO(L^{-1/2})$ and deterministic.
We call such a linear combination a \emph{good linear combination}.
One can see this by an induction argument on $t$.
Indeed, at initialization $t=0$, this claim is trivially true.
Supposing this claim is true for $t$, then it is trivial to see that it remains true at $t+1$ for the backward pass kets $\ket{\delta x^l_t}$.
The only nontrivial part is to show the forward pass for $t+1$.
By the Taylor expansion above, we just need to show that $\phi'(\hatket{W^l_0 x^{l-1}_t}) [\ket{\bar\Delta W^l_t x^{l-1}_t} + \dotket{W^l_0 x^{l-1}_t}]$ is ``well-approximated'' by a good linear combination.
By induction, both $\ket{\bar\Delta W^l_t x^{l-1}_t}$ and $\dotket{W^l_0 x^{l-1}_t}$ are of products between a good linear combination and a term of form $ \phi'(\hatket{W^m_0 x^{m-1}_s})$ for some $s \le t$. Thus $\phi'(\hatket{W^l_0 x^{l-1}_t}) [\ket{\bar\Delta W^l_t x^{l-1}_t} + \dotket{W^l_0 x^{l-1}_t}]$ is of the form $\phi'(\hatket{W^l_0 x^{l-1}_t})\phi'(\hatket{W^m_0 x^{m-1}_s})$ times a good linear combination.
But $\phi'(\hatket{W^l_0 x^{l-1}_t})\phi'(\hatket{W^m_0 x^{m-1}_s})$ only correlates with a single component in this linear combination and is independent with all other components.
Therefore, when summing over depth, the noncorrelated components will experience law of large numbers and one can replace $\phi'(\hatket{W^l_0 x^{l-1}_t})\phi'(\hatket{W^m_0 x^{m-1}_s})$ with its expectation; the correlated components is only a sum of $\cO(L)$ elements each of size $\cO(L^{-3/2})$, so in total they are $o(1)$.
This completes the induction.

In this reasoning, the coefficients of a good linear combination correspond to $\Gamma$ in \Cref{defn:nonlinearGammaC}, and filling in the details of the induction yields the recursive formula in \Cref{defn:nonlinearGammaC}.

Finally, because a good linear combination is a large sum of independent terms, the central limit theorem tells us that $\{\ket{x^l_t}, \ket{\delta x^l_t}\}_{l, t}$ converge in distribution to a Gaussian process.
\end{proof}

\Cref{claim:nonlineardepthmuplimit} is good news for theorists, that we have such a simple form for a fundamental architecture.
At the same time, one may worry that this Gaussian form lacks expressivity.
But in fact, some common architecture or algorithm choices would make the limit non-Gaussian.
For example, the use of adaptive optimizers like Adam or SignSGD.
Or the addition of a nonlinearity before the matrix multiply, i.e., ``prenonlin'', (in addition to the one afterward, ``postnonlin'').

In general, one obtains a stochastic differential equation with McKean-Vlasov elements describing the evolution of $x^l_t$ over depth and time. However, the stochastic integral involved is \emph{not} the usual Ito or Stratonovich integral because the depthwise evolution requires integrating non-adapted process against Brownian motion. Instead, we need to use \emph{Skorohod} integral and the SDE is only defined in the sense of Malliavin calculus.
This is not just a mathematical nitpick; rather, assuming Ito calculus (which amounts to assuming incorrect independence between certain quantities) will lead to the wrong predictions and calculations.
Malliavin calculus is intimately connected to Tensor Programs and we shall develop their relationship as well as the theory of the general infinite-depth limit in a future work.

\section{Heuristics for the proofs in the general case}

The notation in this section is mostly defined in \cref{sec:notations}. The complete notation is defined in \cite{yang2023tensor}.
\subsection{Depth-$\mu$P}\label{heuristics:depth-mup}

Let $\MS(x)=x-\langle x,1\rangle/n=Gx$ where $G=I-11^{\top}/n$,
where $x\in\R^{n}$. Recall the definition of the network and the normalized gradients

\begin{align*}
x^{1} & =U\xi\\
h^{l} & =W^{l}x^{l-1}\\
x^{l} & =x^{l-1}+\frac{1}{\sqrt{L}}G\phi(h^{l})\\
f(\xi) & =V^{\trsp}x^{L}\\
\tilde\del x^{L} & =nV\\
\tilde\del h^{l} & =\phi'(h^{l})\odot(G\tilde\del x^{l})\\
\tilde\del x^{l-1} & =\tilde\del x^{l}+\frac{1}{\sqrt{L}}W^{l\trsp}\tilde\del h^{l}
\end{align*}

where $V=\Theta(1/n)$ coordinatewise, $\tilde\del x^{l}=\Theta(1)$ coordinatewise
and $W^{l}=\Theta(\frac{1}{\sqrt{n}})$ coordinate-wise. 

We also abuse the notation of $G$ and use it as an operator on kets: $G\ket x\odefeq \ket x - \E\ket x$. 

\paragraph{Forward.} Similar to the linear case, one can show that under technical conditions (mostly on the activation function) that the infinite-depth limit of the TP follows the dynamics
\begin{align*}
d\ket{x_{t}^{\lambda}} & =\sqrt{d\lambda}G\phi\left(\oplim{W_{0}^{\lambda}}x_{t}^{\lambda}\ra+\sqrt{d\lambda}\oplim{\widetilde{\bar{\Delta}W_{t}^{\lambda}}}x_{t}^{\lambda}\ra\right)\\
 & =\sqrt{d\lambda}G\phi\left(\oplim{W_{0}^{\lambda}}x_{t}^{\lambda}\ra\right)+d\lambda G\phi'\left(\oplim{W_{0}^{\lambda}}x_{t}^{\lambda}\ra\right)\oplim{\widetilde{\bar{\Delta}W_{t}^{\lambda}}}x_{t}^{\lambda}\ra\\
 & =\sqrt{d\lambda}G\phi\left(\hatoplim{W_{0}^{\lambda}}x_{t}^{\lambda}\ra+\dotoplim{W_{0}^{\lambda}}x_{t}^{\lambda}\ra\right)+d\lambda G\phi'\left(\oplim{W_{0}^{\lambda}}x_{t}^{\lambda}\ra\right)\oplim{\widetilde{\bar{\Delta}W_{t}^{\lambda}}}x_{t}^{\lambda}\ra\\
 & =\sqrt{d\lambda}G\phi\left(\hatoplim{W_{0}^{\lambda}}x_{t}^{\lambda}\ra\right)+d\lambda G\phi'\left(\oplim{W_{0}^{\lambda}}x_{t}^{\lambda}\ra\right)\left(\widetilde{\dotoplim{W_{0}^{\lambda}}x_{t}^{\lambda}\ra}+\oplim{\widetilde{\bar{\Delta}W_{t}^{\lambda}}}x_{t}^{\lambda}\ra\right)
\end{align*}
where $\lambda \in [0,1]$ refers to the fractional layer index ($\lambda$ represents layer index $\lfloor \lambda L\rfloor$ as $L\to \infty$), $t$ refers to the training step, $\oplim{W_{0}^{\lambda}}$ the matrix operator (defined in \Cref{sec:tp}), and the tilde symbol refers to the ``normalized'' version of the object, i.e., multiply the ket with $(d\lambda)^c$ for some $c$ such that the multiplication (normalized ket) is $\Theta(1)$ w.r.t. $L$, and same for the normalized operators. We also simplify $\tilde \del$ to $\del$ if it is already under wider tilde symbol. The first term represents a Gaussian noise.\\

In the linear case, we have
\[
d\ket{x_{t}^{\lambda}}=\sqrt{d\lambda}\left(\hatoplim{W_{0}^{\lambda}}x_{t}^{\lambda}\ra\right)+d\lambda\left(\widetilde{\dotoplim{W_{0}^{\lambda}}x_{t}^{\lambda}\ra}+\oplim{\widetilde{\bar{\Delta}W_{t}^{\lambda}}}x_{t}^{\lambda}\ra\right)
\]
Note
\[
\dotoplim{W_{0}^{\lambda}}x_{t}^{\lambda}\ra=\sqrt{d\lambda}\sum_{s=0}^{t-1}\widetilde{\ket{\delta h_{s}^{\lambda}}}\bra{\nabla_{W_{0}^{\lambda\trsp}\tilde\delta h_{s}^{\lambda}}}x_{t}^{\lambda}\ra=\sqrt{d\lambda}\widetilde{\ket{\delta\hh_{<t}^{\lambda}}}\dbra{W_{0}^{\lambda\trsp}\tilde\delta\hh_{<t}^{\lambda}}x_{t}^{\lambda}\ra
\]
Using multi-vector notation, we write
\begin{align*}
\oplim{\widetilde{\bar{\Delta}W_{t}^{\lambda}}}x_{t}^{\lambda}\ra & =-\eta\widetilde{\ket{\delta\hh_{<t}^{\lambda}}}_{\cchi}\braket{\xx_{<t}^{\lambda}}{x_{t}^{\lambda}} =-\eta \sum_{s<t} \widetilde{\ket{\delta h_{s}^{\lambda}}}_{\chi_s}\braket{x_{s}^{\lambda}}{x_{t}^{\lambda}}\\
\oplim{\bar{\Delta}W_{t}^{\lambda}}x_{t}^{\lambda}\ra & =-\eta\ket{\tilde\delta\hh_{<t}^{\lambda}}_{\cchi}\braket{\xx_{<t}^{\lambda}}{x_{t}^{\lambda}}  =-\eta \sum_{s<t} {\ket{\tilde\delta h_{s}^{\lambda}}}_{\chi_s}\braket{x_{s}^{\lambda}}{x_{t}^{\lambda}}
\end{align*}

\paragraph{Backward.} Similar to the forward prop, we obtain the following dynamics for the infinite-depth TP
\begin{align*}
-d\ket{\tilde\delta x_{\tau}^{\lambda}} & =\sqrt{d\lambda}\oplim{W_{\tau}^{\lambda\trsp}}\phi'(W_{\tau}^{\lambda}x_{\tau}^{\lambda})\odot(G\tilde\delta x_{\tau}^{\lambda})\ra\\
 & =\sqrt{d\lambda}\left(\hatoplim{W_{0}^{\lambda\trsp}}+\dotoplim{W_{0}^{\lambda\trsp}}+\sqrt{d\lambda}\oplim{\bar{\Delta}\widetilde{W_{\tau}^{\lambda}}}^{\dagger}\right)\left[\phi'\left(\hatket{W_{0}^{\lambda}x_{\tau}^{\lambda}}+\sqrt{d\lambda}\dotket{\widetilde{W_{0}^{\lambda}x_{\tau}^{\lambda}}}+\sqrt{d\lambda}\oplim{\widetilde{\bar{\Delta}W_{\tau}^{\lambda}}}x_{\tau}^{\lambda}\ra\right)\ket{G\tilde\delta x_{\tau}^{\lambda}}\right]\\
 & =\sqrt{d\lambda}\hatoplim{W_{0}^{\lambda\trsp}}\left[\phi'(\hatket{W_{0}^{\lambda}x_{\tau}^{\lambda}})\ket{G\tilde\delta x_{\tau}^{\lambda}}\right]+\sqrt{d\lambda}\dotoplim{W_{0}^{\lambda\trsp}}\left[\phi'(\hatket{W_{0}^{\lambda}x_{\tau}^{\lambda}})\ket{G\tilde\delta x_{\tau}^{\lambda}}\right]+d\lambda\oplim{\bar{\Delta}\widetilde{W_{\tau}^{\lambda}}}^{\dagger}\left[\phi'(\hatket{W_{0}^{\lambda}x_{\tau}^{\lambda}})\ket{G\tilde\delta x_{\tau}^{\lambda}}\right]\\
 & \quad+d\lambda\dotoplim{W_{0}^{\lambda\trsp}}\left[\phi''\left(\hatket{W_{0}^{\lambda}x_{\tau}^{\lambda}}\right)\left\{ \dotket{\widetilde{W_{0}^{\lambda}x_{\tau}^{\lambda}}}+\oplim{\widetilde{\bar{\Delta}W_{\tau}^{\lambda}}}x_{\tau}^{\lambda}\ra\right\} \ket{G\tilde\delta x_{\tau}^{\lambda}}\right]
\end{align*}
Here the $(d\lambda)^{3/2}$ term got dropped. The individual terms
can be simplified as follows

\[
\oplim{\bar{\Delta}\widetilde{W_{\tau}^{\lambda}}}^{\dagger}\left[\phi'(\hatket{W_{0}^{\lambda}x_{\tau}^{\lambda}})\ket{G\tilde\delta x_{\tau}^{\lambda}}\right]=-\eta\ketdbra{\xx_{<\tau}^{\lambda}}{\cchi}{\widetilde{\delta\hh_{<\tau}^{\lambda}}}\left[\phi'(\hatket{W_{0}^{\lambda}x_{\tau}^{\lambda}})\ket{G\tilde\delta x_{\tau}^{\lambda}}\right]\approx-\eta\ketdbra{\xx_{<\tau}^{\lambda}}{\cchi}{\widetilde{\delta\hh_{<\tau}^{\lambda}}}\widetilde{\delta h_{\tau}^{\lambda}}\ra
\]
\begin{align*}
\dotoplim{W_{0}^{\lambda\trsp}}\left[\phi'(\hatket{W_{0}^{\lambda}x_{\tau}^{\lambda}})\ket{G\tilde\delta x_{\tau}^{\lambda}}\right] & =\left[\ket{\xx_{<\tau}^{\lambda}}\dbra{W_{0}^{\lambda}\xx_{<\tau}^{\lambda}}+\ket{\xx_{\tau}^{\lambda}}\dbra{W_{0}^{\lambda}\xx_{\tau}^{\lambda}}\right]\left[\phi'(\hatket{W_{0}^{\lambda}x_{\tau}^{\lambda}})\ket{G\tilde\delta x_{\tau}^{\lambda}}\right]\\
 & =\ket{\xx_{<\tau}^{\lambda}}\EV\left[\phi'(\hatket{W_{0}^{\lambda}x_{\tau}^{\lambda}})\frac{\partial\ket{G\tilde\delta x_{\tau}^{\lambda}}}{\partial\hatket{W_{0}^{\lambda}\xx_{<\tau}^{\lambda}}}\right]+\ket{\xx_{\tau}^{\lambda}}\EV\left[\phi''(\hatket{W_{0}^{\lambda}x_{\tau}^{\lambda}})\ket{G\tilde\delta x_{\tau}^{\lambda}}\right]\\
 & =\Theta(\sqrt{d\lambda})
\end{align*}
where the other terms from the product rule drops out because 
\[
\frac{\partial\phi'(\hatket{W_{0}^{\lambda}x_{\tau}^{\lambda}})}{\partial\hatket{W_{0}^{\lambda}\xx_{<\tau}^{\lambda}}}=\frac{\partial\ket{G\tilde\delta x_{\tau}^{\lambda}}}{\partial\hatket{W_{0}^{\lambda}x_{\tau}^{\lambda}}}=0
\]

\subsection{$1/L$ branches}\label{heuristics:inv-depth}

\subsubsection{Forward:}

\begin{align*}
d\ket{x_{t}^{\lambda}} & =d\lambda G\EV\left[\phi\left(\oplim{W_{0}^{\lambda}}x_{t}^{\lambda}\ra+\oplim{\bar{\Delta}W_{t}^{\lambda}}x_{t}^{\lambda}\ra\right)\mid\ket{U_{0},V_{0}}\right]\\
 & =d\lambda G\EV\left[\phi\left(\hatoplim{W_{0}^{\lambda}}x_{t}^{\lambda}\ra+\oplim{\bar{\Delta}W_{t}^{\lambda}}x_{t}^{\lambda}\ra\right)\mid\ket{U_{0},V_{0}}\right]
\end{align*}
where the equality follows because $\ket{x_{t}^{\lambda}}$ is contained
the $\sigma$-algebra of $\ket{U_{0},V_{0}},$so $\dotoplim{W_{0}^{\lambda}}x_{t}^{\lambda}\ra=0$.
Since $\oplim{\bar{\Delta}W_{t}^{\lambda}}\in\sigma(\ket{U_{0},V_{0}})\otimes\sigma(\ket{U_{0},V_{0}})$,
$\oplim{\bar{\Delta}W_{t}^{\lambda}}x_{t}^{\lambda}\ra\in\sigma(\ket{U_{0},V_{0}})$,
and the expectation is really just over $\hatoplim{W_{0}^{\lambda}}x_{t}^{\lambda}\ra$.

\subsubsection{Backward}

\begin{align*}
-d\ket{\tilde\delta x_{\tau}^{\lambda}} & =d\lambda\EV\left[\oplim{W_{\tau}^{\lambda\trsp}}\phi'(W_{\tau}^{\lambda}x_{\tau}^{\lambda})\odot(G\tilde\delta x_{\tau}^{\lambda})\ra\mid\ket{U_{0},V_{0}}\right]\\
 & =d\lambda\EV\left[\oplim{\bar{\Delta}W_{\tau}^{\lambda\trsp}}\phi'(W_{\tau}^{\lambda}x_{\tau}^{\lambda})\odot(G\tilde\delta x_{\tau}^{\lambda})\ra\mid\ket{U_{0},V_{0}}\right]
\end{align*}
Here the $\hatoplim{W_{0}^{\lambda\trsp}}$ and $\dotoplim{W_{0}^{\lambda\trsp}}$
drop out because the former is zero-mean and indepenent from $\ket{U_{0},V_{0}}$
and the latter drops out because $\ket{x_{t}^{\lambda}}$ is contained
the $\sigma$-algebra of $\ket{U_{0},V_{0}}$.

\subsection{$1/L^{\alpha}$ branches, $\alpha\in(1/2,1]$}\label{heuristics:smooth_limits}

\subsubsection{Forward}

\begin{align*}
d\ket{x_{t}^{\lambda}} & =(d\lambda)^{\alpha}G\EV\left[\phi\left(\oplim{W_{0}^{\lambda}}x_{t}^{\lambda}\ra+(d\lambda)^{1-\alpha}\widetilde{\oplim{\bar{\Delta}W_{t}^{\lambda}}}x_{t}^{\lambda}\ra\right)\mid\ket{U_{0},V_{0}}\right]\\
 & =d\lambda\EV\left[\phi'\left(\hatoplim{W_{0}^{\lambda}}x_{t}^{\lambda}\ra\right)\right]G\oplim{\bar{\Delta}W_{t}^{\lambda}}x_{t}^{\lambda}\ra
\end{align*}
because the same reason as above.

\subsubsection{Backward}

\begin{align*}
-d\ket{\tilde\delta x_{\tau}^{\lambda}} & =d\lambda\EV\left[\oplim{W_{\tau}^{\lambda\trsp}}\phi'(W_{\tau}^{\lambda}x_{\tau}^{\lambda})\odot(G\tilde\delta x_{\tau}^{\lambda})\ra\mid\ket{U_{0},V_{0}}\right]\\
 & =d\lambda\EV\left[\oplim{\bar{\Delta}W_{\tau}^{\lambda\trsp}}\phi'(W_{\tau}^{\lambda}x_{\tau}^{\lambda})\odot(G\tilde\delta x_{\tau}^{\lambda})\ra\mid\ket{U_{0},V_{0}}\right]\\
 & =d\lambda\EV\left[\phi'(\hatket{W_{0}^{\lambda}x_{\tau}^{\lambda}})\right]\oplim{\bar{\Delta}W_{\tau}^{\lambda\trsp}}G\tilde\delta x_{\tau}^{\lambda}\ra
\end{align*}

Here, 
\[
\ket{\phi'(W_{\tau}^{\lambda}x_{\tau}^{\lambda})\odot(G\tilde\delta x_{\tau}^{\lambda})}\equiv\EV\left[\phi'(\hatket{W_{0}^{\lambda}x_{\tau}^{\lambda}})\right]G\ket{\tilde\delta x_{\tau}^{\lambda}}+(d\lambda)^{1-\alpha}\EV\left[\phi''(\hatket{W_{0}^{\lambda}x_{\tau}^{\lambda}})\right]G\oplim{\bar{\Delta}W_{\tau}^{\lambda\trsp}}\tilde\delta x_{\tau}^{\lambda}\ra
\]

\subsection{Justifications of the claims}\label{sec:claim_justifications}
\textbf{Claim \ref{clm:stable_non_trivial}.} Stability during training when $\alpha+\gamma\geq1$ is straightforward (some technical conditions on the activation function are required). This is because the weight updates are of order $L^{-\alpha-\gamma}$ and feature updates involve no more than $L$ terms of size $L^{-\alpha-\gamma}$ (plus higher order terms that do not contribute to the update in the large depth limit). When $\alpha +\gamma >1$, the contribution of a sum of at most $L$ terms of order $L^{-\alpha - \gamma}$ will decrease to zero, and the network output $f_t$ will converge to $f_0$ in this case, yielding a trivial limit. However, when $\alpha + \gamma=1$, the updates remain important in the infinite depth limit, yielding a non-trivial limit.\\

\textbf{Claim \ref{clm:faithful}.} Consider a stable and nontrivial parametrization (i.e. $\alpha+\gamma=1$). Faithfulness at initialization is achieved only when $\alpha\geq 1/2$. This was proven in \cite{hayou21stable} in a more general setup. Faithfulness during training is ensured as long as $\alpha\leq 1$ because feature updates are always $\Theta(1)$ in depth. With $\alpha>1$, $\gamma<0$ and the weight updates explode with depth in this case, which yield exploding behaviour for $\hh$.\\

\textbf{Claim \ref{clm:redundant}} When $\alpha \in (1/2,1]$, we obtain smooth limiting dynamics when $L\to \infty$ as demonstrated in \Cref{heuristics:smooth_limits}. This limiting process is a smooth process (no Brownian jumps) that satisfies the required definition of redundancy.

\textbf{Claim \ref{clm:depth_mup}.} It remains to prove that Depth-$\mu$P is non-redundant. This is a result of the limiting dynamics in this case (\Cref{heuristics:depth-mup}) . With Depth-$\mu$P, the randomness of the initialization in the hidden layer remains present throughout training, inducing a Brownian-like term that breaks redundancy. 

\textbf{Claim \ref{clm:layerwise-linearization}.} In Depth-$\mu$P, $W_t^l-W_0^l$ is $\Theta(1/\sqrt L)$ which is much smaller than $W_0^l$. Therefore, $\phi(W^l_t\xx_t^{l-1})-\phi( W^l_0\xx_t^{l-1}) - \phi'(W_0^l\xx_t^{l-1})\odot ((W_t^l-W_0^l)\xx_t^{l-1})=o(1/\sqrt L)$, thus satisfies \Cref{defn:layerwise-linearization}.
Similar to the depth-$\mu$P case, for $\alpha \in [1/2,1)$, the activation in the forward pass can be linearized which indicates layerwise linearization when $\alpha+\gamma=1$.

\section{Additional Experiments}\label{sec:additional_exps}

\subsection{Failure of Standard Parametrization at Large Depths}

\begin{figure}[h]
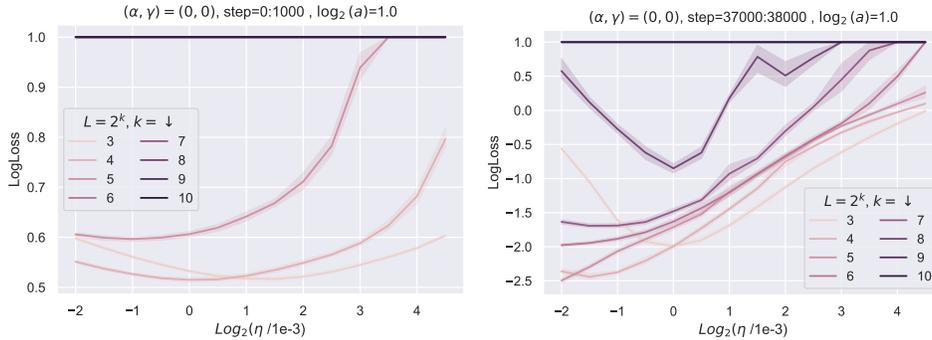

    \centering
    \includegraphics[width=0.45\linewidth]{fig/sp_0.pdf}
    \includegraphics[width=0.45\linewidth]{fig/sp_37000.pdf}
    \caption{Training with Standard Parametrization fails at large depth due to numerical issues.}
    \label{fig:sp_failure}
\end{figure}

\subsection{Experiments with Block Depth 2}
\begin{figure}
    \centering
    \includegraphics[width=0.45\linewidth]{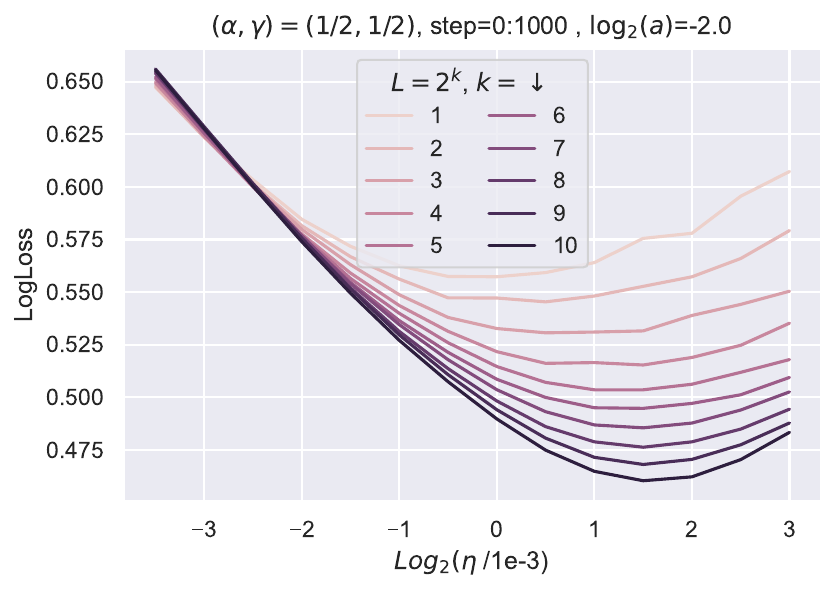}
    \includegraphics[width=0.45\linewidth]{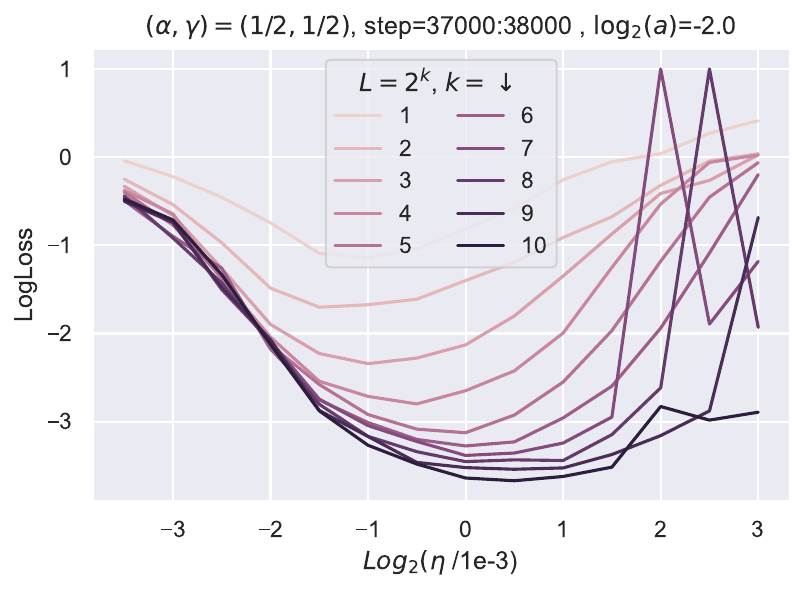}
    \caption{Same setup as \Cref{fig:lr_transfer}, with block depth 2 instead.}
    \label{fig:bd2}
\end{figure}

Currently, our theory covers resnets with block depth 1, and our experiments confirm the theoretical findings. We conducted similar experiments for clock depth 2 (i.e. the residual block consists of 2 fully connected layers) to see whether the learning rate transfers with Depth-$\mu$P. The results are reported in \Cref{fig:bd2}. The results show a significant shift in the learning rate which might indicate that as block depth increases, adjustments are needed to stabilize hyperparameters with depth.

\subsection{Other experiments}

\begin{figure}[h]
    \centering
    \includegraphics[width=\linewidth]{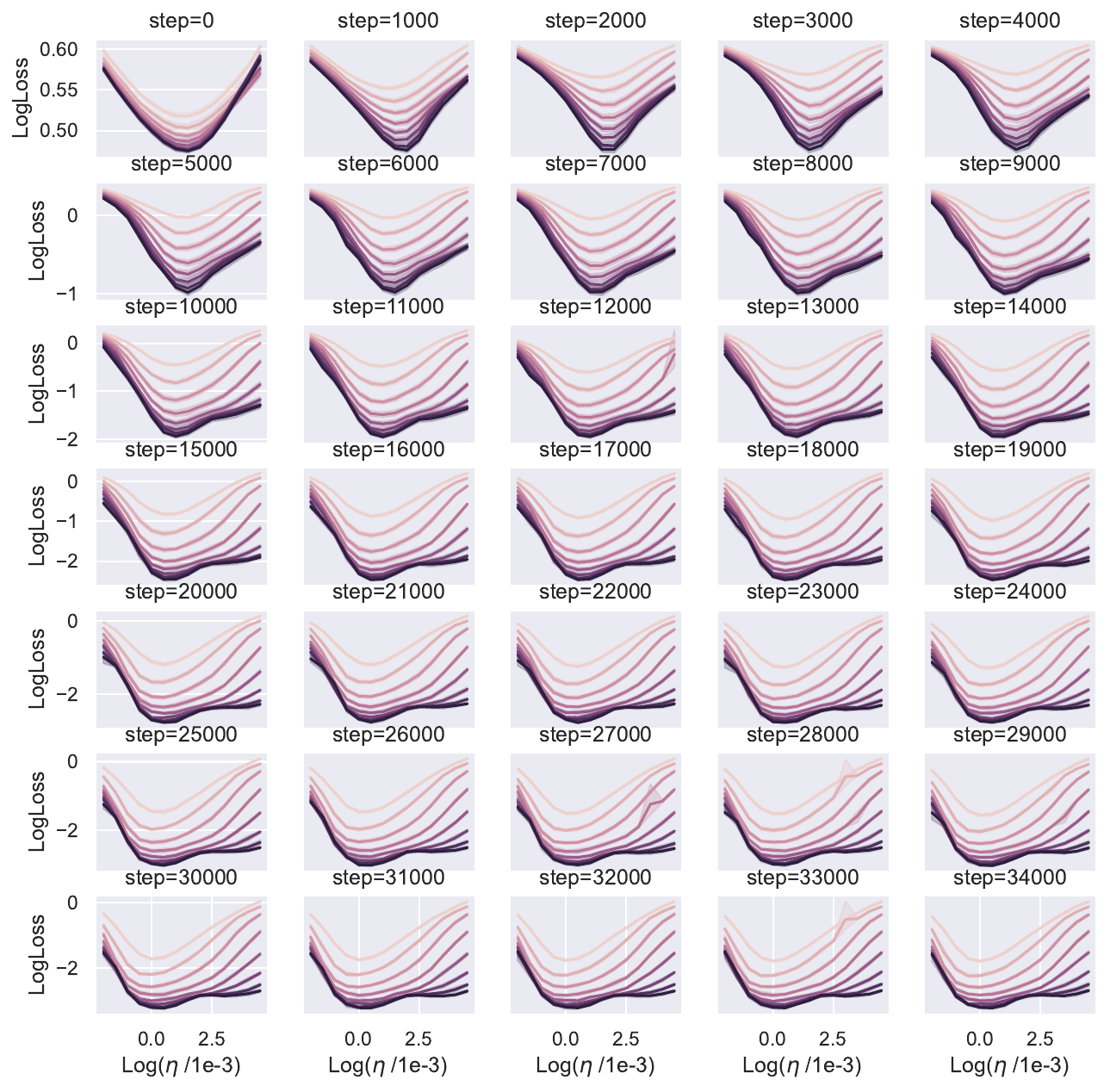}
    \caption{Same as \cref{fig:lr_transfer} (\textbf{Up}, Depth-$\mu$P) with multiple time slices.}
    \label{fig:lr_transfer_all_mup}
\end{figure}

\begin{figure}[h]
    \centering
    \includegraphics[width=\linewidth]{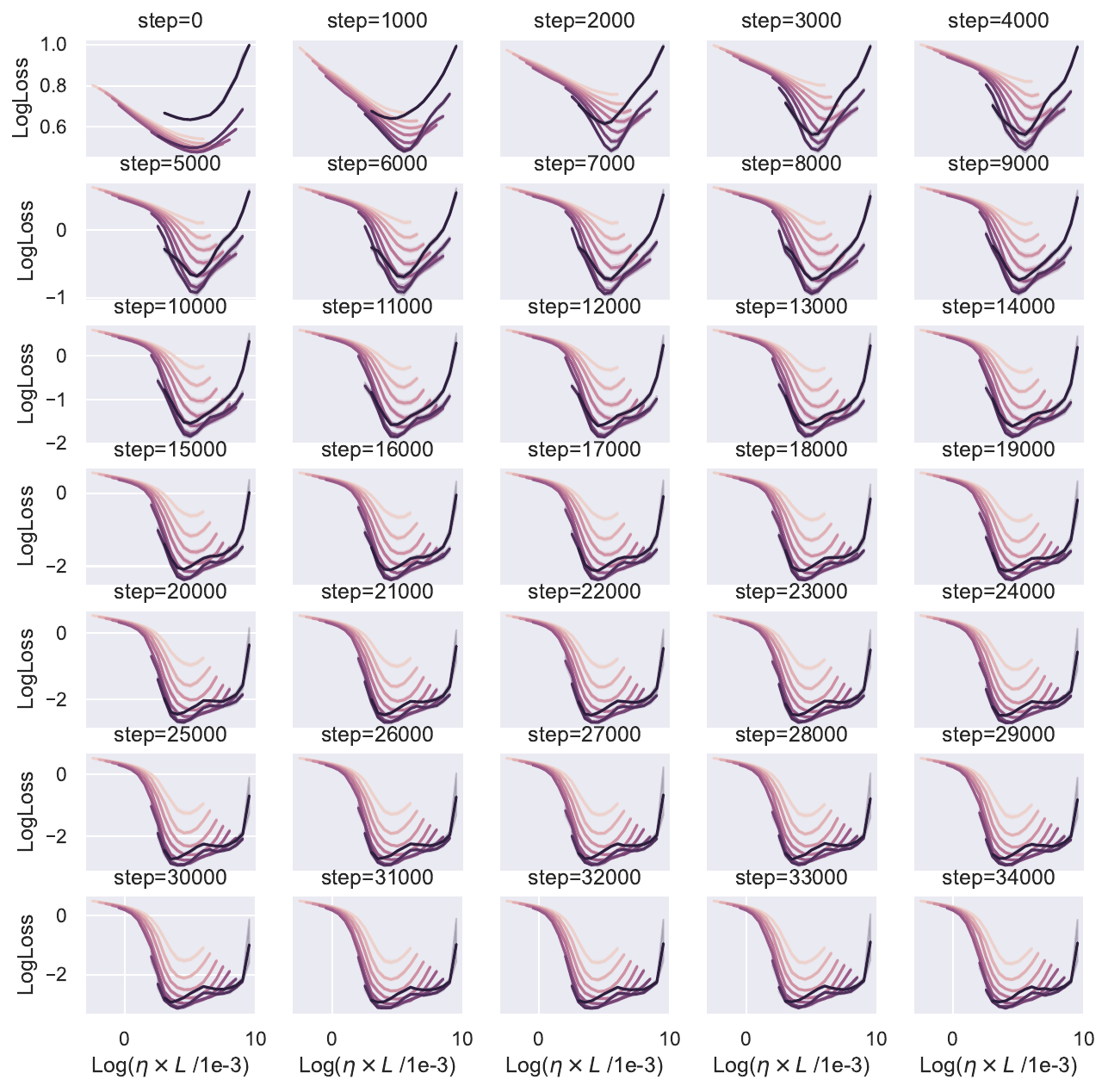}
    \caption{Same as \cref{fig:lr_transfer} (\textbf{Middle}, Standard Parametrization with $\gamma = 1$) with multiple time slices.}
    \label{fig:lr_transfer_all_normalized_sp}
\end{figure}

\begin{figure}[h]
    \centering
    \includegraphics[width=\linewidth]{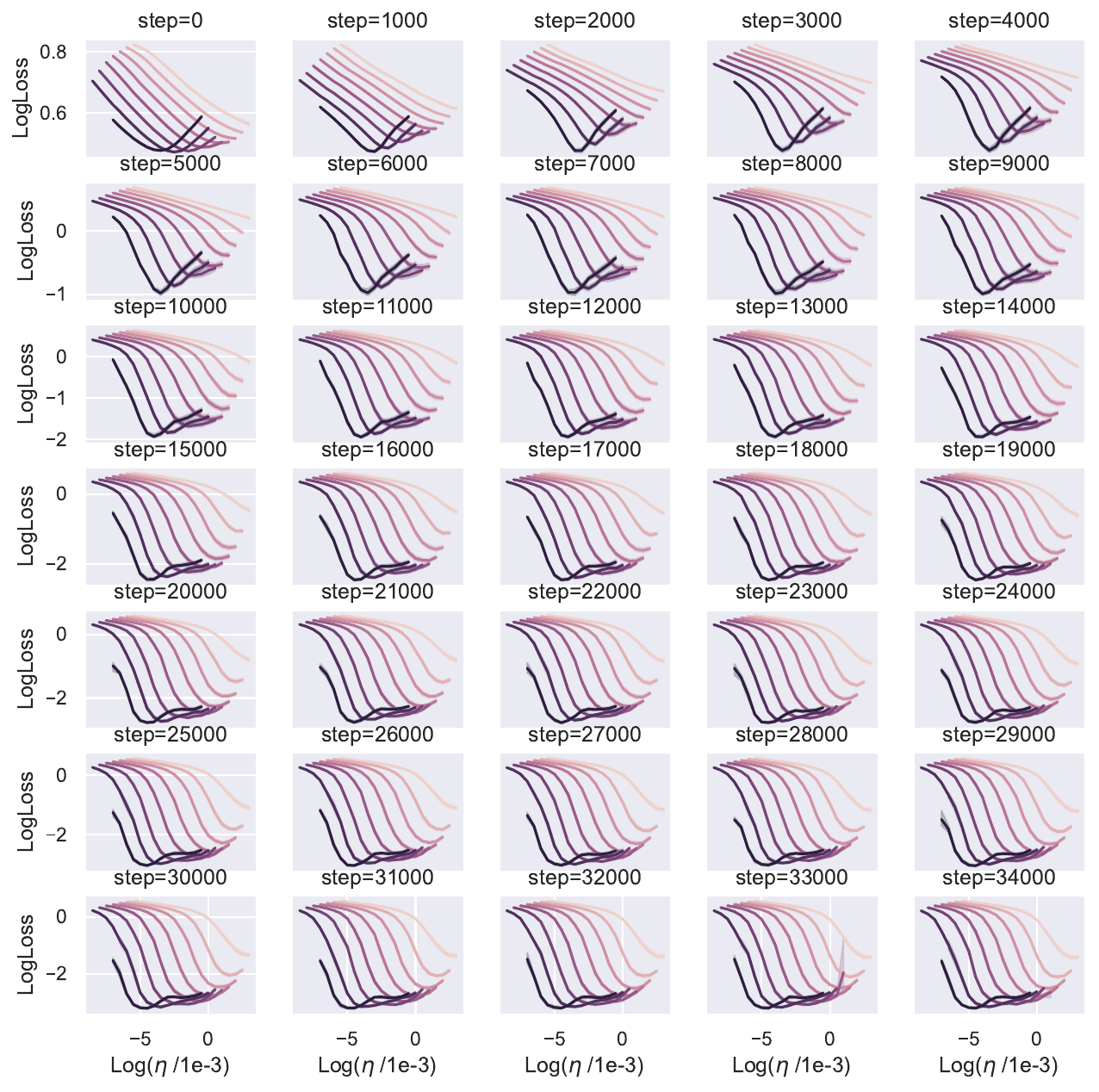}
    \caption{Same as \cref{fig:lr_transfer} (\textbf{Bottom}, Standard Parametrization with no scaling, $\alpha=0, \gamma=0$) with multiple time slices.}
    \label{fig:lr_transfer_all_sp}
\end{figure}

\begin{figure}
    \centering
    \includegraphics[width=\linewidth]{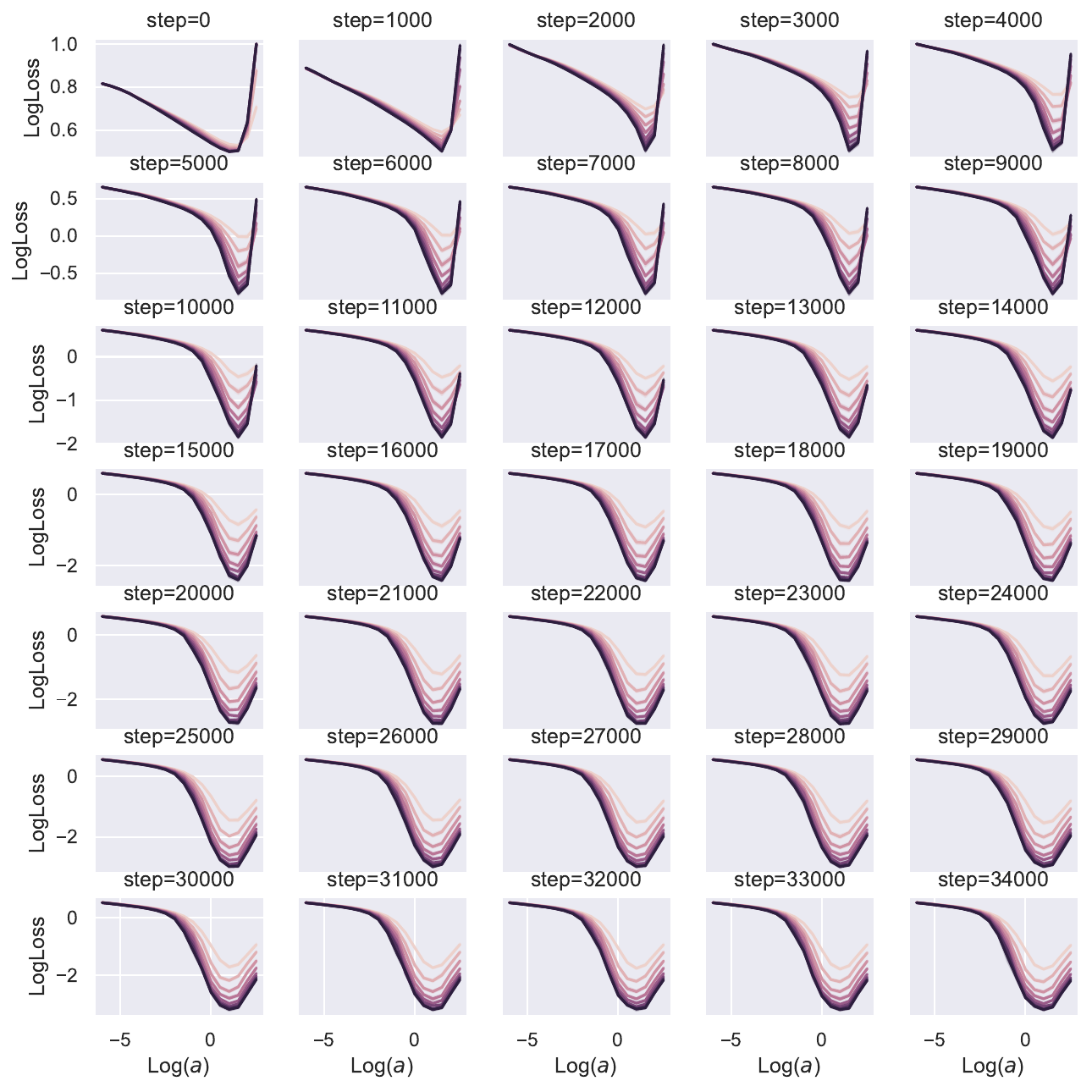}
    \caption{Same as \cref{fig:blockmult_transfer} with multiple time slices.}
    \label{fig:blockmult_transfer_all}
\end{figure} 

\end{document}